\documentclass[acmsmall,screen,nonacm]{acmart}
\AtBeginDocument{%
  }

\setcopyright{acmcopyright}
\copyrightyear{2025}
\acmYear{2025}
\acmDOI{XXXXXXX.XXXXXXX}

\acmJournal{IMWUT}
\acmVolume{0}
\acmNumber{0}
\acmArticle{0}
\acmMonth{0}

\usepackage{subcaption}
\usepackage{cleveref}
\usepackage{tabularx}
\usepackage{booktabs}
\usepackage{xcolor}
\usepackage{booktabs} % For professional-looking rules (\toprule, \midrule, \bottomrule)
\usepackage{siunitx}  % For the S column type (intelligent number alignment)
\usepackage{makecell} % For \thead (better multi-line table headers)
\usepackage{array} % Required for the fix
\usepackage{multirow} % For multi-row cells in tables
\usepackage{tikz}
\usetikzlibrary{shapes, arrows, positioning}

\begin{document}

\title{Exploiting Light To Enhance The Endurance and Navigation of Lighter-Than-Air Micro-Drones}

\author{Harry Huang}
\email{C.Huang-4@tudelft.nl}
\affiliation{%
  \institution{TU Delft}
  \city{Delft}
  \state{Zuid-Holland}
  \country{The Netherlands}
}

\author{Talia Xu}
\affiliation{%
  \institution{The University of Auckland}
  \city{Auckland}
  \country{New Zealand}
  }
\email{talia.xu@auckland.ac.nz}

\author{Marco Zúñiga Zamalloa}
\affiliation{%
  \institution{TU Delft}
  \city{Delft}
  \country{The Netherlands}
}
\email{m.a.zuniga.zamalloa@tudelft.nl}

\renewcommand{\shortauthors}{Huang et al.}

\begin{abstract}
% What is the problem?
Micro-Unmanned Aerial Vehicles (UAVs) are rapidly expanding into tasks from inventory to environmental sensing, yet their short endurance and unreliable navigation in GPS-denied spaces limit deployment. Lighter-Than-Air (LTA) drones offer an energy-efficient alternative: they use a helium envelope to provide buoyancy, which enables near-zero-power drain during hovering and much longer operation. LTAs are promising, but their design is complex, and they lack integrated solutions to enable sustained autonomous operations and navigation with simple, low-infrastructure.
\textit{We propose a compact, self-sustaining LTA drone that uses light for both energy harvesting and navigation.} Our contributions are threefold: (i) a high-fidelity simulation framework to analyze LTA aerodynamics and select a stable, efficient configuration; (ii) a framework to integrate solar cells on the envelope to provide net-positive energy; and (iii) a ``point-and-go'' navigation system with three light-seeking algorithms operating on a single light beacon.
Our LTA-analysis, together with the integrated solar panels, not only saves energy while flying, but also enables sustainable operation: providing 1 minute of flying time for every 4 minutes of energy harvesting, under illuminations of 80\,klux. We also demonstrate robust single-beacon navigation towards a light source that can be up to 7\,m away, in indoor and outdoor environments, even with moderate winds. The resulting system indicates a plausible path toward persistent, autonomous operation for indoor and outdoor monitoring. More broadly, this work provides a practical pathway for translating the promise of LTA drones into a persistent, self-sustaining aerial system.
\end{abstract}



\maketitle

\section{Introduction}

Micro-Unmanned Aerial Vehicles (UAVs) have expanded rapidly into a wide range of applications from aerial inspection~\cite{8067469, 7870116, nooralishahi2021drone} and environmental monitoring~\cite{s22207872, woodget2017drones, carpentiero2017swarm} to indoor inventory management~\cite{cristiani2020inventory, li2021reloc, kwon2019robust}. 
However, despite rapid progress, the capabilities of micro-drones, those weighing 250 g or less, remain constrained by two key limitations: short flight endurance and restricted navigation in GPS-denied environments. These shortcomings stem from \textit{stringent energy and weight constraints}: limited battery capacity and a high cost of hovering reduce flight times, and limitations on carrying extra sensors or performing computationally intensive tasks constraint the use of alternative navigation systems.

An emerging class of Lighter-Than-Air (LTA) drones, which integrates micro-drones with helium-filled envelopes~\cite{burri2013design, xu2023sblimpdesignmodeltranslational, cho2017autopilot, sharma2023beavis, gorjup2020low, zufferey2006flying}, provides a promising path forward to increase flight times and carry more advanced hardware. By leveraging buoyant lift to offset gravity, such platforms can hover at near-zero energy cost, extending endurance dramatically. This feature makes them ideal for periodic indoor operations, such as inventory scanning in warehouses, crop monitoring in greenhouses, or air-quality sensing in industrial spaces. 

Considering the unique advantages of LTA micro-drones, this work investigates whether we can exploit \textit{light, both natural and artificial}, to extend the flight time and assist the navigation of these novel platforms. Our central research question is:
How can light be leveraged as both an energy source and a navigational aid to enable long autonomous operation of LTA micro-drones, particularly in indoor environments?

\vspace{3mm}
\textbf{Research Challenges.} Answering this question requires tackling three intertwined research challenges:

\begin{itemize}
    \item \textit{Challenge 1: Selecting an efficient LTA architecture}. While LTA platforms reduce hover energy, the various designs proposed in the literature: balloon geometry, rotor placement, and control strategy, significantly influence maneuverability and drag. Yet, the literature offers no systematic comparison of LTA platforms. To build upon the best LTA design, we must first address an important question: Which LTA platform provides a suitable tradeoff between energy efficiency and maneuverability?

    \item \textit{Challenge 2: Integrating lightweight solar energy harvesting}. Although larger airframes can benefit from solar cell augmentation~\cite{chu2021development, articleLiller, oettershagen2016perpetual}, micro-drones are extremely sensitive to additional mass. Some attempts have been made to add solar cells to micro-drones~\cite{abidali2024development, goh2019100, articlelifeNano, manikandan2024parametric}. However, two critical questions remain open: What is the optimal balance between the solar cells' weight and harvested energy? And, what is the most efficient aerodynamic design to embed solar cells onto an LTA micro-drone?
    
    \item \textit{Challenge 3: Simplifying navigation using limited light sources}. Visible-light-based localization can achieve centimeter-level accuracy with drones but often requires dense lighting infrastructure~\cite{10257235, bastiaens2021received, zhu2024survey, almadani2019novel}. For typical indoor spaces with sparse fixtures, enabling navigation with a single light source is necessary but poses a significant challenge if it has to be done with a simple receiver and minimal changes in the lighting infrastructure.
\end{itemize}

\begin{figure}[t]
  \centering
  \includegraphics[width=\linewidth]{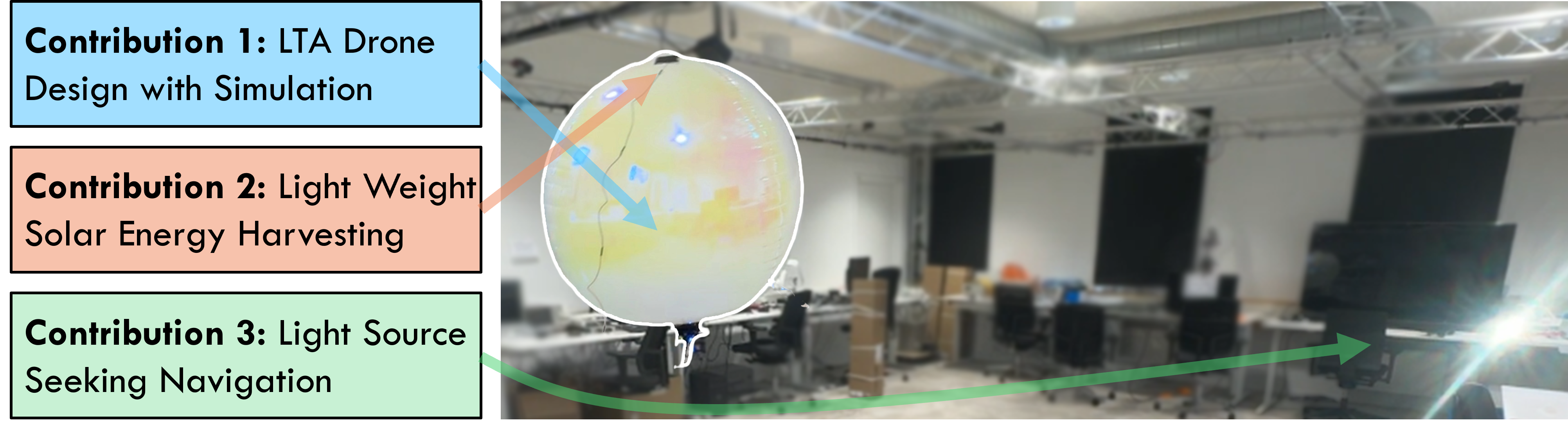}
  \caption{Main contributions: (1) Analyzing different LTA designs to identify the optimal one. (2) Enabling the placement of solar cells while balancing harvesting and aerodynamic properties. (3) Designing a robust navigation algorithm that works with a single light source.}
  \label{fig:intro_pic}
\end{figure}

\textbf{Contribution.} Considering the above challenges, our contribution is an end-to-end design for compact, self-sustaining LTA drones, as shown in \autoref{fig:intro_pic}. \textit{First}, we propose a novel high-fidelity simulator to model the complex aerodynamics of LTA micro-drones. Our simulation framework, validated empirically, allows us to identify the best configuration and control system that exploits buoyancy while maintaining stable flight. \textit{Second}, we analyze multiple types of solar cells, together with the aerodynamic effect on the envelope’s surface. Our analysis turns the LTA micro-drone into an energy-harvesting asset that extends endurance without compromising agility or stability. \textit{Finally}, we design a light-based ``point-and-go'' navigation mechanism that uses a single light source as a beacon. 
Our final evaluation shows two key results. 
\begin{itemize}

\item \textit{Contribution 1: A novel LTA platform that provides enough harvesting power to provide sustainable and autonomous operation of micro-drones.} Our novel simulator enables a thorough and reliable analysis of state-of-the-art (SOA) LTA platforms. After identifying the best design that trades-off energy-efficiency and maneuverability, we provide a framework to investigate the harvesting and aerodynamic properties of different solar cell configurations. Our final designs shows that, under illuminations of 80 klux (which resemble standard daylight conditions), our platform can enable sustainable operation: providing 1 minute of flying time for every 4 minutes
of energy harvesting.  

\item \textit{Contribution 2: A robust navigation system for indoor and outdoor scenarios with a single light anchor and with winds of 30\,km/h}. Even though our system is aimed at indoor scenarios, our strong focus on (i) selecting a compact and maneuverable LTA design, and (ii) analyzing three navigation algorithms that can operate with a single light anchor, allow our system to handle moderate outdoor winds of 30\,km/h (8\,m/s), making the platform reliable for operations that might traverse indoor-outdoor boundaries. Other SOA approaches report wind tolerance below 3.5\,m/s~\cite{burri2013design} and 8\,m/s~\cite{sharma2023beavis}, but without empirical validation. 
\end{itemize}

Overall, this paper introduces a practical pathway toward energy-autonomous, light-guided LTA micro-drones. By integrating buoyant design, solar energy harvesting, and minimalist optical navigation, we move a step closer to persistent, self-sustaining aerial systems capable of reliable operation in large indoor spaces where conventional micro-drones struggle\footnote{We demonstrate our drone's navigation performance in \url{https://github.com/LightEnhanceLTAmicodrone/Light-Enhance-LTA-micro-drone.git}. The code will be published in the same repository.}.

\section{Analysis of Lighter-Than-Air Platforms}
\label{sec:platform_comparison}

LTA platforms offer a strong foundation for endurance drones. Conventional micro-drone designs spend most of their energy simply staying airborne, constantly forcing air downward to counteract gravity. LTA systems eliminate this fundamental cost by relying on buoyant lift, requiring little power to maintain altitude. However, buoyancy poses a trade-off: the envelope that provides lift also creates aerodynamic drag once the drone moves. Overcoming this drag reduces maneuverability, in particular in the presence of wind.

Analyzing the trade-offs between buoyancy and maneuverability among the various alternatives in the SOA is far from straightforward. Comparing platforms fairly not only involves the slow process of building and modifying the physical prototypes but also the difficult task of maintaining the same conditions for a fair evaluation. 
Key parameters like buoyancy vary between tests due to imprecise helium filling and gradual leakage. These changes alter the system’s mass and lift, making results difficult to reproduce or compare. 
When testing different platforms it is hard to determine if a design change fails
because of its new aerodynamic properties, its new inertial properties, or simply because the test conditions
(like buoyancy) are different. This uncertainty can reduce the evaluation process to speculative trial and error. A new
methodology is therefore required: one that can decouple these variables so we can run fast, repeatable comparisons that enable reliable design decisions.

\subsection{Finding a Starting Point}

\begin{figure}[t]
    \centering % Center the entire figure content

    % --- A SINGLE ROW OF 5 SUBFIGURES ---

    \begin{subfigure}[b]{0.19\linewidth}
        \centering
        \includegraphics[width=\linewidth]{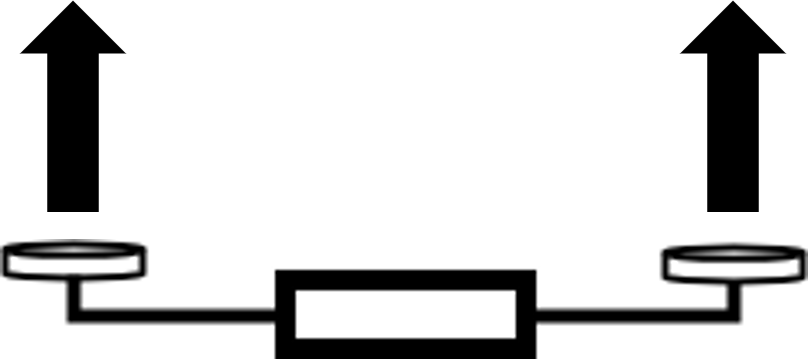}
        \caption{Baseline: no balloon}
        \label{fig:mutli-rotor}
    \end{subfigure}
    \hfill % Adds horizontal space
    \begin{subfigure}[b]{0.19\linewidth}
        \centering
        \includegraphics[width=0.6\linewidth]{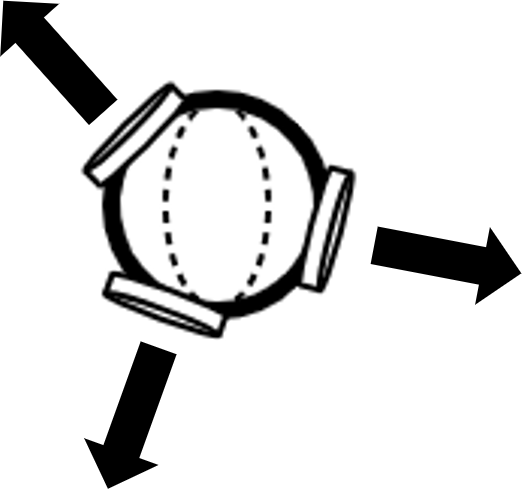}
        \caption{Skye}
        \label{fig:skye}
    \end{subfigure}
    \hfill % Adds horizontal space
    \begin{subfigure}[b]{0.19\linewidth}
        \centering
        \includegraphics[width=0.8\linewidth]{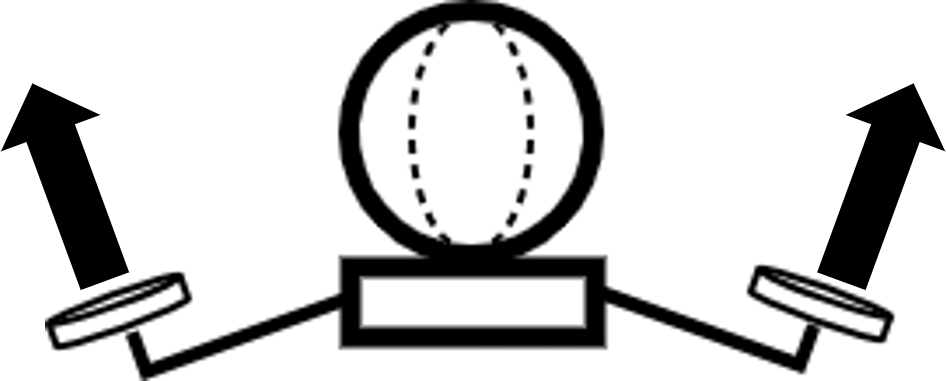}
        \Description{Image of SBlimp platform.}
        \caption{SBlimp}
        \label{fig:sblimp}
    \end{subfigure}
    \hfill % Adds horizontal space
    \begin{subfigure}[b]{0.19\linewidth}
        \centering
        \includegraphics[width=0.8\linewidth]{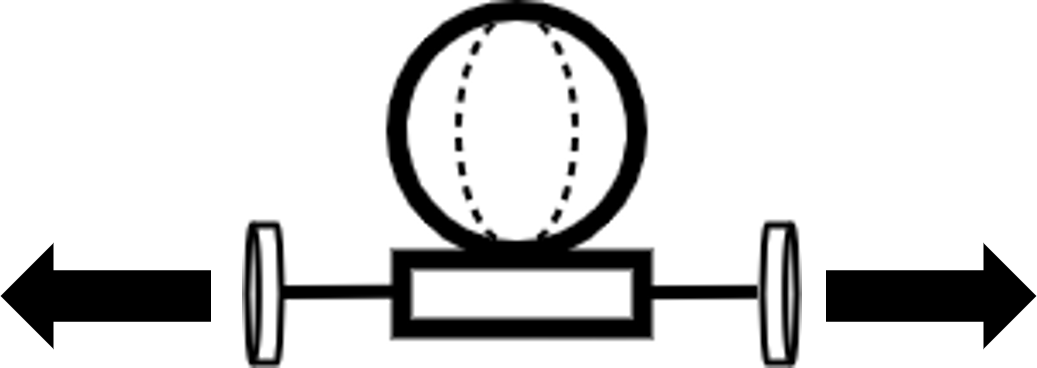}
        \caption{BEAVIS}
        \label{fig:beavis}
    \end{subfigure}
    \hfill % Adds horizontal space
    \begin{subfigure}[b]{0.19\linewidth}
        \centering
        \includegraphics[width=\linewidth]{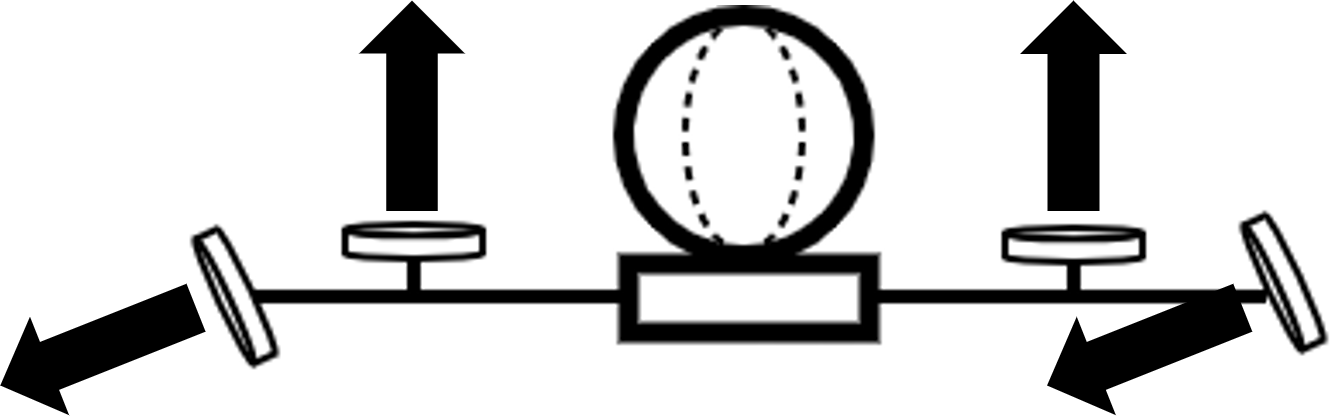}
        \caption{GT-MAB}
        \label{fig:GT_MAB}
    \end{subfigure}

    \caption{Propulsive System Comparison. All platforms have four rotors but aligned in different ways.}
    \label{fig:propulsivDifferent}
\end{figure}

To develop a solar-powered LTA platform, we first need to identify the best design to build upon. \autoref{fig:propulsivDifferent} shows the main aerodynamic designs, and \autoref{tab:uav_analysis} outlines their trade-offs alongside a standard drone baseline with four rotors. From specifications alone, such as maneuverability, size, and complexity, we can make some early calls. For example, Skye has a large physical size (2.7\,m diameter) and requires custom hardware to place the rotors in specific locations~\cite{burri2013design}, making it difficult to reproduce and less flexible in smaller space. 

The other three platforms have a similar diameter (0.9\,m), the same number of rotors (4) but different rotor alignments: tilted (SBlimp), lateral (BEAVIS), and lateral \& vertical (GT-MAB). 
The different alignments provide different abilities to perform the two most important drone movements: lateral and vertical.
SBlimp uses fixed tilted propellers, which is a simple design but cannot decouple vertical and lateral movements~\cite{xu2023sblimpdesignmodeltranslational}. In fact, the authors report that the design can sustain a maximum lateral speed of 0.8\,m/s. Beyond this value, the system becomes unstable as the tilted rotors cannot maintain simultaneously the desired (vertical) altitude and (lateral) speed. Due to this reason we discard the SBlimp platform from further analysis. 

Overall, two platforms provide a useful middle ground for our study: \textit{GT-MAB} and \textit{BEAVIS}. Both have simple mechanical structures but provide a different trade-off: BEAVIS is better at lateral movements, while GT-MAB is better at vertical movements. BEAVIS mounts all four propellers laterally. 
That yields strong and flexible lateral forces, while vertical lift depends on complex rotor-envelope airflow interactions. GT-MAB has two vertical propellers and two lateral ones, decoupling altitude control from lateral motion. 
The altitude control of GT-MAB is better than BEAVIS but the lateral motion is less flexible. To move into a given direction, GT-MAB must first rotate into the desired direction and then move forward. 

The design choices of BEAVIS and GT-MAB lead to different controller requirements and energy usage profiles that cannot be fully understood from specifications alone.
To make an informed choice, a direct comparison under identical conditions is necessary, but performing these tests physically is a significant challenge. The performance of LTA platforms varies because helium leaks are common and affect flight patterns~\cite{gorjup2020low}. Furthermore, if the helium leak exceeds a threshold, the parameters of the drone controllers need to be modified, requiring costly and iterative fine-tuning experiments~\cite{1453566}.
These shortcomings require the use of high-fidelity close-loop simulations.

\begin{table}[t]
\centering
\footnotesize
\caption{Comparative Analysis of Different LTA Platforms}
\label{tab:uav_analysis}
\begin{tabular}{@{}lcccccc@{}}
\toprule
\textbf{Platform} & \textbf{\begin{tabular}{@{}c@{}}Planar\\Movement\end{tabular}} & \textbf{\begin{tabular}{@{}c@{}}Vertical\\Movement\end{tabular}} & \textbf{\begin{tabular}{@{}c@{}}Atttude\\Movement\end{tabular}} & \textbf{Endurance} & \textbf{Size} & \textbf{Complexity} \\
\midrule
\textbf{Multi-Rotor Drone} & High & High & High & Low & Low & Low \\
\addlinespace
\textbf{Skye \cite{burri2013design}} & Medium & Medium & Medium & High & High & High \\
\addlinespace
\textbf{SBlimp \cite{xu2023sblimpdesignmodeltranslational}} & Medium Low & Medium & Medium Low & High & Medium & Medium \\
\addlinespace
\textbf{GT-MAB \cite{cho2017autopilot}} & Medium Low & Medium & High & High & Medium & Low \\
\addlinespace
\textbf{BEAVIS \cite{sharma2023beavis}}  & High & Medium  Low & Medium Low & High & Medium & Low \\
\bottomrule
\end{tabular}
\end{table}

\subsection{A Hybrid, Closed-Loop Simulation Framework}

Simulators are critical for the design of drone controllers and algorithms, and there multiple simulators for standard micro-drones ~\cite{shah2017airsimhighfidelityvisualphysical, kong2022marsimlightweightpointrealisticsimulator, yang2021digital, panerati2021learningflygym}. However, for the more nascent area of LTA-based platforms, there are only three simulators~\cite{8715570, zufferey2006flying, Price_2022}, and all suffer from one more of the following shortcomings: (1) Do not model the propeller forces, which prevents evaluating the various rotor configurations in the SoA, for example, the airflow that BEAVIS uses for lift; (2) Do not model damping and drag forces, which requires time-consuming physical experiments to determine the platform's drag, a process that must be repeated for any design change; and (3) Do not model close-loop behavior, which means that the simulator cannot predict whether a platform will actually be stable, hold altitude, or complete autonomous tasks such as position keeping or velocity tracking. Having a close-loop controller is central to evaluate a platform's performance. 

\begin{figure}[t]
  \centering
  \includegraphics[width=\linewidth]{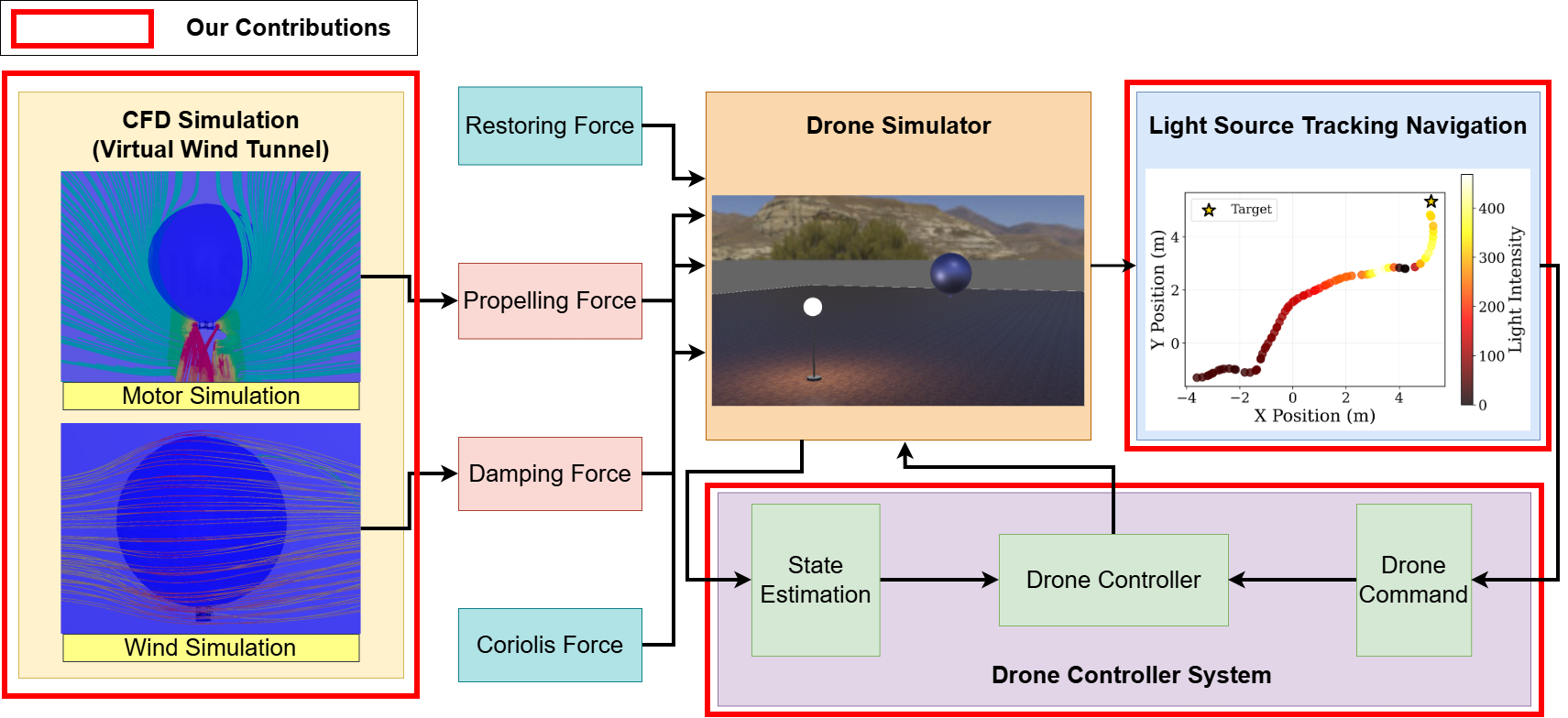}
  \caption{Simulation Pipeline. Our contributions are highlighted in red boxes. We provide (1) a CFD component (Computational Fluid Dynamics) to model propeller and damping (drag) forces from first principle physics and (2) a new drone controller system to analyze three different navigation algorithms that operate with a single light source.}
  \label{fig:oursimulation}
\end{figure}

Overall, the above limitations mean that existing simulators cannot capture how different designs will fly. Our framework introduces two key components to solve these problems: a CFD (Computational Fluid Dynamics) component to model propeller and damping forces; and a close-loop drone controller system. The next subsections describe our contributions, which are also highlighted in our full pipeline in \autoref{fig:oursimulation}. We then use this tool to evaluate GT-MAB and BEAVIS under identical tasks, validate simulation results with physical tests, and select the platform we carry forward for solar-cell integration and navigation.

\subsubsection{The baseline model and its limitations}
\label{sec:cfd-based-model}

Modeling a micro-drone’s motion requires accounting for the various forces acting on its body. Our simulator is based on the Newton–Euler dynamics framework proposed by Zufferey et al.~\cite{zufferey2006flying}, which describes the drone’s six degrees of freedom motion (6-DoF). This LTA model has the best documentation, and it is the most popular in the community. In this model, the drone's motion (i.e. its linear acceleration $\boldsymbol{\dot{v}}$) depends on the inertia of the system ($\boldsymbol{M}$) and the combined effect of four main external forces: Buoyancy ($\boldsymbol{F}_{R}$), Propellers ($\boldsymbol{F}_{P}$), Damping/Drag ($\boldsymbol{F}_{D}$), and Gyroscopic effects ($\boldsymbol{F}_{C}$).

\begin{equation}
\boldsymbol{M}\boldsymbol{\dot{v}} = \Sigma\boldsymbol{F}_{ext} = \boldsymbol{F}_{R} + \boldsymbol{F}_{P} + \boldsymbol{F}_{D} + \boldsymbol{F}_{C}
\end{equation}

While effective for the original platform, this baseline model has two key limitations that hinder its generalization to new drone designs.

\textit{(1) Propeller force ($\boldsymbol{F}_{P}$).} The SOA simulator does not explicitly model rotor aerodynamics. Instead, it measures the rotors' thrust empirically and represents them as a constant input force inside the simulator (in newtons). Although this simplification is adequate for a fixed configuration, it fails when the platform geometry changes—for example, when rotor alignment, body structure, or the proximity of rotors to the LTA envelope are modified. In these cases, the real aerodynamic interactions differ significantly, and the predefined thrust values no longer represent the actual forces.

\textit{(2) Drag force ($\boldsymbol{F}_{D}$).} The same issue arises for aerodynamic drag. In the baseline model, drag is measured empirically for only one platform (the authors’ prototype). Because drag is not modeled from first principles, these measurements lack generality and must be re-derived for each new design.

Together, these limitations prevent the simulator from accurately reproducing the behavior of modified or novel platforms such as BEAVIS or GT-MAB. Evaluating a new configuration requires repeating the full set of time-consuming physical experiments to recalibrate both propulsion and drag forces—making design iteration slow and cumbersome.

Our first major contribution is to eliminate this dependency on physical testing by replacing these empirical steps with a fast, reliable Computational Fluid Dynamics (CFD)–based model. With our approach, only a 3D model of the LTA drone is required (which is easy to obtain). A one-time, offline CFD analysis is then performed to extract aerodynamic parameters for both $\boldsymbol{F}_{P}$ and $\boldsymbol{F}_{D}$, providing a general, repeatable, and scalable foundation for simulating diverse LTA drone configurations.

\begin{figure}[t]
    \centering % Center the entire figure
    
    % First subfigure
    \begin{subfigure}[b]{0.48\linewidth}
        \centering
        \includegraphics[width=\linewidth]{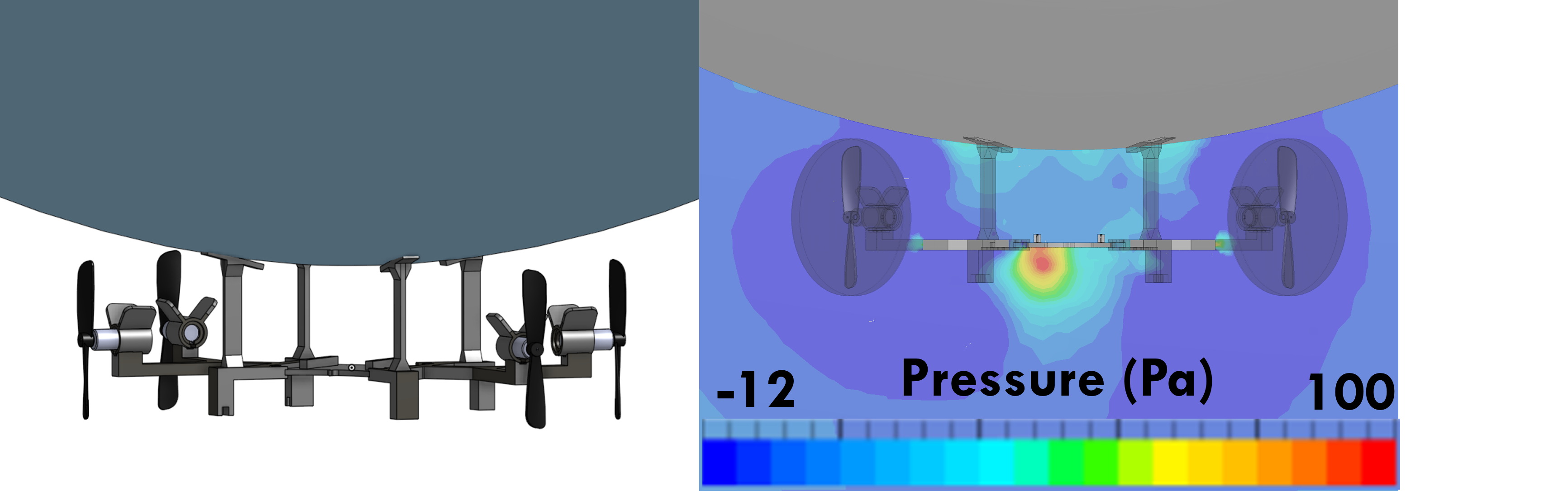}
        \Description{BEAVIS CAD model and CFD simulation result.}
        \caption{BEAVIS's CAD model and CFD simulation result.}
        \label{fig:CFD_beavis}
    \end{subfigure}
    \hfill % Adds horizontal space between the figures
    % NO BLANK LINE HERE
    % Second subfigure
    \begin{subfigure}[b]{0.48\linewidth}
        \centering
        \includegraphics[width=\linewidth]{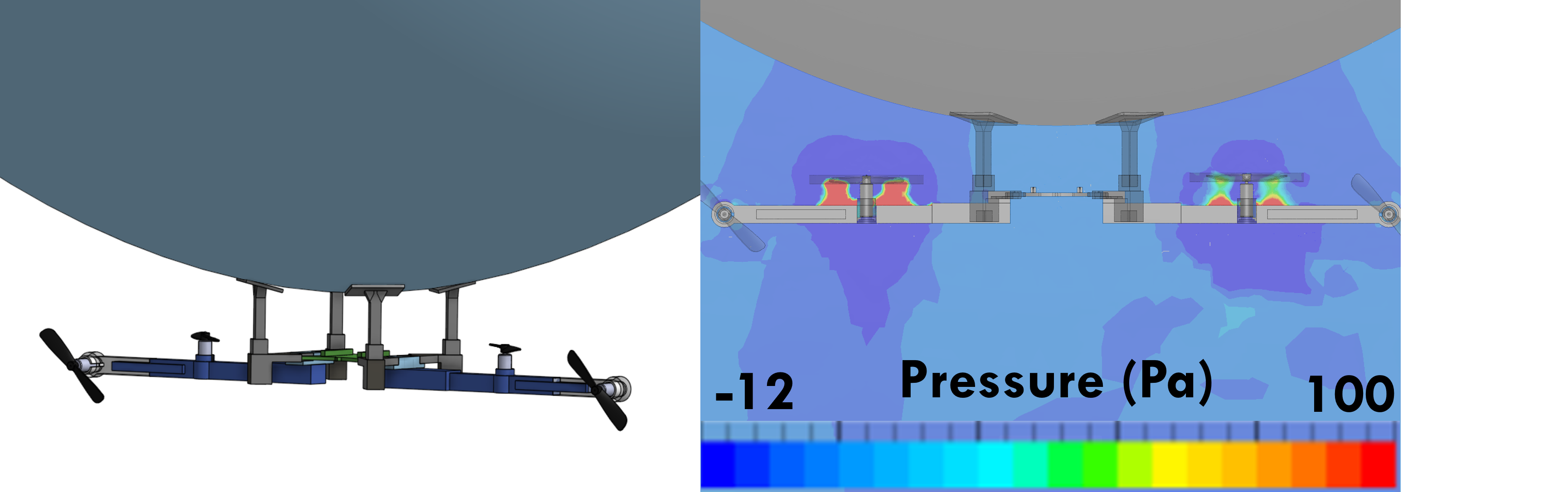}
        \caption{GT-MAB's CAD model and CFD simulation result}
        \label{fig:CFD_gtmab}
    \end{subfigure}
    \caption{Comparison of the BEAVIS and GT-MAB platform
designs with our enhanced simulator.}
    \label{fig:CFD_result}
\end{figure}

\subsubsection{Propeller Forces ($F_P$)} To accurately model the total propeller forces ($\boldsymbol{F}_P$), we need to address two key aspects. First, given the electronic signals provided to the propellers, we must determine the thrust produced by each rotor. Second, we need to combine the forces generated by all motors to produce the net push and twist (force and torque) that controls the entire platform.

\textit{For the first challenge}, calculating the motor thrust, the SoA model assumes a simple linear relationship between the electrical signal (Pulse Width Modulation, PWM) and the resulting force~\cite{zufferey2006flying}. However, real motors are non-linear. To capture this accurately, we adopt an empirical quadratic model based on data from Bitcraze \cite{bitcraze_platform_2025} (Bitcraze are the drones we use in our empirical evaluation):

\begin{equation}
    F_{thrust} = a * pwm^2 + b * pwm.
\end{equation}

Where $a=0.0915$ and $b=0.0677$ are component-specific coefficients provided by Bitcraze from a one-time physical test of the motor/propeller combination. Since these coefficients depend only on the rotor components, not the overall drone design, we can reuse them for any platform (like GT-MAB and BEAVIS). With this principled approach, we make the simulator more generalizable than methods requiring new empirical data for every design.

\textit{For the second challenge}, combining all the rotors' thrust, we must derive a function that translates the individual thrusts from all four motors (both GT-MAB and BEAVIS have four motors, $T_1...T_4$) into a net force and twist on the drone's body. This mapping function is the thrust allocation matrix ($\boldsymbol{B}$):

\begin{equation}
    % Complete propeller wrench equation
\boldsymbol{F}_{propellers} = \boldsymbol{B} \begin{bmatrix} F_{thrsut_{1}}. \\ F_{thrsut_{2}} \\ F_{thrsut_{3}} \\ F_{thrsut_{4}} \end{bmatrix}
\end{equation}

This matrix is highly design-specific because airflow from one propeller travel around the balloon, changing its effects. This is a critical interaction that the old model misses and is essential for LTA platforms. 
Our CFD approach solves this shortcoming by first running high-fidelity simulations of the aerodynamics of the 3D model (\autoref{fig:CFD_result}). After simulating all the motors in the CFD model, we can precisely measure the total resulting force ($\boldsymbol{F}_{propellers}$) and solve for the unique $\boldsymbol{B}$ matrix for that design. 
Our CFD results are validated against prior work. \autoref{fig:CFD_beavis} shows that the simulated airflow on the BEAVIS model reproduces the aerodynamic responses reported by the original work \cite{sharma2023beavis}: even though all rotors are laterally aligned, there is a single and small up-lifting force generated by the design (red center region). And \autoref{fig:CFD_gtmab} shows that GT-MAB creates four small forces with the two rotors that are aligned vertically (four small red regions). The SoA simulator cannot capture these critical responses~\cite{zufferey2006flying}.

\subsubsection{Damping/Drag Forces ($F_D$)} 

Our CFD approach is also used to model the second major force acting on LTA drones: aerodynamic damping (drag). This term represents the air resistance that opposes the drone’s motion—a dominant effect for LTA platforms because of their large surface area. Accurately capturing drag is essential for predicting both energy consumption and maneuverability.
In the original framework by Zufferey et al. ~\cite{zufferey2006flying}, drag was determined empirically through physical flight tests. The drone was flown at various constant speeds and directions, and the steady-state propeller thrust required to counteract air resistance was recorded. Although effective, this process is time-consuming and must be repeated entirely for any modification of the original design.

In our approach, these physical experiments are replaced by CFD-based simulations. The goal is to estimate the damping parameters of a standard six-degree-of-freedom model (6-DoF). Those parameters express drag as a function of the drone’s linear and angular velocities ($\boldsymbol{v,w}$). The linear components represent movements, and the angular velocities represent rotations. The damping model requires both linear and quadratic terms. Linear coefficients (e.g., $D_{v_{x}}$, $D_{\omega_{x}}$) dominate at low speeds, while quadratic coefficients (e.g., $D_{v_{x}^2}$, $D_{\omega_{x}^2}$) become significant at higher velocities, where drag increases nonlinearly. The complete drag model is expressed as:

\begin{equation}
    \boldsymbol{F}_D = -diag\begin{bmatrix}
        D_{v_{x}} + D_{v_{x}^{2}}\lVert \boldsymbol{v_{x}} \rVert \\
        D_{v_{y}} + D_{v_{y}^{2}}\lVert \boldsymbol{v_{y}} \rVert \\
        D_{v_{z}} + D_{v_{z}^{2}}\lVert \boldsymbol{v_{z}} \rVert \\
        D_{\omega_{x}} + D_{\omega_{x}^{2}}\lVert \boldsymbol{\omega_{x}} \rVert \\
        D_{\omega_{y}} + D_{\omega_{y}^{2}}\lVert \boldsymbol{\omega_{y}} \rVert \\
        D_{\omega_{z}} + D_{\omega_{z}^{2}}\lVert \boldsymbol{\omega_{z}} \rVert \\
    \end{bmatrix}\boldsymbol{v}
\end{equation}

\begin{figure}[t]
    \centering % Center the entire figure
    
    % Second subfigure (larger)
    \begin{subfigure}[b]{0.45\linewidth}
        \centering
        \includegraphics[width=0.9\linewidth]{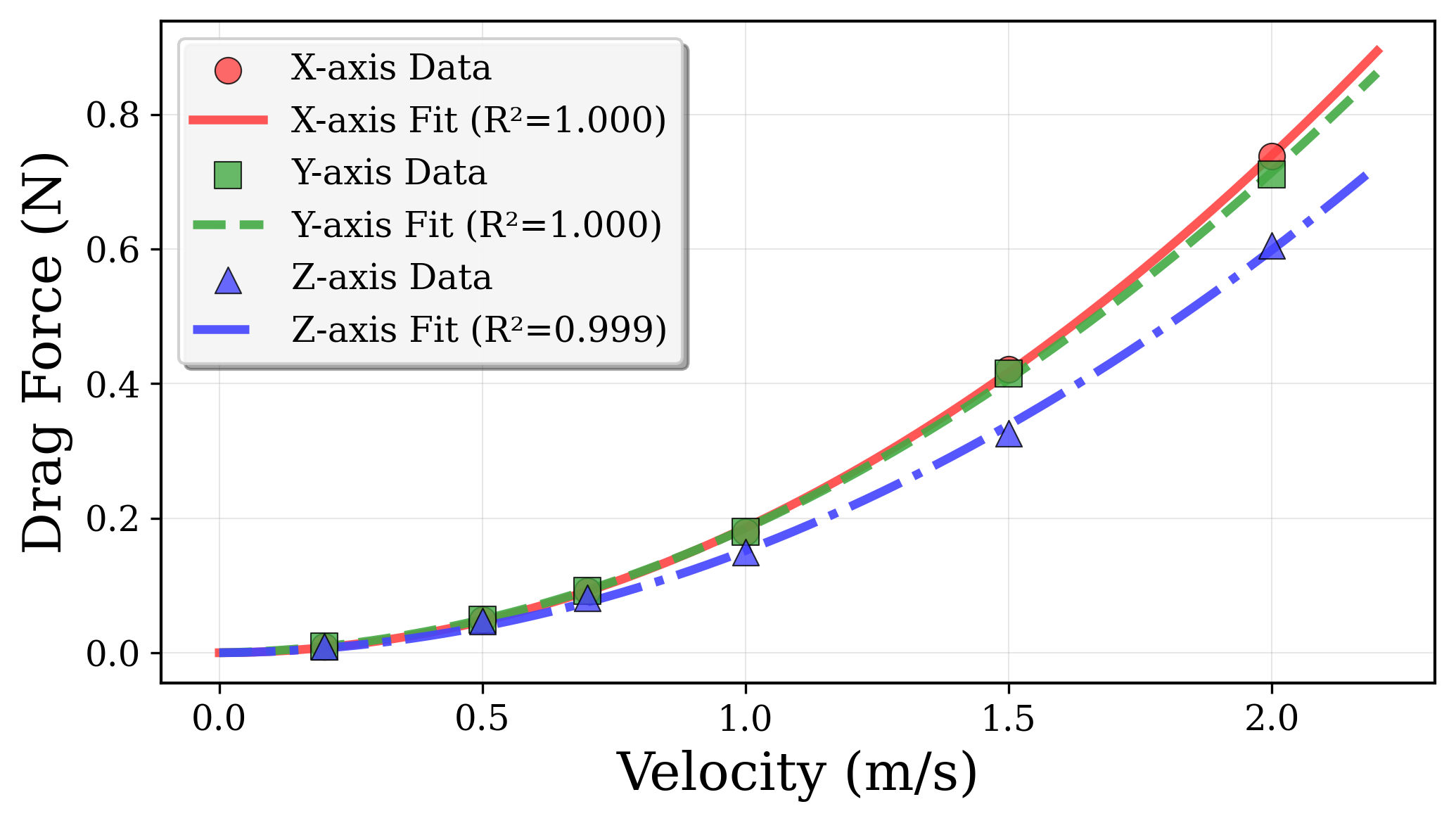}
        \Description{Plot showing translational drag characteristics.}
        \caption{Translational drag.}
        \label{fig:trans_drag}
    \end{subfigure}
    \hspace{5mm}
    \begin{subfigure}[b]{0.45\linewidth}
        \centering
        \includegraphics[width=0.9\linewidth]{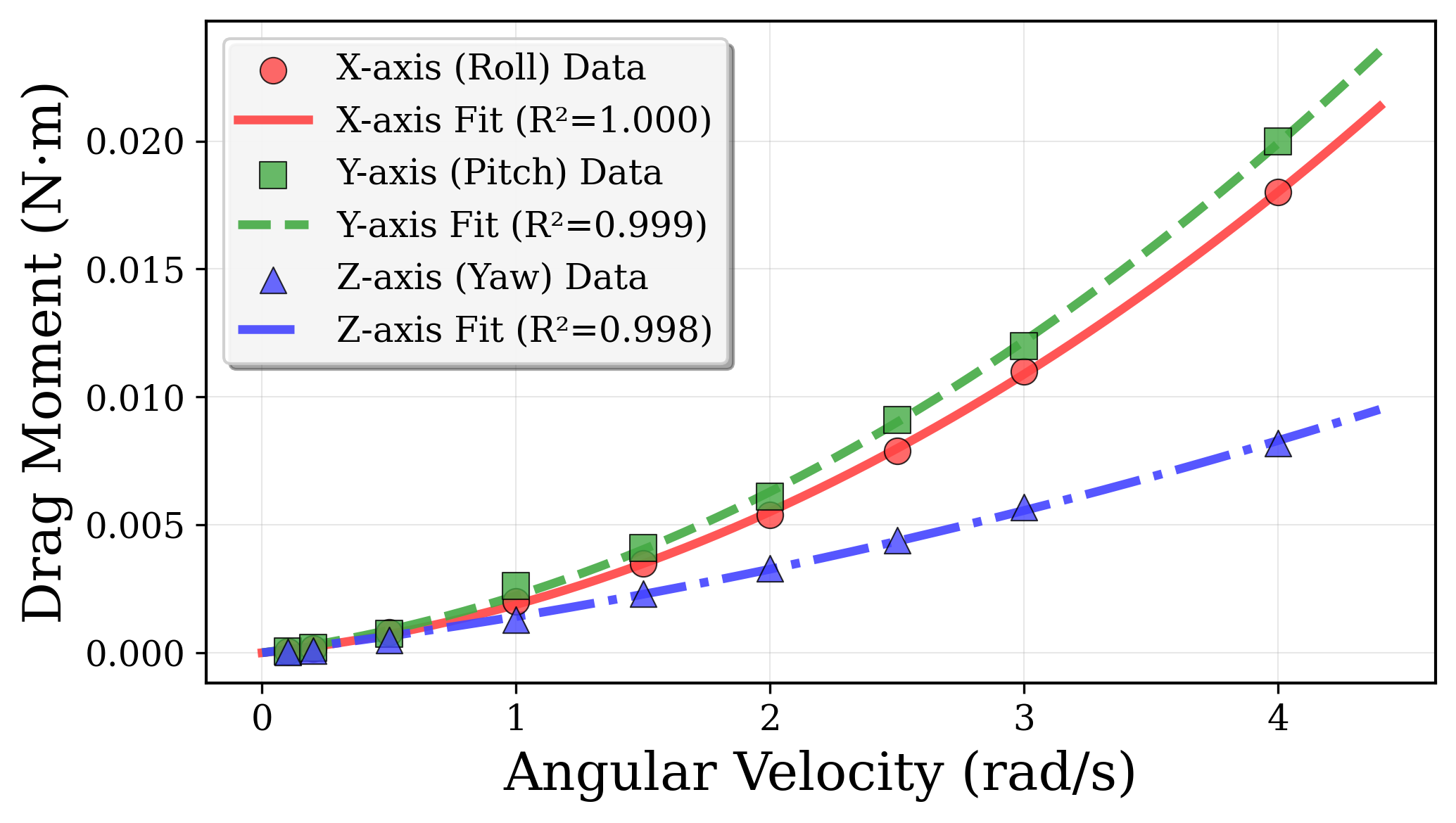}
        \caption{Rotational drag.}
        \label{fig:rot_drag}
    \end{subfigure}
    
    \caption{Estimation of linear and quadratic terms to simulate Damping/Drag forces.}
    \label{fig:drag_characterization}
\end{figure}

To estimate these coefficients, we use our CFD simulator to create a virtual wind tunnel and we place the drone’s 3D model inside the tunnel. The wind tunnel simulates airflow over the drone at multiple velocities—for example, around 0.2 m/s to capture linear terms and up to 2.0 m/s for quadratic effects. This process is repeated for all translational and rotational axes. The drag forces computed from the CFD simulations are then fitted to the model  to extract the full set of $D$ coefficients (as illustrated in \autoref{fig:drag_characterization}).

Overall, our CFD component produces a complete, design-specific model for both propeller and drag forces, without requiring a single physical flight. By replacing slow, hard-to-repeat experiments with high-fidelity CFD analysis, we enable a generalizable and precise physics foundation for the next stage of our work—developing and integrating a closed-loop controller for LTA drone simulation and control.

\begin{figure}[t]
    \centering
    \begin{subfigure}[b]{0.48\textwidth}
        \centering
        \includegraphics[width=0.8\textwidth]{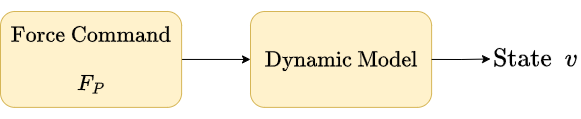}
        \caption{Open-Loop Model (Zufferey et al. \cite{zufferey2006flying})}
        \label{fig:open_loop}
    \end{subfigure}
    \hfill % Adds space between the two subfigures
    \begin{subfigure}[b]{0.48\textwidth}
        \centering
        \includegraphics[width=\textwidth]{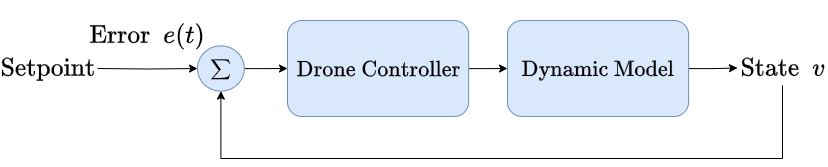}
        \Description{Closed-loop model diagram showing the inclusion of a flight controller.}
        \caption{Closed-Loop Model (Our Approach)}
        \label{fig:closed_loop}
    \end{subfigure}
    \caption{Comparison of an open-loop simulation model (a) versus a closed-loop model that includes a flight controller (b). The closed-loop approach is necessary to evaluate real-world performance.}
    \label{fig:control_loop}
\end{figure}

\subsubsection{A Closed-loop Controller}

The force model developed in the previous subsections describes how the drone responds to motor commands, but it cannot capture real flight behavior because the simulator does not provide a controller. Real flight dynamics depend on a flight controller—the drone’s “brain”—which continuously senses the drone’s state (e.g., altitude, attitude, and velocity) and adjusts motor commands to maintain stability, counteract disturbances, and perform specific maneuvers.

The original baseline simulator \cite{zufferey2006flying} is open-loop (\autoref{fig:open_loop}). This means that while it can compute some of the external forces acting on the drone, it lacks the feedback mechanism to provide real-time control. As a result, it cannot simulate how the drone actively compensates for drag, responds to buoyancy changes, or maintains altitude during navigation. Consequently, it cannot answer the most important practical questions, such as ``Will this design be stable?'' or ``How much effort will it take to hold altitude?''
Our second major contribution is therefore to close the loop by integrating a flight controller into the simulation (\autoref{fig:closed_loop}). %This addition transforms the framework from a passive force calculator into a true flight dynamics simulator, enabling the comparison of different drone designs and the evaluation of autonomous navigation algorithms.

\begin{figure}[tb!]
  \centering
  \includegraphics[width=0.7\linewidth]{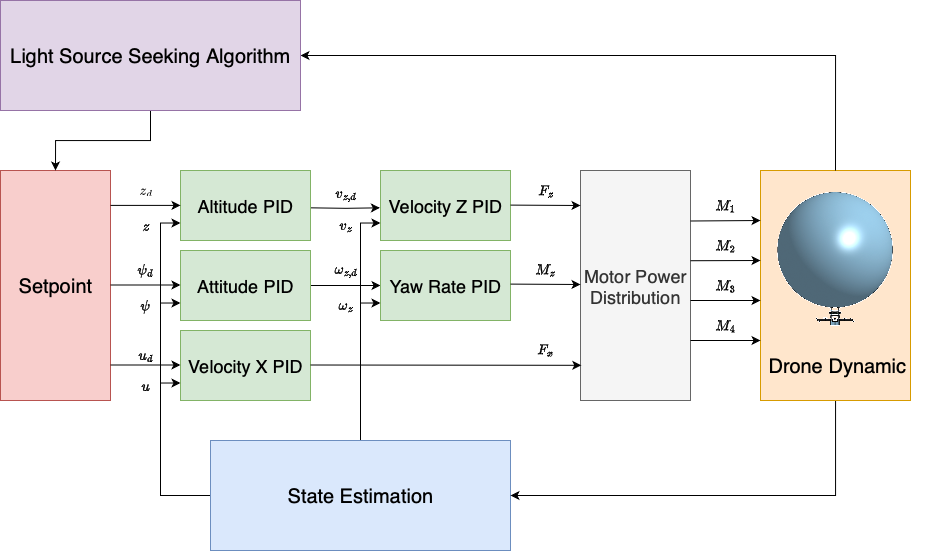}
  \caption{Block diagram of the drone's flight control with the navigation algorithm (Light Source Seeking).}
  \label{fig:drone_controller}
\end{figure}

To ensure a fair evaluation across platforms, we first standardize the control architectures used in our experiments. For the BEAVIS platform,  this is straightforward: we directly adopt the controller proposed by its original authors~\cite{sharma2023beavis}. For GT-MAB, however, no existing controller offered the right balance of simplicity and generality. Prior implementations are either too simplistic—such as basic PID controllers prone to oscillations~\cite{gorjup2020low}—or too specialized, relying on complex gain-scheduling strategies tied to specific hardware. To address this gap, we implement a cascaded PID control architecture for GT-MAB (\autoref{fig:drone_controller}). This is a
widely used and highly stable controller design that balances performance with simplicity. %Its strength comes from a two-loop hierarchical structure: an outer loop computes the target velocity required to reach the desired position, while a faster inner loop drives the motors to achieve that velocity. This separation minimizes overshoot and results in a smooth and stable flight, which is required for autonomous tasks, such as navigation.

Our closed-loop framework also resolves a critical challenge for LTA platforms: controller tuning due to helium leakage. On physical LTA-drones, PID tuning is slow and error-prone because buoyancy changes between tests (due to helium leakage). The changes in buoyancy alter flight characteristics, making it unclear whether poor performance arises from the controller or the drone. In our simulator, tuning becomes both fast and consistent. Since the controller can be tuned with precise drag and thrust forces (thanks to our CFD model),  we can obtain all the controller gains in
our fast virtual environment and then transfer those exact gains to the physical drone. This approach significantly reduces real-world setup and testing time.

\begin{figure}[t]
    \centering
    \begin{subfigure}[b]{0.45\linewidth}
        \centering
        \includegraphics[width=0.9\linewidth]{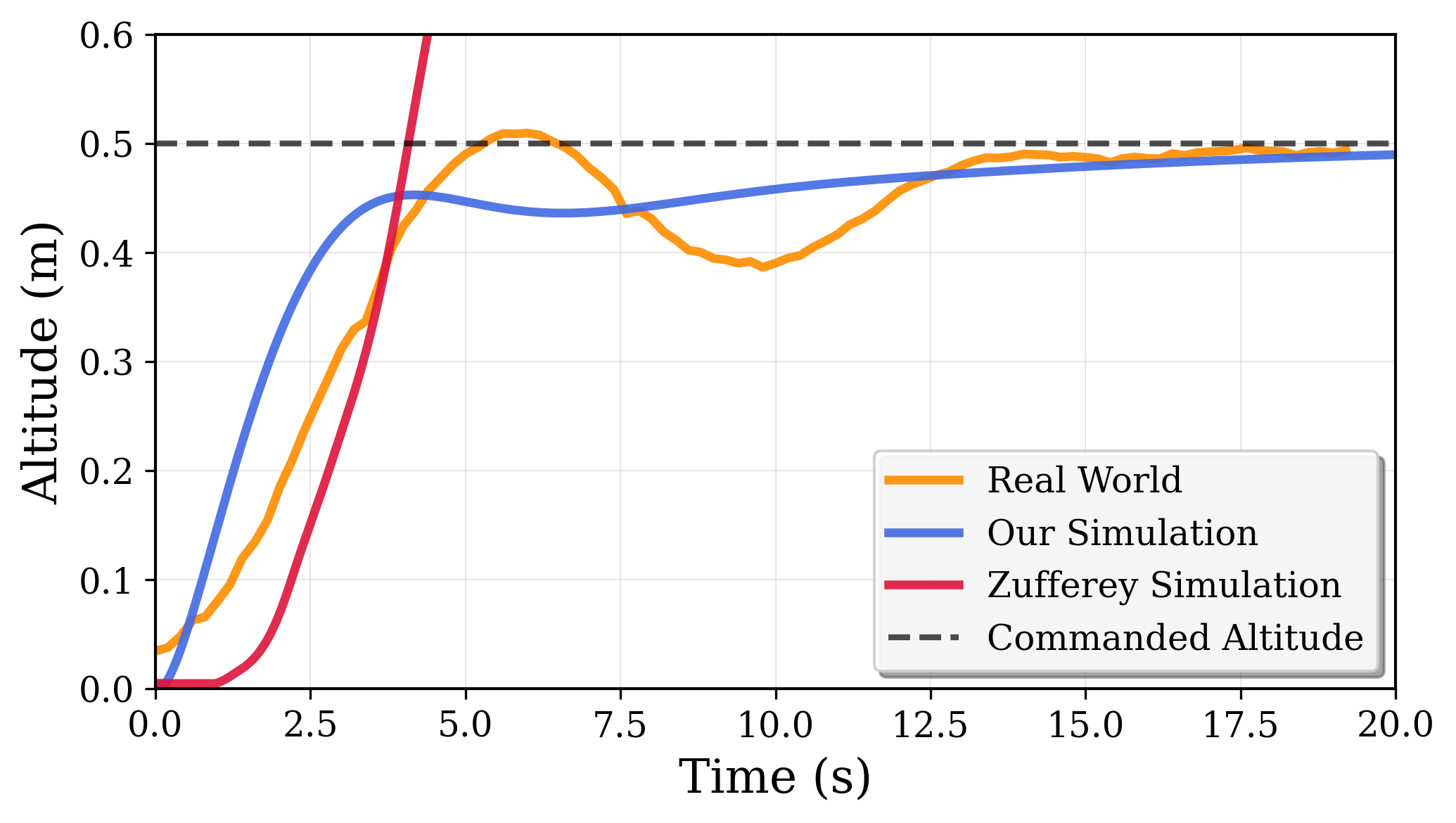}
        \Description{Simulation performance for altitude.}
        \caption{Altitude task}
        \label{fig:altitudeEva}
    \end{subfigure}
    \hspace{5mm} 
    \begin{subfigure}[b]{0.45\linewidth}
        \centering
        \includegraphics[width=0.9\linewidth]{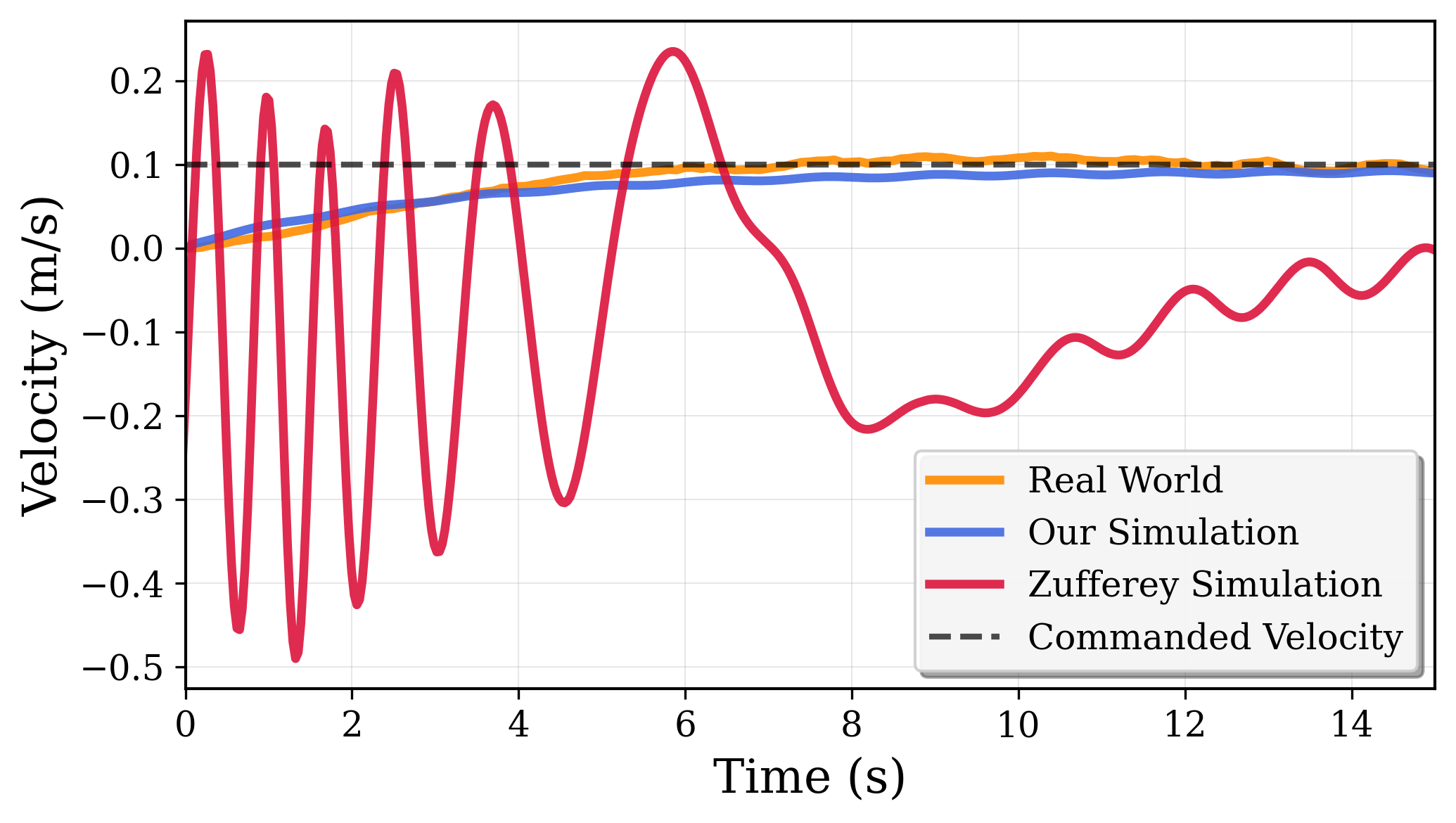}
        \caption{Velocity task}
        \label{fig:velocityEva}
    \end{subfigure}
    \caption{Performance comparison of simulation results and real-world performance. Zufferey is the SoA~\cite{zufferey2006flying}.}
    \label{fig:sim2realPerform}
\end{figure}

To test our closed-loop simulator, we perform two common tasks on drones: altitude-hold and velocity-tracking. This preliminary evaluation only considers the GT-MAB platform; in the next section, we thoroughly compare BEAVIS and GT-MAB.
For GT-MAB, the cascaded PID controller is tuned entirely in simulation and then applied unmodified to the real drone. The results, shown in \autoref{fig:sim2realPerform}, highlight the predictive accuracy of our approach: the baseline open-loop simulation (red) is unstable and fails to follow the reference commands. In the altitude task, the baseline overshoots, and in the velocity task, the baseline is unstable. On the other hand, our closed-loop simulation (blue) closely matches the real-world performance (orange). It correctly captures the initial overshoot and settling time for the altitude task and matches the stable tracking for the velocity task.

Our overall results validate that our CFD-derived models and closed-loop control architecture jointly provide a high-fidelity, predictive simulation environment. This allows controllers to be tuned virtually and deployed directly on physical LTA drones with minimal additional adjustment.

\subsection{Comparing LTA Platforms: BEAVIS and GT-MAB}

With our validated closed-loop simulator, we can now revisit one of our core research questions: which platform, BEAVIS or GT-MAB, provides a more stable and controllable baseline? To answer this question, we perform a comprehensive comparison, considering simulation and physical tests.

% \subsubsection{Static Force Analysis. A ``virtual bench test'' in our simulator to determine which design is better at generating vertical and horizontal forces.} 
% \tocheck{To write. Pending horizontal force results.}

%To determine which platform has better vertical movement performance, we analyzed their aerodynamics using a CFD simulation. The results, visualized in Figure~\autoref{fig:CFD_result}, reveal two distinct lift performance. The BEAVIS platform creates a wide and mild high-pressure zone (max 102.0 Pa) beneath its body. In contrast, the GT-MAB concentrates a much stronger, narrow area of high pressure (max 571.9 Pa) directly below its vertical propellers. However, it is not clear how this pressure generates the lift force. Thus, to quantify what this pressure difference means for actual flight, we calculate the total upward force produced by each design. The GT-MAB generated a net upward force of 0.62 N. The BEAVIS platform generated only 0.05 N. This shows that the GT-MAB has a superior vertical maneuver ability. This analysis also shows that the GT-MAB has a greater capacity to carry a payload since it can generate more vertical force required to take off.

%We further evaluated both platforms in our custom simulation environment. For each test, the drones were configured to be 2g negatively buoyant to account for potential helium leakage during real-world operation. We first command the drone to hover at a given altitude. Second, to see the rotating ability, we command the drone to rotate and maintain a specific heading.

\subsubsection{Closed-Loop Simulation: Performing a dynamic flight test in our simulator to determine how both platforms perform under autonomous control.}
The goal is to determine which aerodynamic design translates into practical flight stability. Both original studies claim good maneuverability, but, which one is the best? Is it better to have all rotors laterally (BEAVIS) or combine them laterally and vertically (GT-MAB)?
We run BEAVIS and GT-MAB in our simulator with their respective controllers, and we give them two tasks: 1) ascend and hold a target altitude, and 2) rotate and hold a target heading.

To create a realistic setup, both drones are configured with a 2\,g negative buoyancy, representing a common real-world condition where minor helium leakage makes the platform slightly “heavy”. This scenario tests whether the drone can maintain control under imperfect buoyancy—a likely operational state. For GT-MAB, which provides ample vertical thrust (0.62 N), we expect smooth compensation for this deficit. For BEAVIS, however, with its limited vertical thrust (0.05\,N), even a small buoyancy loss could compromise stability and maneuverability.

\begin{figure}[t]
    \centering % Center the entire figure
    % First subfigure
    \begin{subfigure}[b]{0.45\linewidth}
        \centering
        \includegraphics[width=0.9\linewidth]{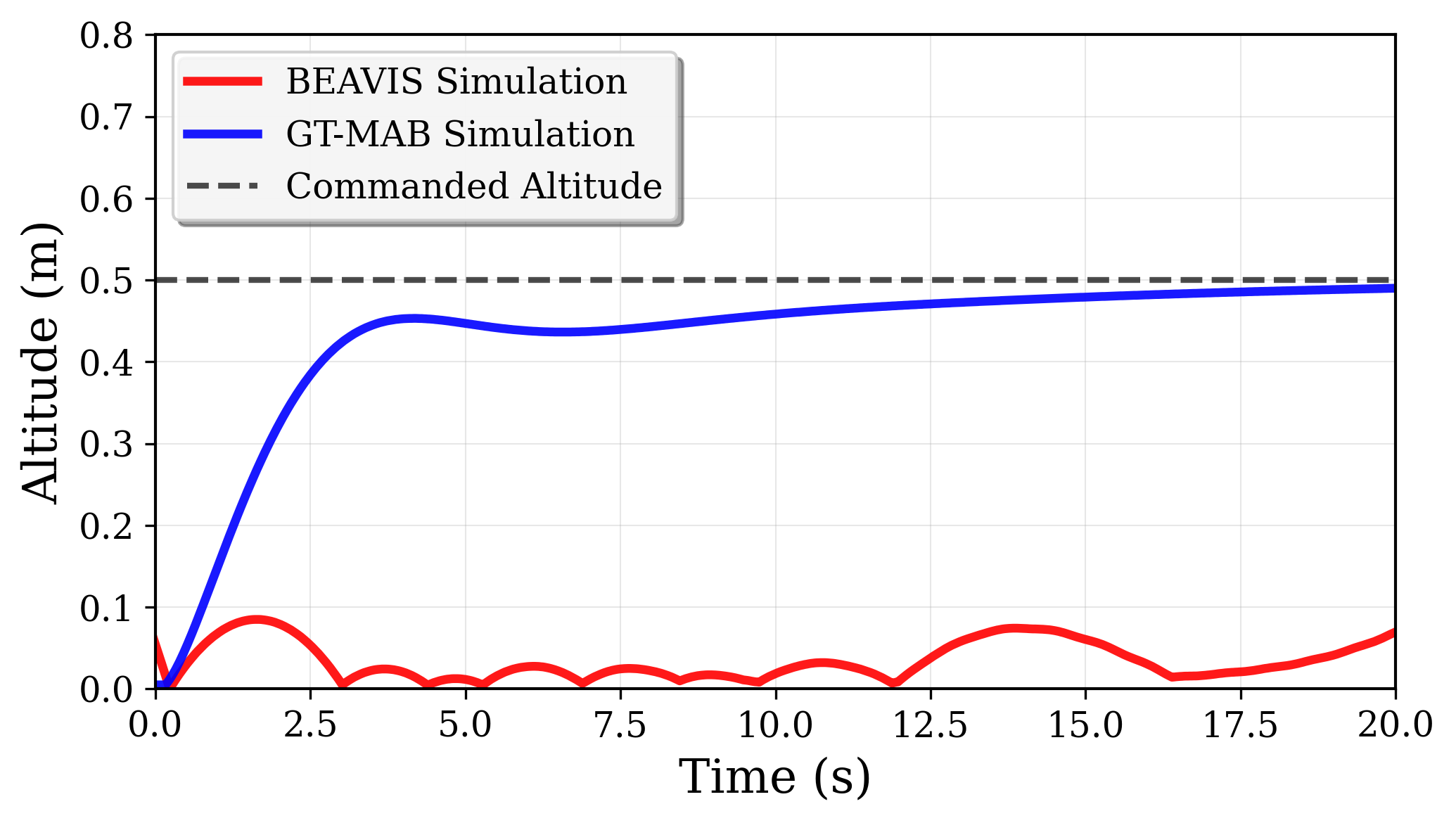}
        \Description{Altitude maintenance comparison plot showing the performance of BEAVIS and GT-MAB in simulation.}
        \caption{Altitude maintenance comparison.}
        \label{fig:altitudeSim}
    \end{subfigure}
    \hspace{5mm} % Adds horizontal space between the figures
    % NO BLANK LINE HERE
    % Second subfigure
    \begin{subfigure}[b]{0.45\linewidth}
        \centering
        \includegraphics[width=0.9\linewidth]{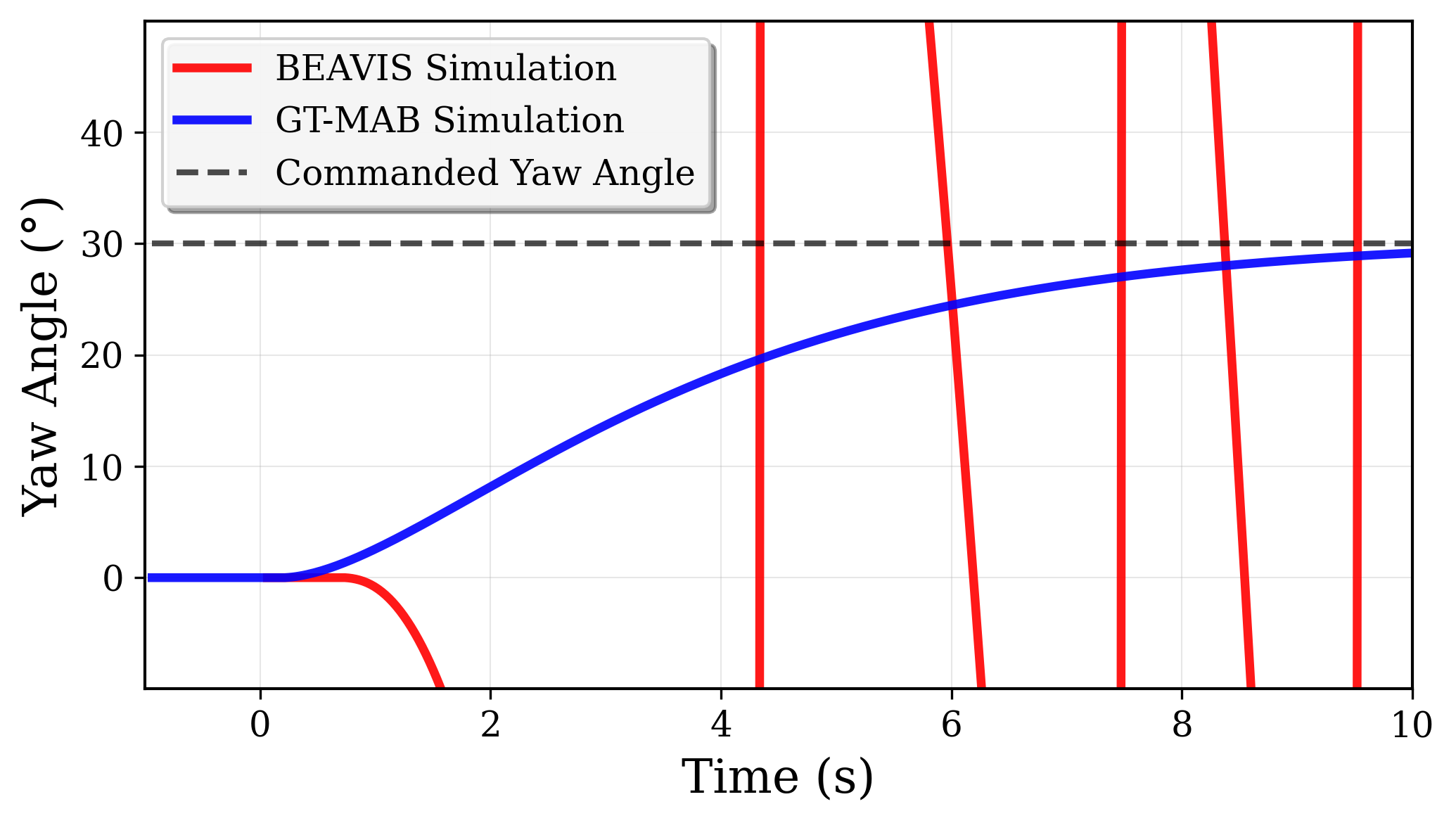}
        \caption{Yaw angle response comparison.}
        \label{fig:YawSim}
    \end{subfigure}
    \caption{Performance comparison of BEAVIS and GT-MAB in simulation}
    \label{fig:compareSim}
\end{figure}

The simulation results in \autoref{fig:compareSim} confirm this hypothesis. During the altitude-hold task (\autoref{fig:altitudeSim}), GT-MAB achieves a smooth ascent and maintains its target altitude with minimal oscillation. In contrast, BEAVIS struggles to take off and cannot reach the commanded height. The yaw test (\autoref{fig:YawSim}) shows a similar outcome. While GT-MAB rotates steadily and holds the target angle, BEAVIS exhibits uncontrolled rotation and fails to stabilize. These results reveal that BEAVIS, which depends on near-perfect buoyancy, is poorly suited for practical use. Its aerodynamic design and controller cannot simultaneously handle small disturbances and stabilization tasks. GT-MAB, on the other hand, maintains robust control even under non-ideal conditions, making it a more reliable candidate for an autonomous platform.

%
%This 2g deficit is a deliberate stress test that links directly to our CFD analysis. For GT-MAB, which has ample vertical thrust (0.62 N), we expect it to have no problem compensating for this 2g deficit. For BEAVIS, however, with its minimal vertical thrust (0.05 N), this 2g deficit might already derail its operations, leaving little to no margin for stabilization or maneuvering.

%The simulation results in \autoref{fig:compareSim} confirm this hypothesis. For the altitude task (\autoref{fig:altitudeSim}), the GT-MAB demonstrates a smooth and stable ascent, locking onto the commanded altitude. The BEAVIS platform, however, struggles to take off and reach the desired altitude. The yaw test (\autoref{fig:YawSim}) shows a similar outcome: the GT-MAB rotates smoothly and holds the angle, while the BEAVIS platform is unstable, rotating continuously and unable to hold a fixed angle.
%
%These finding demonstrate that the BEAVIS design, which relies on perfect buoyancy, is not well-suited for real-world conditions. Its coupled control system struggles to handle both a small, realistic disturbance (the 2g deficit) and its stabilization tasks simultaneously. The GT-MAB design, on the other hand, is more stable and controllable even under these non-ideal conditions, making it a more reliable candidate for an autonomous platform.

\subsubsection{Physical Validation: A real-world flight test with a more challenging yaw task.}
The simulation results clearly show that GT-MAB is the most versatile platform. We now need to verify that this significant performance gap holds true on actual flights.
To do this, we build 3D structures for both platforms, following the dimensions and guidelines of the original papers. Both platforms use the same type of envelope and perform two experiments to assess their real-world flight performances. For the first experiment, we perform the same altitude-hold task as in the simulation. For the yaw experiment, we make the task more challenging by commanding the drones to track a continuously changing yaw angle. This allows us to further probe their dynamic tracking ability in a more complex scenario.

\begin{figure}[t]
    \centering 
    \begin{subfigure}[b]{0.45\linewidth}
        \centering
        \includegraphics[width=0.9\linewidth]{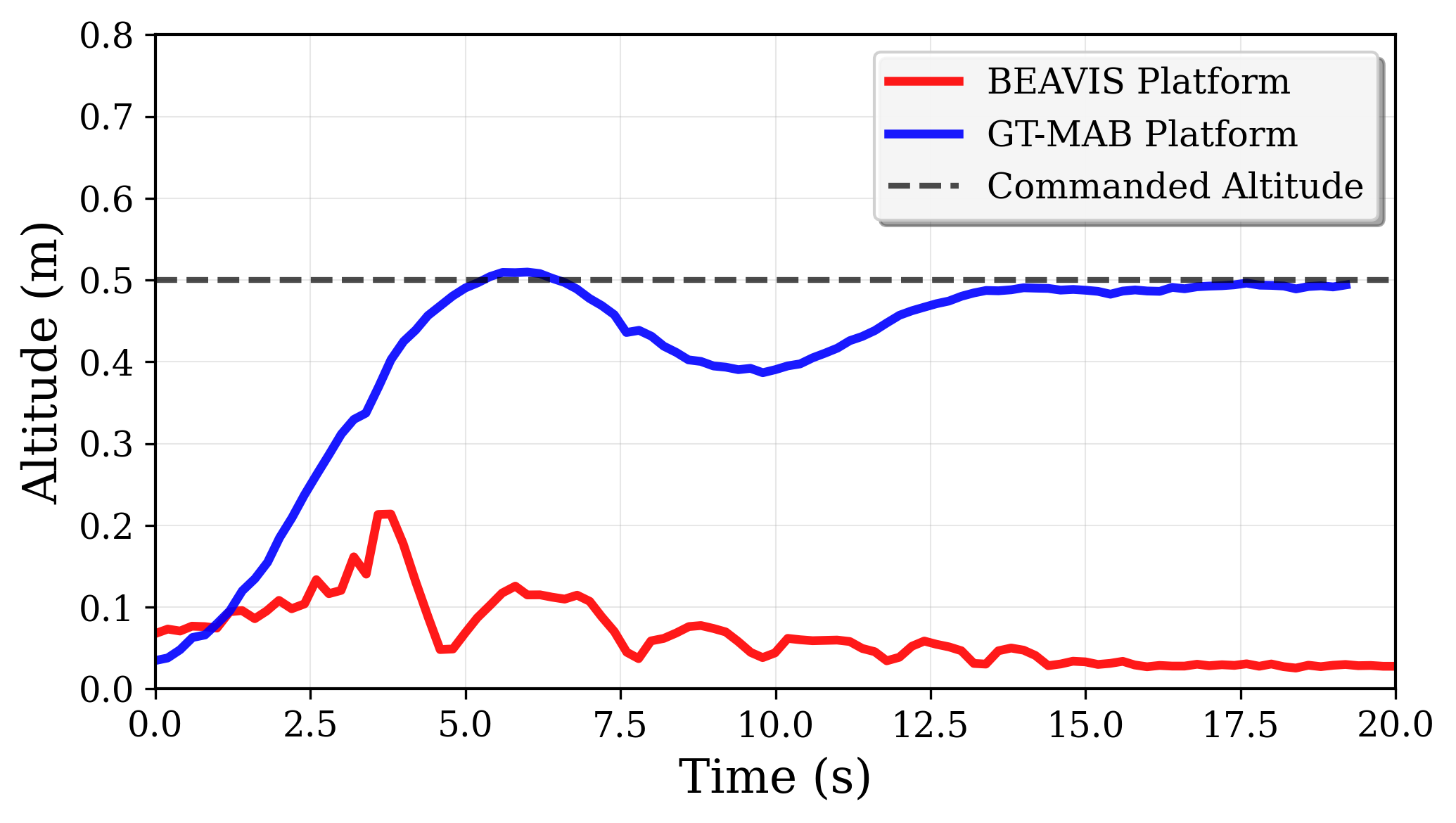}
        \Description{Altitude maintenance comparison image.}
        \caption{Altitude maintenance comparison.}
        \label{fig:altitude_compare}
    \end{subfigure}
    \hspace{5mm} 
    \begin{subfigure}[b]{0.45\linewidth}
        \centering
        \includegraphics[width=0.9\linewidth]{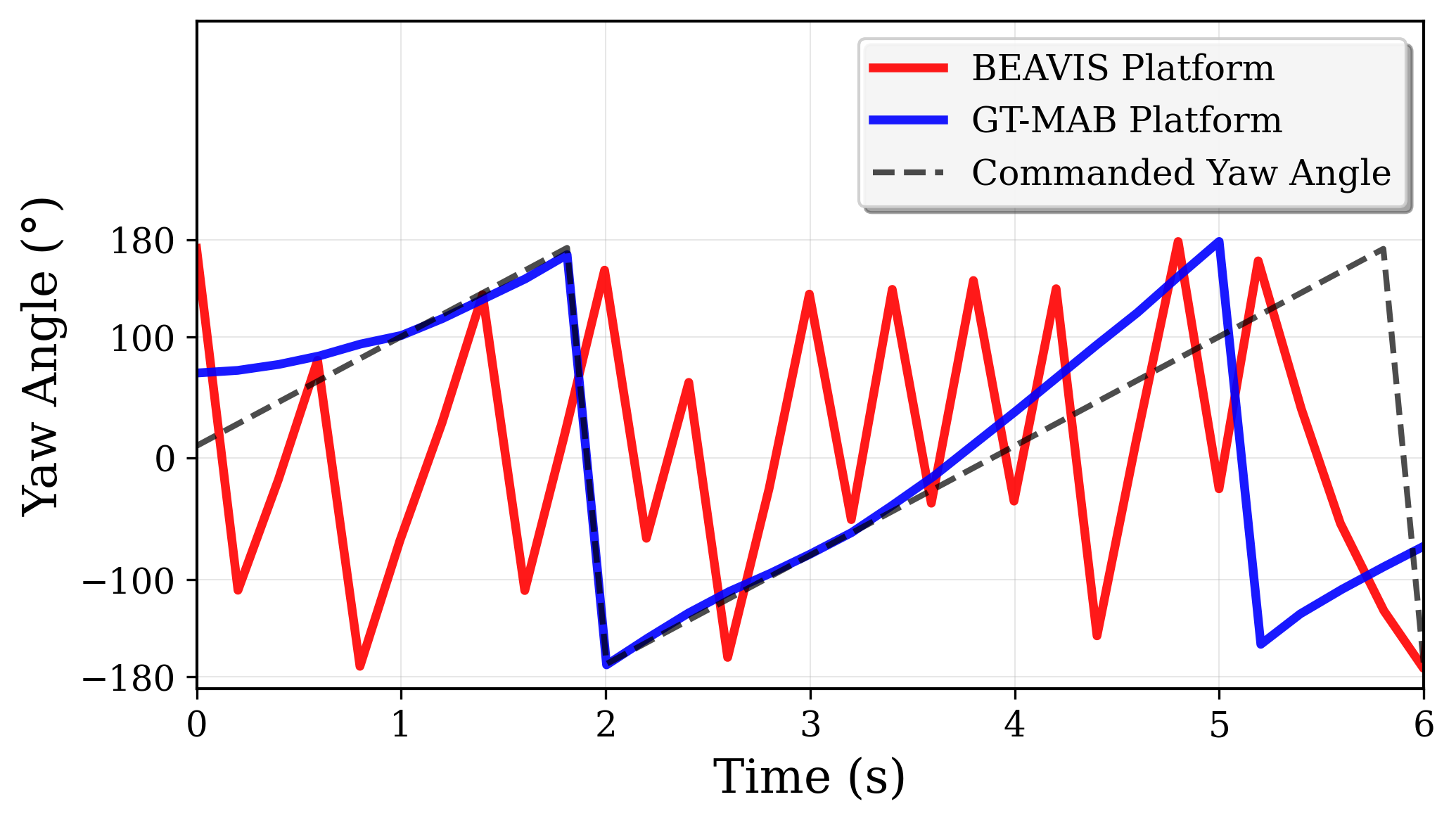}
        \caption{Yaw angle response comparison.}
        \label{fig:yaw_compare}
    \end{subfigure}
    
    \caption{Performance comparison of BEAVIS and GT-MAB in the  real world.}
    \label{fig:realworld}
\end{figure}

The real-world results, shown in \autoref{fig:realworld}, strongly corroborate our simulation findings. The GT-MAB platform (blue) effectively tracks both the altitude and the dynamic yaw commands with only minor delays or overshooting behavior. The BEAVIS platform (red), as predicted by our simulator, is visibly unstable, showing large altitude errors and an inability to control its heading.

\subsubsection{Discussion and Broader Implications}

Our three enhancements (propeller force, drag force, and controller) allows us to reliably identify the more suitable platform. Given its stability and agility in simulations and the real world, we select GT-MAB as the baseline platform to enhance with solar panels and visible light navigation.

While our analysis focuses on selecting a baseline platform, our enhanced closed-loop simulation framework provides a powerful tool for the broader design of LTA systems. As established earlier, physical prototyping faces significant limitations: slow iteration cycles, inconsistency from factors such as helium leakage, and the inability to rapidly test new design features —like embedding solar panels or new sensors for navigation. Our simulator overcomes these challenges. By combining a predictive CFD-based model with a validated closed-loop controller, our framework not only allows fair, reproducible comparisons between existing platforms but also fast and physics-based evaluation of new LTA-drone designs.

\section{Extending Endurance With Energy Harvesting}
\label{sec:energy-harvesting}

With the GT-MAB platform chosen as a baseline design (\autoref{sec:platform_comparison}), our focus now shifts to developing enhanced capabilities for applications like indoor monitoring.
Consider, for instance, a scenario where a drone performs periodic environmental checks in a large greenhouse. An LTA platform is particularly well-suited for such scenarios as it can hover efficiently at various locations while gathering sensor data over extended periods. However, making such long deployments truly autonomous requires addressing two key challenges. First, although buoyancy dramatically reduces the energy needed to hover, the overall endurance of the drone remains restricted by the small battery capacity. Frequent re-charging is needed due to the continuous power draw from electronics, sensors, and stabilization motors (especially as helium slowly leaks). Second, navigating reliably within these large indoor spaces, where satellite-based GPS signals are often unavailable, requires a simple and effective navigation method.

The LTA architecture provides a natural advantage in addressing both issues. The buoyant lift allows the platform to carry additional payload without drawing extra power from the rotors, freeing capacity to embed  solar panels to harvest energy and specialized sensors to assist with navigation.
This section tackles the first of these challenges: extending operational time through energy harvesting. We investigate the selection, mounting, and integration of solar panels; exploring how the large surface area of the LTA envelope can be leveraged to offset the continuous power draw of essential subsystems. The second challenge, autonomous navigation, is addressed in the following section.

\subsection{Selecting Solar Cells}

Integrating solar cells onto the GT-MAB’s curved envelope requires balancing three critical factors: weight, form factor (surface area), and energy-harvesting efficiency. Flexible (thin-film) cells are lightweight and can be placed easily over the balloon’s curvature, but they typically offer lower harvesting efficiency. In contrast, rigid (monocrystalline) panels provide higher efficiency but are heavier and more difficult to mount on curved surfaces.

\begin{table}[h!]
    \centering
    % \footnotesize % Resizebox handles sizing, so this isn't strictly needed
    \caption{Comparison of different solar cells.}
    \label{tab:solar_cells}
    % Resizebox forces the table to fit exactly within the text width
    \resizebox{\textwidth}{!}{%
        \begin{tabular}{@{}ccccccc@{}}
            \toprule
            \text{Manufacturer} & \text{Model} & \thead{\text{Dimensions}\\(mm)} & \thead{\text{Weight}\\(g)} & \thead{\text{Normalized Weight}\\(g/cm$^2$)} & \thead{\text{Charging Time}\\(s)} & \thead{\text{Normalized Charging Time}\\(s/cm$^2$)} \\
            \midrule
            Dracula     & Layer          & 63x67xN/A       & 1.66   & 0.039 & 646.2  & 15.3  \\
            Adafruit    & 5368           & 113x66x2.9      & 32.293 & 0.433 & 19     & 0.25  \\
            SEEED       & 31307002       & 137x81x1.5      & 50.417 & 0.454 & 14     & 0.126 \\
            PowerFilm   & MPT3.6-75      & 26.9x9.9x0.22   & 9      & 0.034 & 116.02 & 2.15  \\
            Kitronik    & 3605           & 110x110x3       & 60     & 0.495 & 12     & 0.099 \\
            ANYSOLAR    & SM811K08L      & 8.9x7.8x2.1     & 23.6   & 0.34  & 6.2    & 0.089 \\
            \bottomrule
        \end{tabular}%
    }
\end{table}

To navigate this trade-off, we evaluate several commercially solar panels, comparing their weight, active area, and charging performance (\autoref{tab:solar_cells}). In each experiment, a solar cell is connected to a depleted supercapacitor, and we measure the time required to charge it from 0\,V to 4\,V under no-load conditions. From these measurements, we derive two normalized metrics:

\begin{itemize}
    \item \textit{Weight per unit area (g/cm²)}, reflecting the weight added to the drone.
    \item \textit{Charging time per unit area (s/cm²)}, reflecting relative harvesting efficiency.
\end{itemize}

% \begin{figure}[t]
%   \centering
%   \includegraphics[width=0.35\linewidth]{./img/solar_panel_weight_comparison_20251028_170509.png}
%   \caption{Analysis of solar cells: Pareto Frontier}
%   \label{fig:TversusWeight}
% \end{figure}

\begin{figure}[tb!]
    \centering
    \begin{minipage}[t]{0.36\linewidth}
        \centering
        \includegraphics[width=\linewidth]{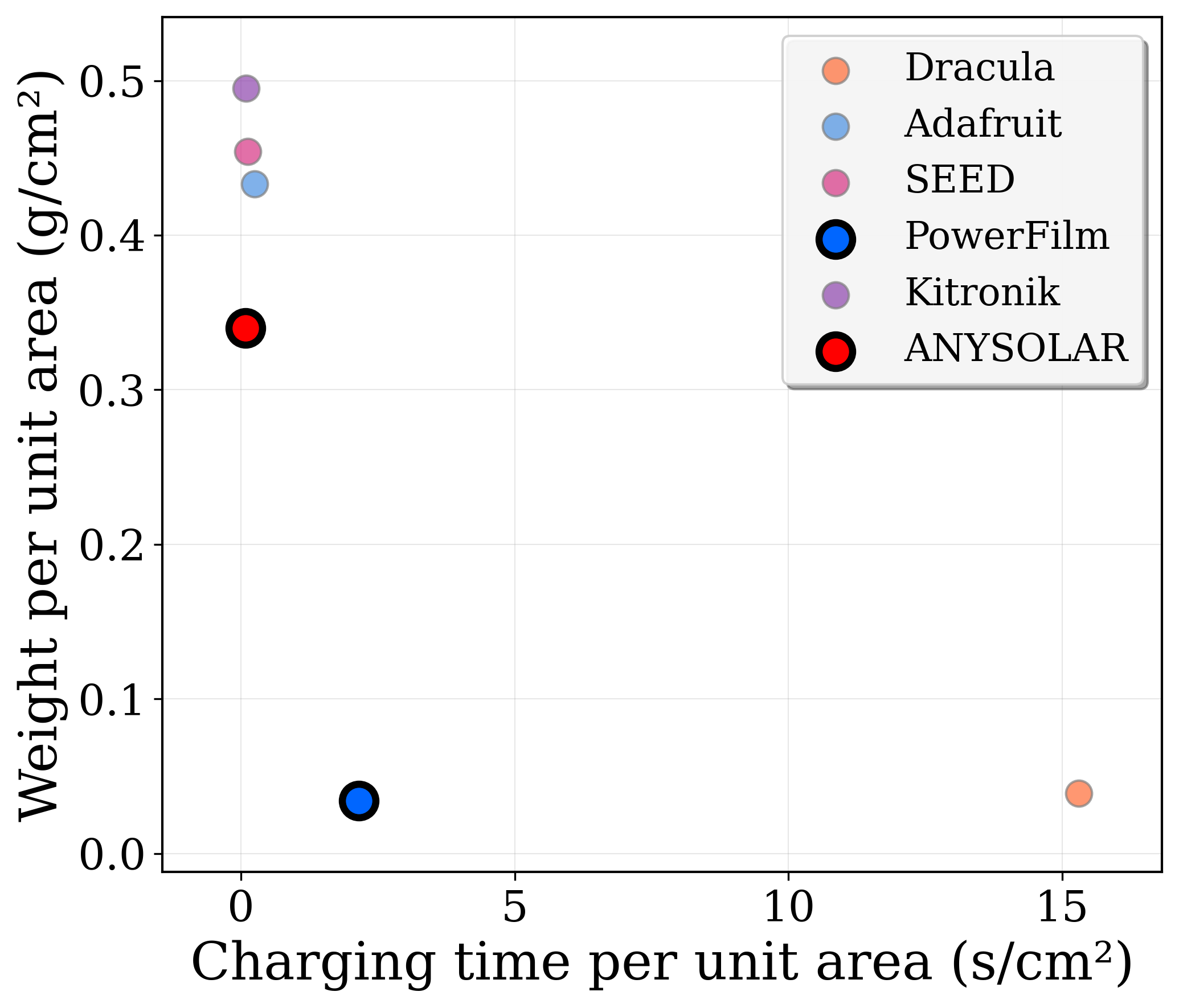}
        \captionof{figure}{Analysis of solar cells: Pareto Frontier}
        \label{fig:TversusWeight}
    \end{minipage}
    \hspace{5mm} 
    \begin{minipage}[t]{0.54\linewidth}
        \centering
        \includegraphics[width=\linewidth]{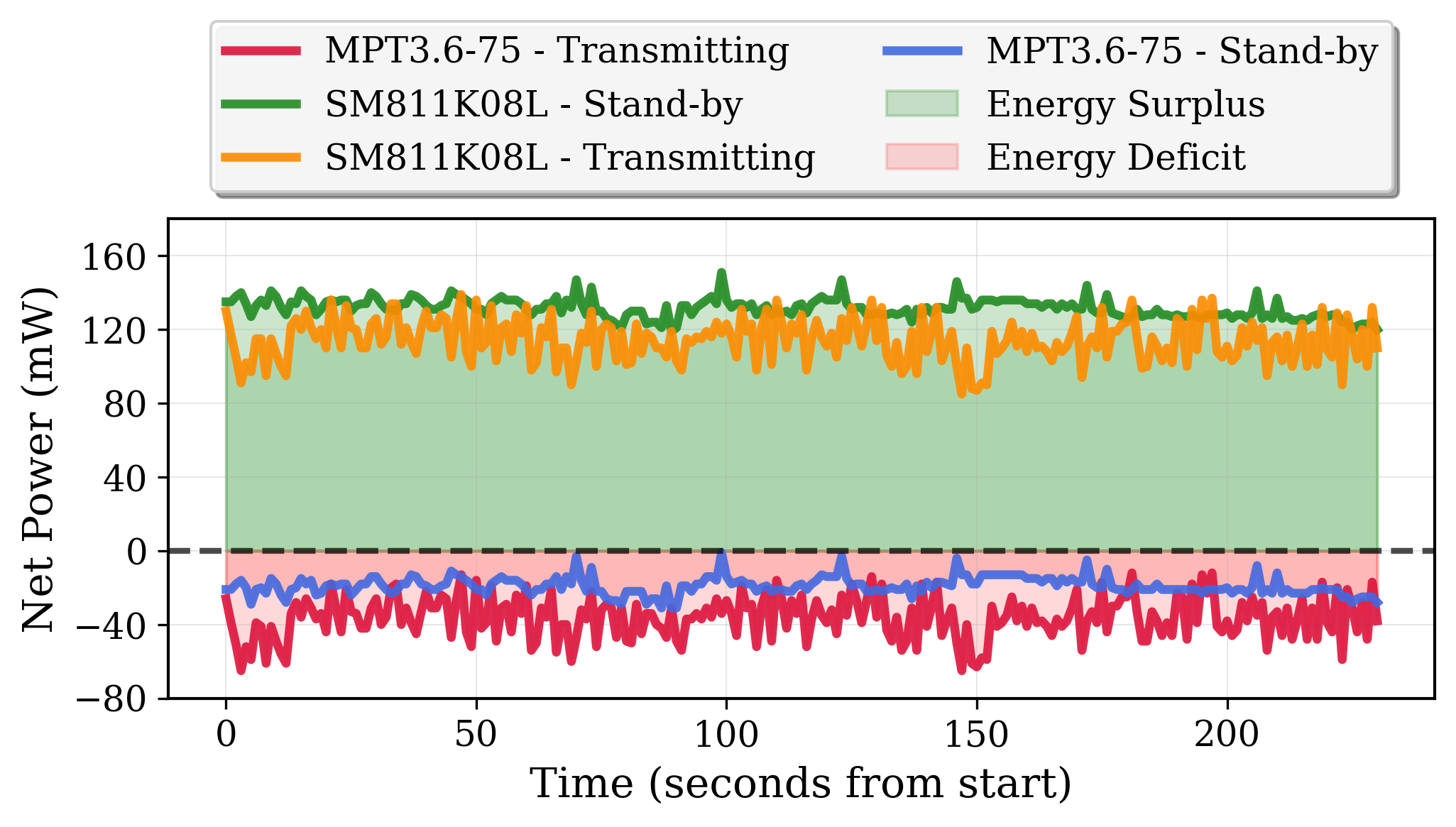}
        \caption{Net power harvested from PowerFilm and ANYSOLAR.}
        \label{fig:solarPower}
    \end{minipage}
\end{figure}

% \begin{figure}[tb!]
%   \centering
%   \includegraphics[width=0.5\linewidth]{./img/energy_balance_analysis_20251010_175820.png}
%   \caption{Net power analysis of the PowerFilm array versus the ANYSOLAR array.}
%   \Description{A plot comparing the net power output of the PowerFilm and ANYSOLAR solar arrays, showing the ANYSOLAR array producing a net energy surplus and the PowerFilm array a deficit.}
%   \label{fig:solarPower}
% \end{figure}

For a given surface area $A$, the ideal solar cell minimizes both weight and charging time. \autoref{fig:TversusWeight} plots these two metrics, revealing a clear Pareto frontier between lightweight and high-efficiency options. Two options emerge as optimal candidates: the PowerFilm (blue dot), a flexible thin-film module offering the lowest weight per area, and the ANYSOLAR (red dot), a rigid monocrystalline cell providing the shortest charging time per area.

To evaluate both solar cell candidates under realistic operating conditions, we construct solar cell arrays using each type while constraining the total weight to 46–49\,g. This weight matches the lifting capacity of our LTA platform. The PowerFilm array consists of four panels (2S2P configuration, 1065 mm² total area, 46.5\,g), while the ANYSOLAR array uses two panels connected in series (138.84 mm² total area, 48.8\,g). Despite their similar weights, the PowerFilm array covers a substantially larger surface area due to its much lower unit weight.

To measure the electrical power generated by each array, we use an LED source that provides a uniform illumination of 18,000\,lux. The power harvesting process is managed by a maximum power point tracking (MPPT) controller (SparkFun Sunny Buddy), while an INA219 current and voltage sensor records both the array’s output power and the power consumed by the drone’s main microcontroller (MCU). From these measurements, we compute the net power (generated power minus consumed). This analysis is critical because we need to make sure that while the micro-drone is floating, the solar cells can generate enough energy to power the MCU operations and to have a surplus that can charge the battery. Considering this goal, we evaluate two representative scenarios:
\begin{itemize}
    \item \textit{Stand-by} – where the MCU performs state estimation and the motors are off.
    \item \textit{Transmitting} – same as stand-by but adding the energy costs of wireless data transmissions.
\end{itemize}

The results, presented in \autoref{fig:solarPower}, plot net power (y-axis, in mW) over time (x-axis, in seconds). Positive values (green-shaded regions) indicate an energy surplus, while negative values (red-shaded regions) represent an energy deficit. The PowerFilm array (blue line for standby, red line for transmission) consistently operated with a net energy deficit: even in idle mode, its output was insufficient to sustain the MCU’s baseline power demand. In contrast, the ANYSOLAR array (green line for standby, orange line for transmission) produced a clear and sustained energy surplus in both operating conditions.

This surplus demonstrates that the ANYSOLAR array can successfully power the drone's core functionalities (processing and communication) while simultaneously charging the battery. Therefore, despite its rigid form factor, the ANYSOLAR SM811K08L array is selected. The next critical step is to determine how to mount this rigid array on the platform without compromising stability.

\subsection{Solar Cell Mounting}

\begin{figure}[tb!]
    \centering
    % --- First Subfigure ---
    \begin{subfigure}[b]{0.25\textwidth}
        \centering
        \includegraphics[width=\textwidth]{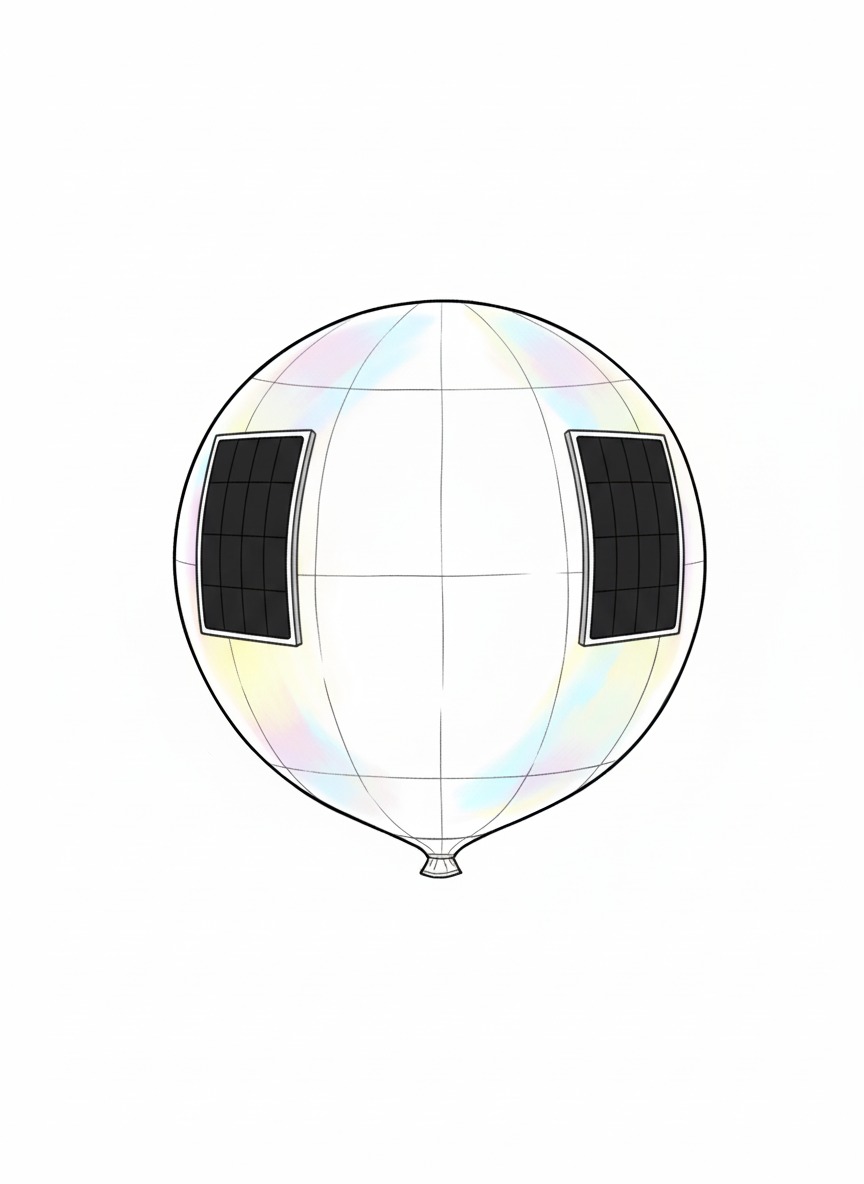}
        \caption{Sides}
        \label{fig:sidemount}
    \end{subfigure}
    \hfill % Adds space
    % --- Second Subfigure ---
    \begin{subfigure}[b]{0.25\textwidth}
        \centering
        \includegraphics[width=\textwidth]{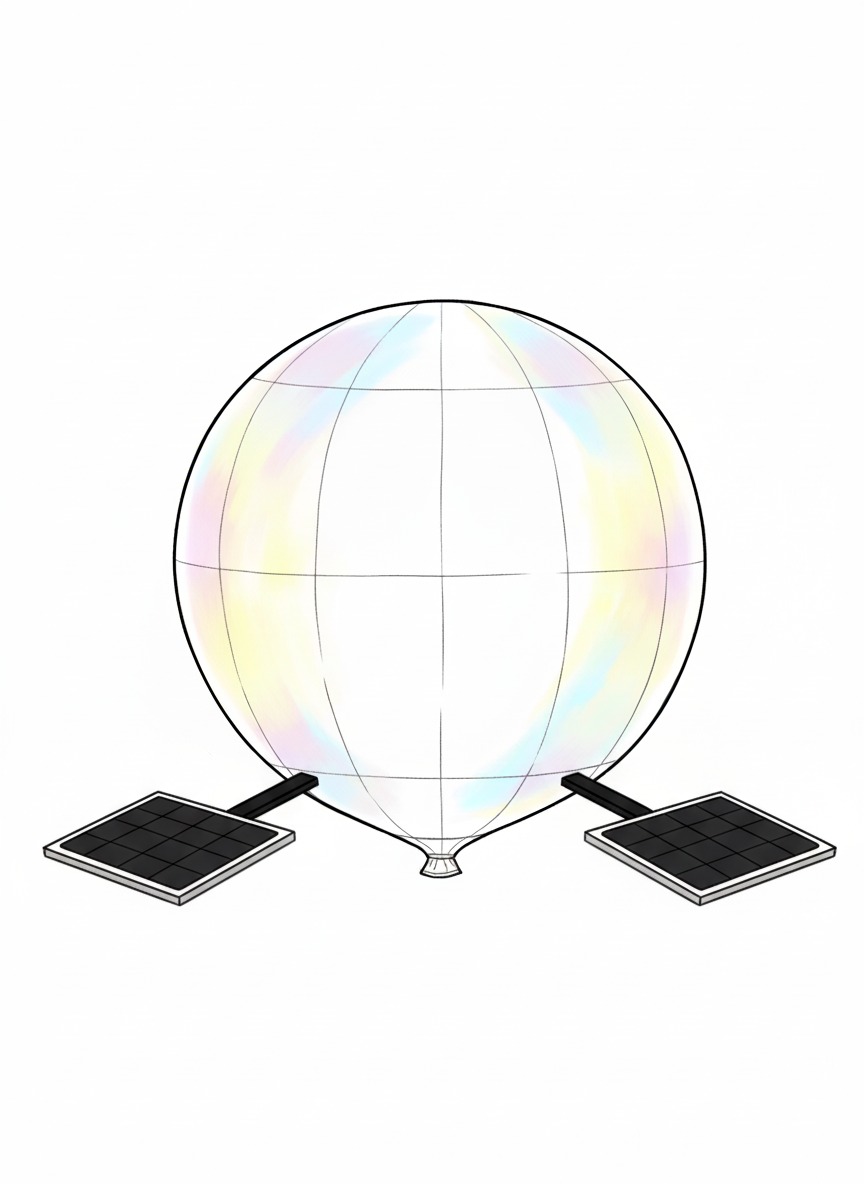}
        \caption{Bottom}
        \label{fig:bottomount}
    \end{subfigure}
    \hfill % Adds space
    % --- Third Subfigure ---
    \begin{subfigure}[b]{0.25\textwidth}
        \centering
        \includegraphics[width=\textwidth]{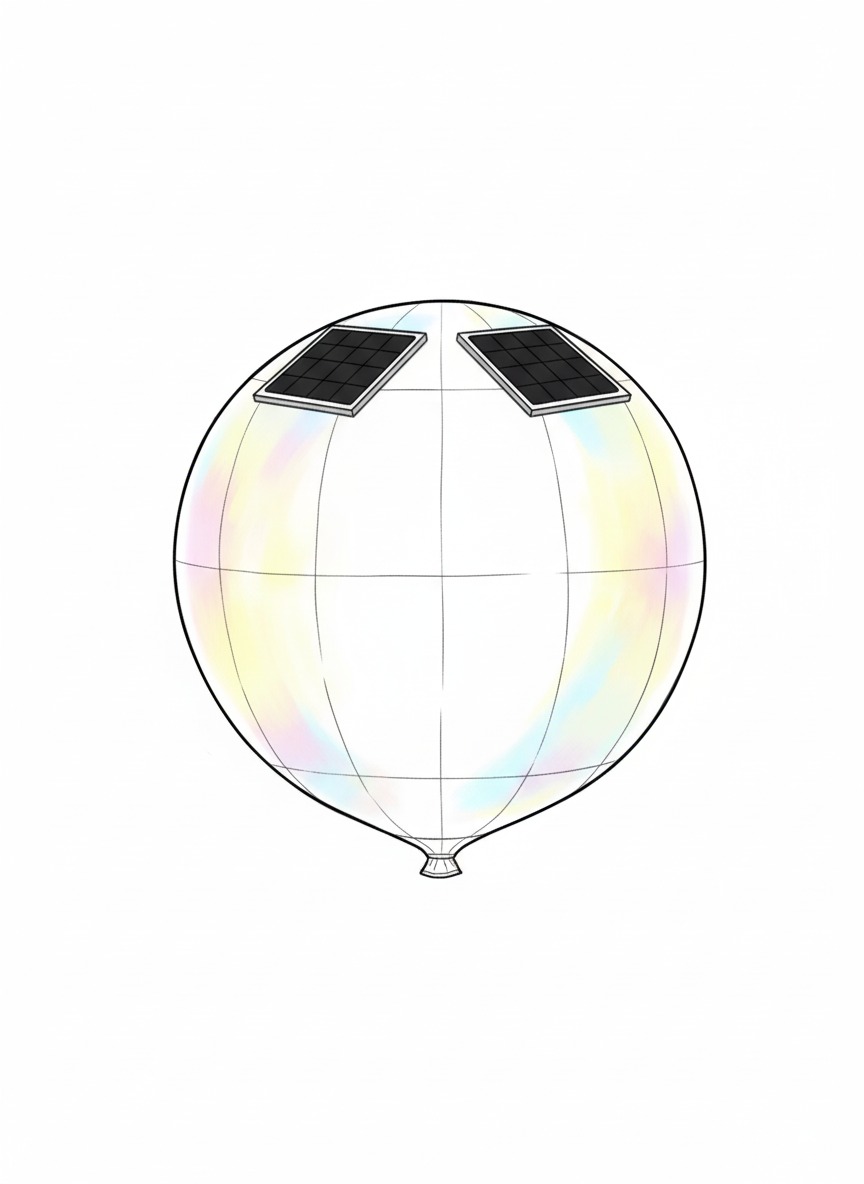} % <-- Add your 3rd image path
        \caption{Top}
        \label{fig:topmount}
    \end{subfigure}
    
    % --- Main Caption ---
    \caption{Different structural methods to mount solar panels}
    \label{fig:mountSolar}
\end{figure}

\begin{figure}[tb!]
    \centering
    % --- Bottom Row (Two Figures) ---
    \begin{subfigure}[t]{0.3\textwidth} % <-- Set width for a single centered figure
        \centering
        \includegraphics[width=\textwidth]{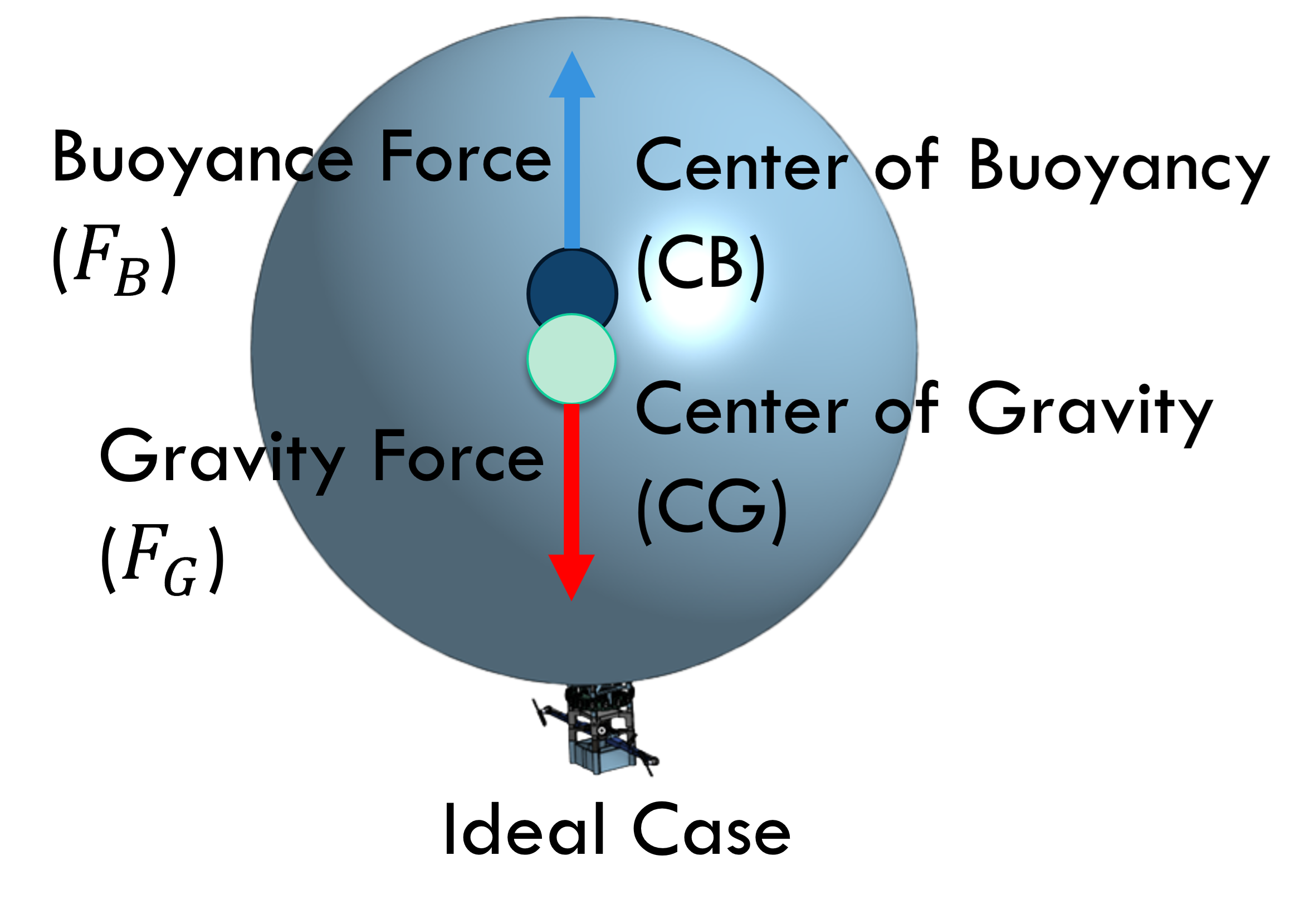}
        \caption{Case when CB and CG overlap}
        \label{fig:centerpoints}
    \end{subfigure}
    \hfill % Adds space between the two bottom figures
    \begin{subfigure}[t]{0.3\textwidth} % <-- Width for side-by-side figures
        \centering
        \includegraphics[width=0.6\textwidth]{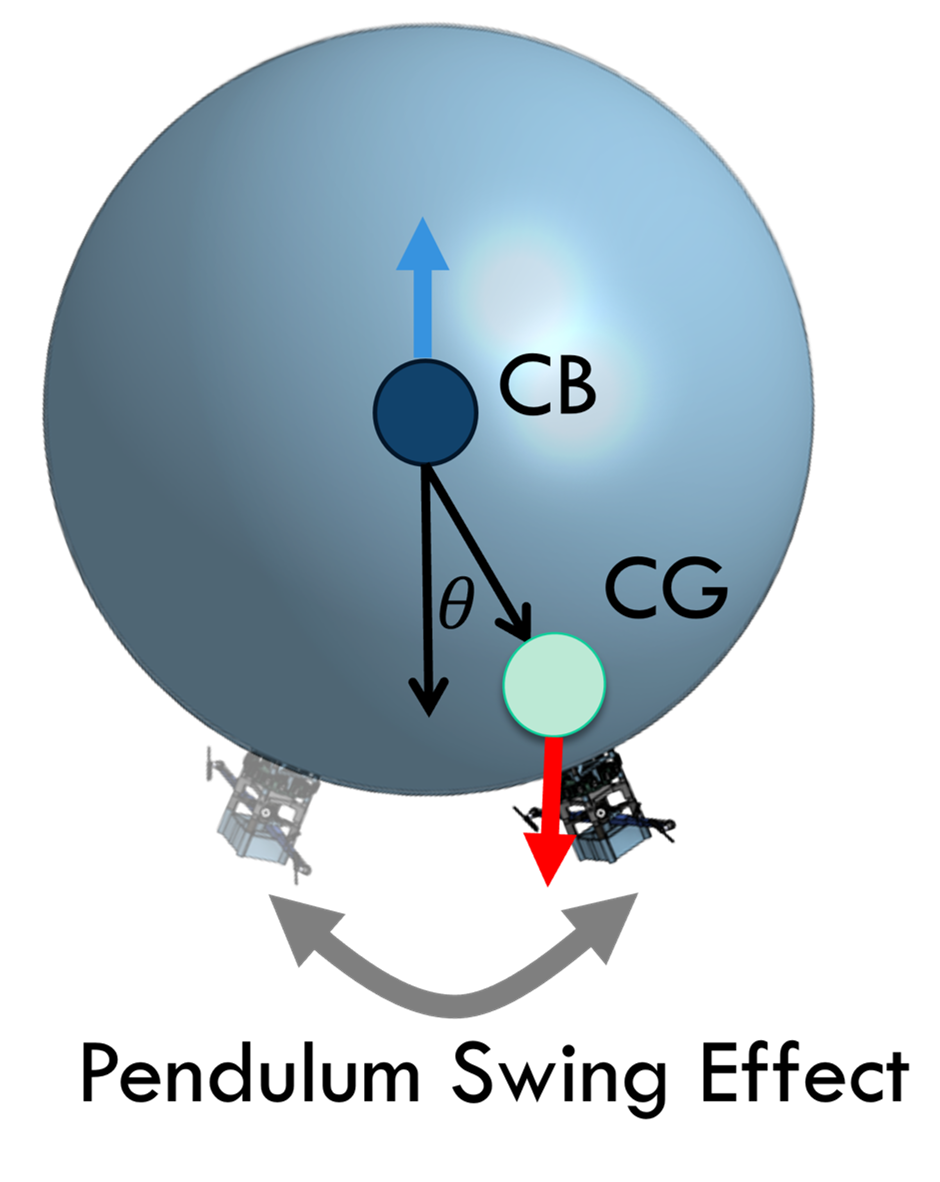}
        \caption{Case when CG is under CB}
        \label{fig:CGOscillate}
    \end{subfigure}
    \hfill % Adds space between the two bottom figures
    \begin{subfigure}[t]{0.3\textwidth} % <-- Width for side-by-side figures
        \centering
        \includegraphics[width=0.6\textwidth]{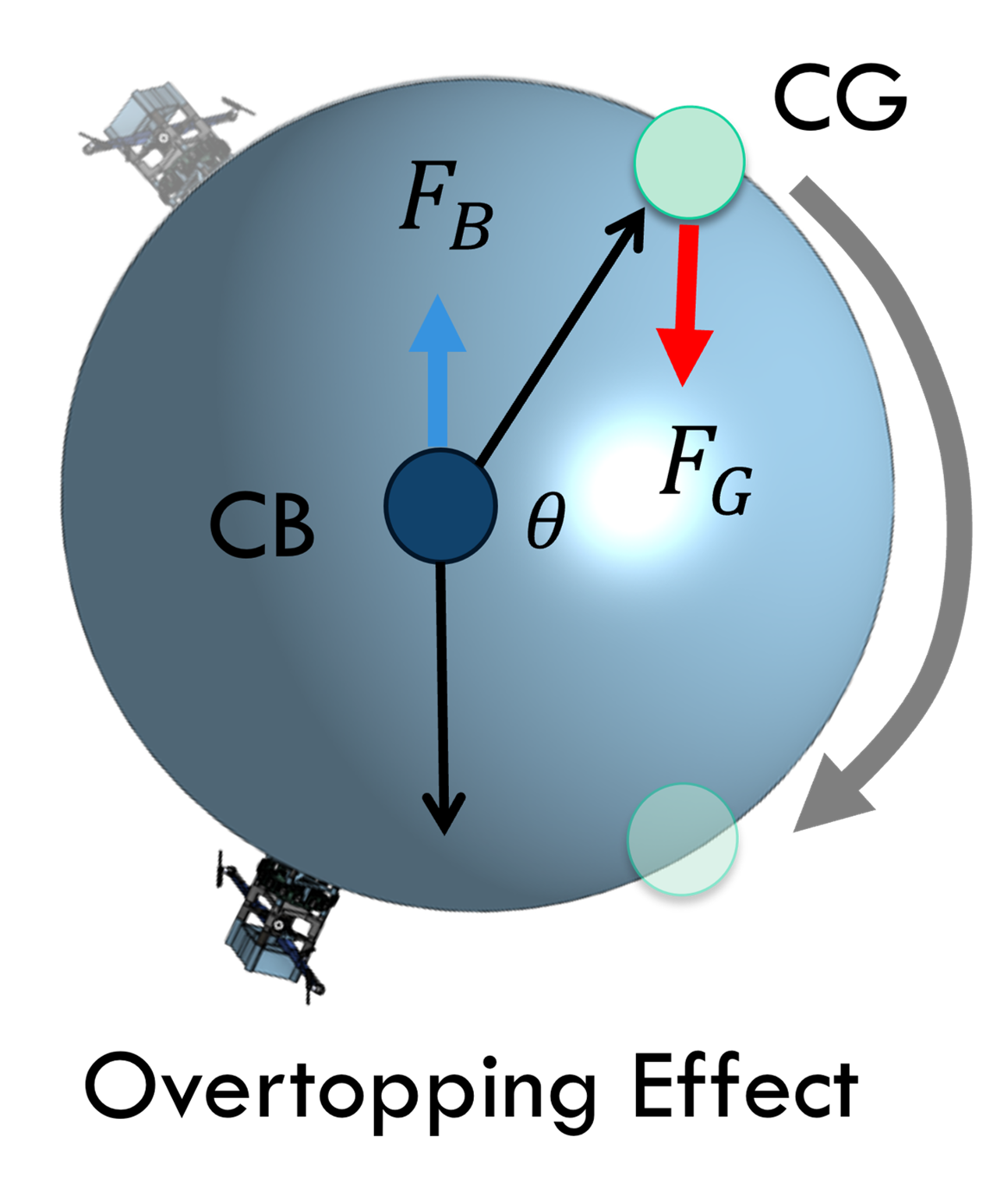}
        \caption{Case when CG is above CB}
        \label{fig:CGOMore}
    \end{subfigure}
    % --- Main Caption (Updated to match your images) ---
    \caption{Analysis of the system's stability: (a) Shows the Center of Buoyancy and Center of Gravity, (b) Illustrates standard oscillation, and (c) Shows overtopping effects.}
    \label{fig:cg_stability_analysis} % <-- Updated label
\end{figure}

Mounting the rigid ANYSOLAR array presents an important trade-off: the system must maximize light exposure while maintaining flight stability. The solar panels can be mounted in multiple ways, on the side, at the bottom or at the top of the platform, as shown in \autoref{fig:mountSolar}. Each configuration, however, affects the platform’s stability differently and must therefore be analyzed carefully.

The stability of an LTA platform is governed by the relative positions of two key points: the \textit{Center of Buoyancy} (CB), the point where the balloon's lift is centered, and the \textit{Center of Gravity} (CG), the point with the combined weight of all components. Ideally, these two centers should coincide to maximize stability (\autoref{fig:centerpoints}). In practice, however, most designs exhibit some offset. When the CG lies \textit{below} the CB, the platform behaves like a pendulum--it may swing slightly during motion but will naturally return to equilibrium (\autoref{fig:bottomount}). However, if the CG lies \textit{above} the CB, the platform becomes unstable and may flip over (\autoref{fig:topmount}). Thus, while neither condition is perfect, maintaining the CG below the CB is generally safer and more controllable.

To assess how solar panel placement affects these dynamics, we evaluate the three mounting options of \autoref{fig:mountSolar} in our simulator. In each configuration, the platform is instructed to move in a straight line at 0.1m/s, allowing us to measure its pitch stability and oscillation behavior.

\begin{figure}[tb!]
    \centering
    % --- Bottom Row (Two Figures) ---
    \begin{subfigure}[t]{0.32\textwidth} % <-- Set width for a single centered figure
        \centering
        \includegraphics[width=\textwidth]{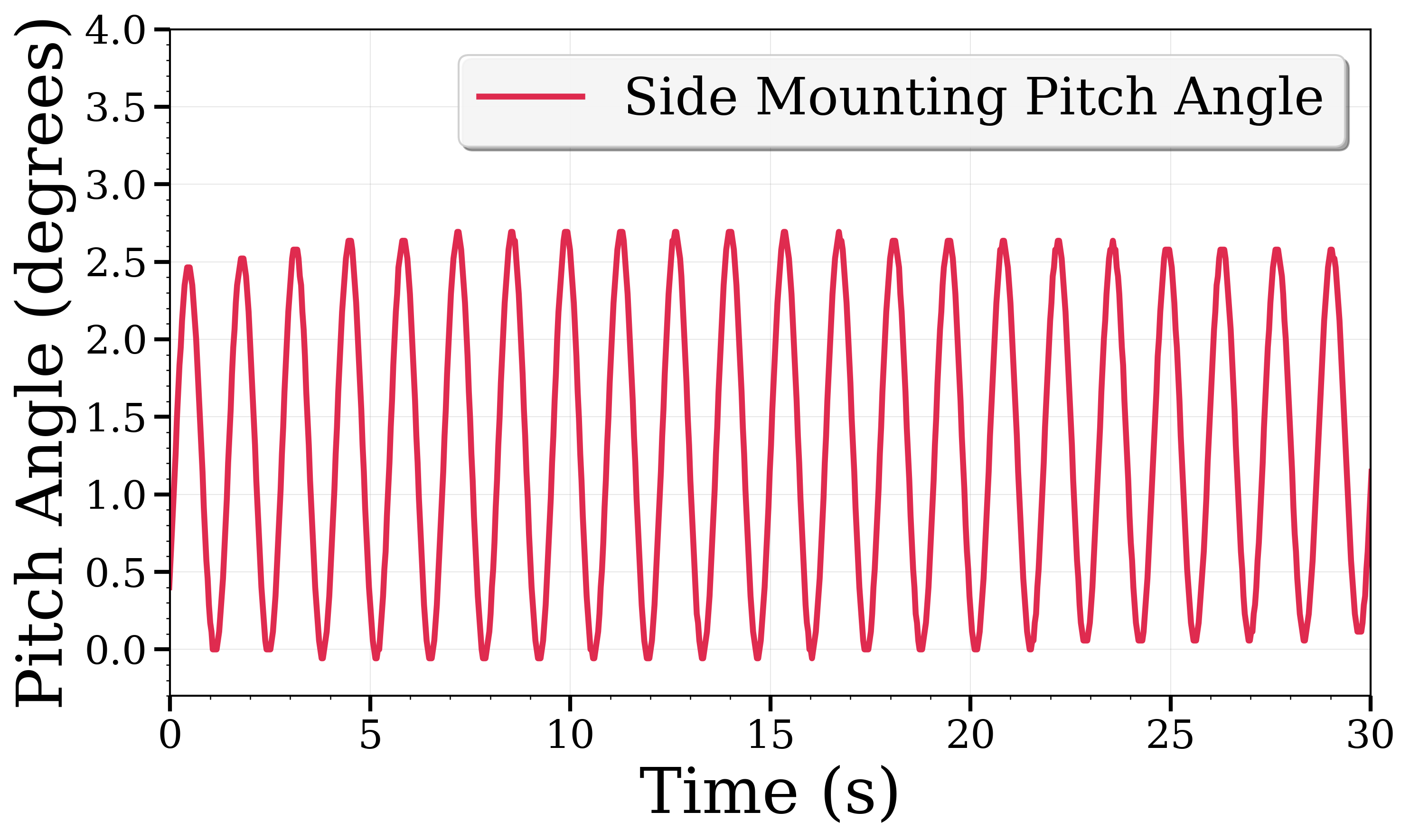}
        \caption{Sides}
        \label{fig:lateralMount}
    \end{subfigure}
    \hfill % Adds space between the two bottom figures
    \begin{subfigure}[t]{0.32\textwidth} % <-- Width for side-by-side figures
        \centering
        \includegraphics[width=\textwidth]{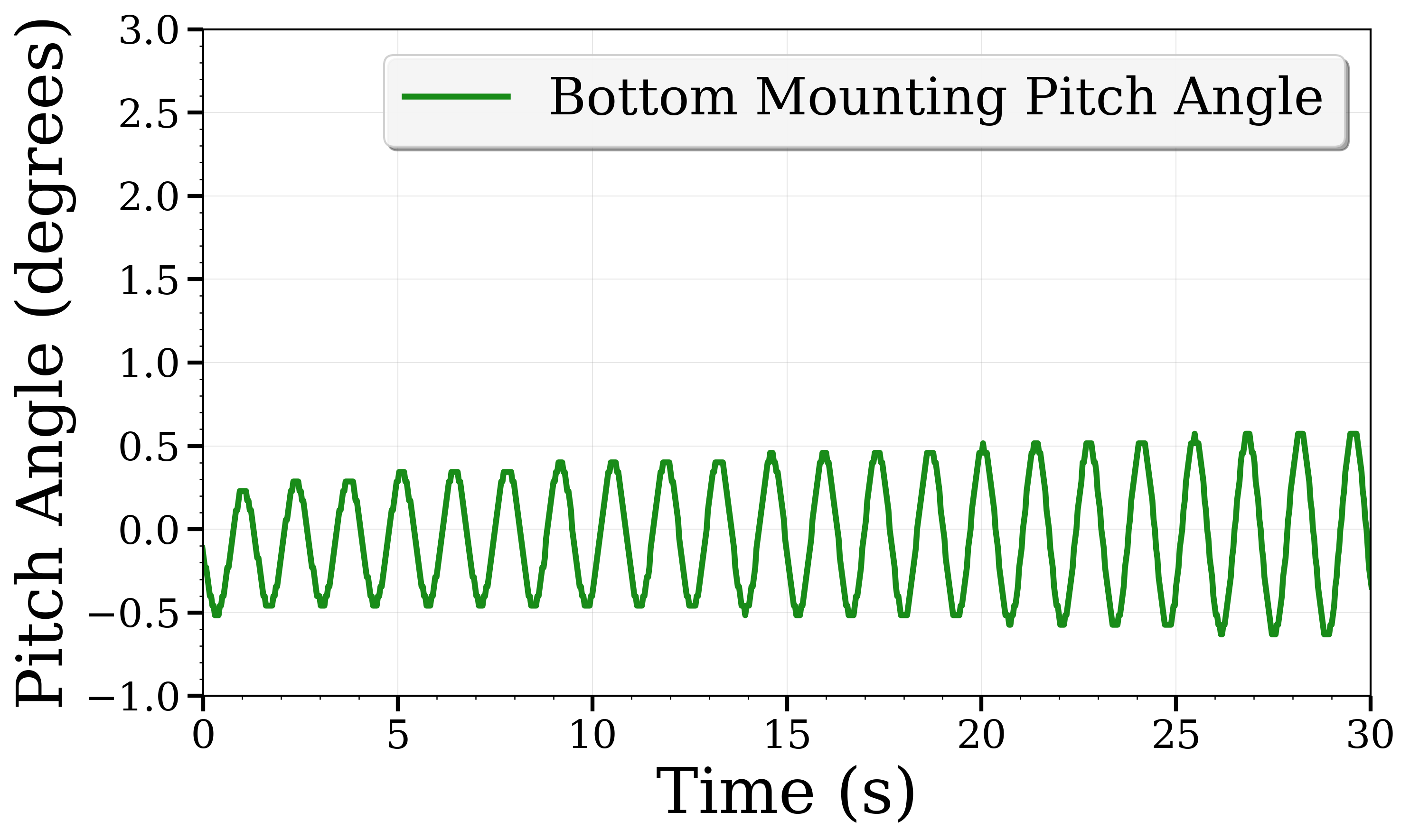}
        \caption{Bottom}
        \label{fig:bottomMount}
    \end{subfigure}
    \hfill % Adds space between the two bottom figures
    \begin{subfigure}[t]{0.32\textwidth} % <-- Width for side-by-side figures
        \centering
        \includegraphics[width=\textwidth]{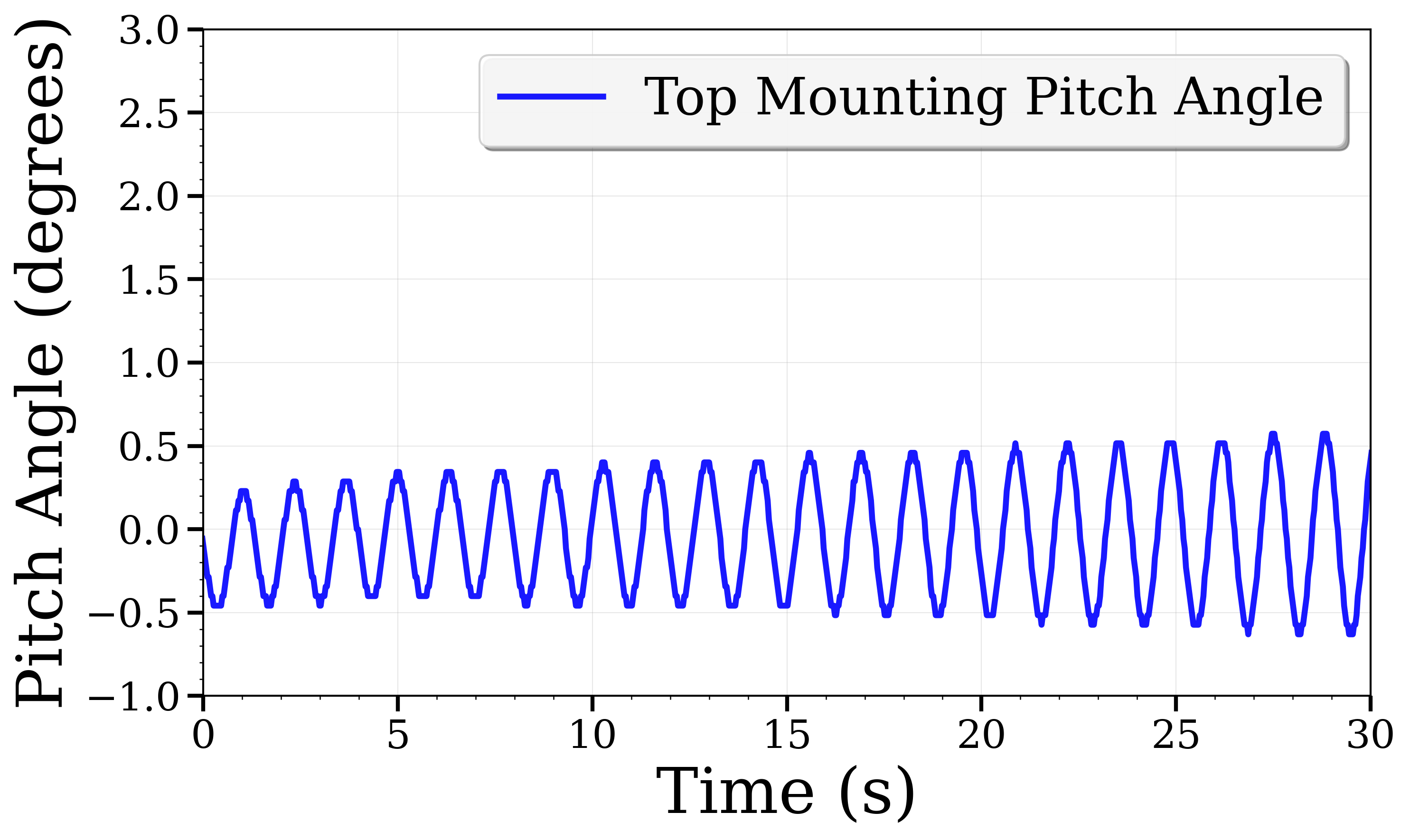}
        \caption{Top}
        \label{fig:TopMount}
    \end{subfigure}
    % --- Main Caption (Updated to match your images) ---
    \caption{Drone oscillation effect for different mounting strategies of the solar panels.}
    \label{fig:oscillation} % <-- Updated label
\end{figure}

\begin{enumerate}
    \item \textit{Lateral (side) mounting}. As shown in Figure 13a, the panels could be placed around the envelope. However, if the solar cells are not placed carefully, there will be an asymmetric weight that can shift the CG horizontally, away from the CB’s vertical axis. That horizontal difference between the centers would create a constant torque that induces tilt, causing strong oscillations during movements. The simulation results confirm that this configuration compromises stability (\autoref{fig:lateralMount}), exhibiting strong pitch oscillations around $\pm 2.5^ {\circ}$. Therefore, this option is discarded.

    \item \textit{Bottom mounting}. Mounting the panels on an extended frame at the bottom of the platform (\autoref{fig:bottomount}) keeps the CG low and vertically aligned with the CB, yielding better stability. The simulation results (\autoref{fig:bottomMount}) confirm that this configuration reduces oscillation amplitude. However, to prevent the balloon’s shadow from blocking the cells, the frame must extend significantly outward (40\,cm in our test), adding 45.5\,g of weight. Given the platform’s payload limitations, this added mass is not desirable.

    \item \textit{Top mounting}. The third configuration places the solar array on top of the envelope (\autoref{fig:topmount}). This option is the most weight-efficient, requiring no additional framing while maintaining vertical alignment between the CG and CB. A minor drawback of this placement is that it raises the overall CG slightly, reducing the pendulum’s restorative force. This means that a sudden forward acceleration can create drag, causing some tilt. However, our simulation shows that our smooth, cascaded PID controller minimizes the effect of rapid accelerations, as shown in \autoref{fig:topmount}. The top-mounted configuration achieves stability comparable to the bottom-mounted setup but without the added weight.
\end{enumerate}

Based on these findings, we select the top-mounted configuration as the optimal solution. It provides the best balance between low weight, stable center-of-gravity alignment, and adequate light exposure, making it the most practical and efficient choice for indoor energy-harvesting operations.

\subsection{Integration of Solar Cells with the Drone Platform}

\begin{figure}[tb!]
  \centering
  \includegraphics[width=\linewidth]{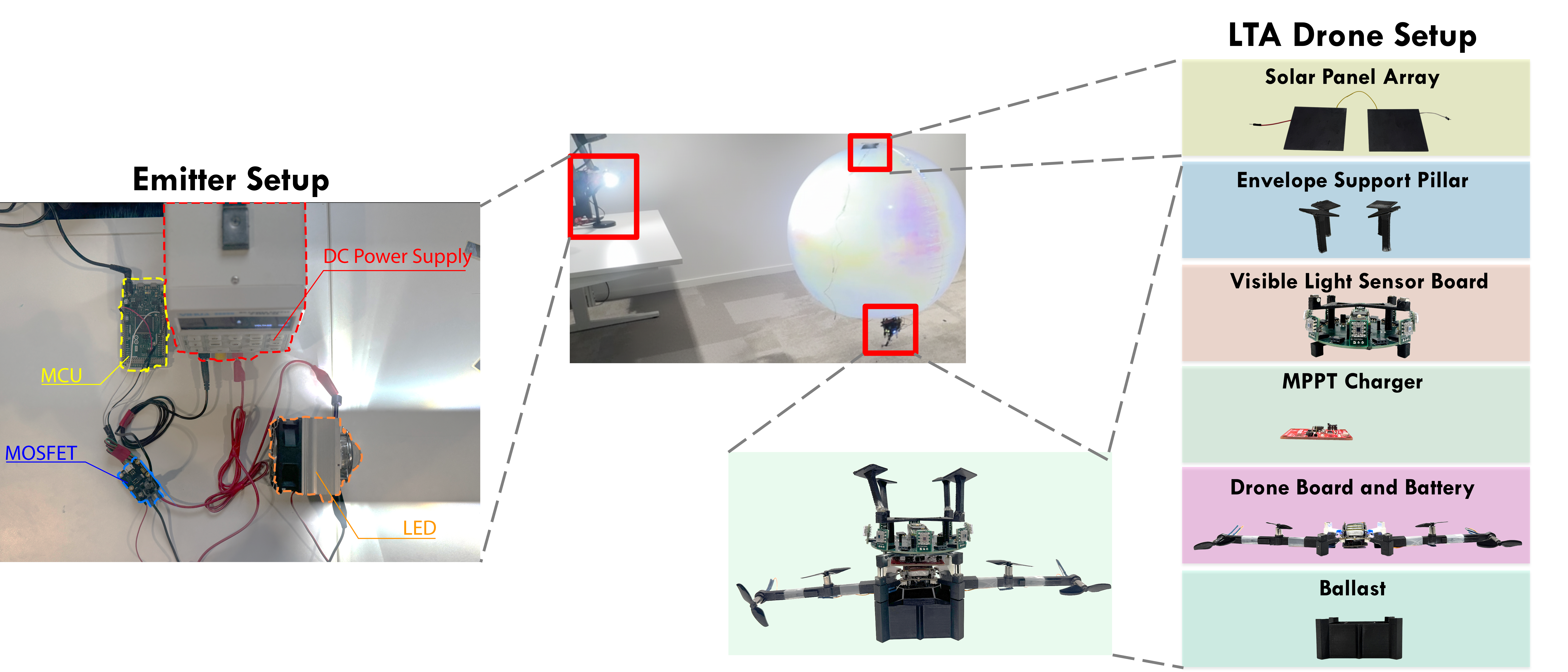}
  \caption{Hardware Setup. The emitter consists of a single  light source with a simple MOSFET to modulate the frequency (explained in the next section). The LTA platform hosts the solar cells on top and the main frame at the bottom has multiple layers: Ballast (for weight control), Rotors and drone control, MPPT charger (for energy harvesting), Board with photosensors (for light-based navigation), and support pillars to connect the main frame with the envelope.}
  \label{fig:exploded_view}
\end{figure}

To integrate the envelope and solar cells, we construct a GT-MAB–inspired LTA drone (\autoref{fig:exploded_view}). Rather than relying on a single rigid 3D frame, we adopt a modular architecture that allows individual layers to be stacked and reconfigured. Each layer is connected using pairs of N50 neodymium magnets (1.2 kg adhesive force), providing a robust yet easily detachable assembly that simplifies replication, maintenance, and design iteration.

The platform consists of four functional layers:
\begin{enumerate}
    \item \textit{Base layer — Ballast module (bottom)}:
This layer houses adjustable weights that allow fine-tuning of the system’s net buoyancy.
    \item \textit{Propulsion and control layer}:
This layer contains the drone's main frame, including the four rotors, and an STM32F405-based flight controller responsible for stabilization and thrust control.
    \item \textit{Solar energy harvesting layer}:
This layer hosts the SparkFun Sunny Buddy charge controller, selected for its maximum power point tracking (MPPT) algorithm to optimize energy harvesting from the solar array.
    \item \textit{Navigation Sensor layer (top)}:
The uppermost layer holds a custom sensor board equipped with photodiodes and a dedicated Teensy MCU, used for the navigation task described in the next section.
\end{enumerate}

To preserve privacy, the drone does not include any camera sensors. Instead, it estimates its orientation (attitude) using the onboard inertial measurement unit (IMU). Planar velocity is derived by integrating accelerometer data over time, while altitude is estimated by fusing data from a barometer and a downward-facing Time-of-Flight (ToF) sensor.
All structural components—including the frame, pillars, and ballast container—are 3D printed in PLA with 15\% infill, minimizing weight while maintaining rigidity. A summary of key hardware specifications is provided in \autoref{tab:specs}.

\begin{table}[t]
  \centering
  \footnotesize
  \caption{Key Specifications of the (Name of our platform)}
  \label{tab:specs}
  \begin{tabularx}{\linewidth}{l X c c}
    \toprule
    \textbf{Component} & \textbf{Overview} & \textbf{Quantity} & \textbf{Weight (g)} \\
    \midrule
    Envelope & 100 cm diameter metallized foil balloon & 1 & 116.8 \\
    Envelope Support Pillar & 3D-printed structural support & 4 & 9.6 \\
    VLC Control Board & Custom board with Teensy 4.1 MCU & 1 & 61.1 \\
    Visible Light Sensor & OPT101 with custom board & 8 & 11.6 \\
    Solar Charger & SparkFun Sunny Buddy MPPT charger & 1 & 11.6 \\
    Propulsion & 7mm brushed coreless DC motor with 47mm propeller & 4 & 11.6 \\
    Propeller Frame & Motor mount and support frame & 1 & 33.3 \\
    Flight Controller & PCB with STM32F405, BMI088, \& nRF51822 & 1 & 6.8 \\
    ToF Sensor & Time of Flight board with VL53L1x & 1 & 1.6 \\
    Battery & 3.7V, 250mAh Lithium Polymer (LiPo) & 1 & 7.0 \\
    \midrule
    \multicolumn{3}{r}{\textbf{Total Weight (Platform Weight)}} & \textbf{282.5} \\
    \midrule
    \addlinespace % Adds a little vertical space for separation
    Ballast Case & Housing for adjustable weights & 1 & 40.7 \\
    % \midrule
    % \multicolumn{3}{r}{\textbf{Total Weight (with Ballast)}} & \textbf{40.7} \\
    \bottomrule
  \end{tabularx}
\end{table}

\section{Enabling Simple Navigation with Light}
\label{sec:light-navigation}

With energy harvesting addressed, we now tackle the second key challenge: enabling simple, reliable indoor navigation with light.
While sophisticated light-based positioning systems can achieve high precision, they require using multiple ceiling beacons. Our objective is not to provide precise cartesian coordinates; instead, we propose a simple “point-and-go” navigation that depends on a single light source. In this way, the light fixtures in a warehouse or greenhouse  can be used as "light houses'' that guide drones. The path of a drone can be determined by the sequence of light fixtures that it needs to follow. 

The transmitter can be placed anywhere—on a wall, stand, or ceiling—and does not require any major modification, only the ability to change its frequency, which is a feature that many modern lighting systems already have to control the radiated illumination.
This design choice, however, introduces a fundamental trade-off: by simplifying the transmitter infrastructure, we shift the navigational complexity onto the LTA platform (receiver). A single light source provides significantly less directional information than a multi-beacon system that allows for triangulation. The drone must therefore employ different onboard strategies to interpret this limited information and determine the correct heading. 

\subsection{Light-Based Navigation Strategies}

To systematically analyze the trade-offs identified above, we develop and evaluate three different algorithms: %. These methods represent different points in the design space:

\begin{enumerate}
    \item \textit{Bearing Angle Guidance (BAG)}: Uses a photodiode array to directly compute the light’s direction. The sensors with higher intensity indicate the source's bearing. This method is based on a prior study that uses an array of radio sensors to localize drones~\cite{6580223}.\\
    \textit{Pros:} Fast, accurate, and responsive.\\
    \textit{Cons:} Requires several photodiodes, increasing weight and hardware complexity.
    %BAG uses a photodiode array for direct angle calculation. This is the most direct method, but it requires multiple photodiodes.
    \item \textit{Dither Extremum Seeking (DES)}: Uses a single photodiode and small “wiggling” maneuvers to find the direction of maximum light intensity. This method is based on a prior study that uses single light beacons with a normal drone~\cite{wu2020inflightrangeoptimizationmulticopters}.\\
    \textit{Pros:} Lightweight; minimal sensor hardware.\\
    \textit{Cons:} Indirect and inefficient flight paths due to continuous oscillation.
    %DES uses only a single photodiode and active ``wiggling'' movements to find the light gradient. This minimizes the sensor payload but it could be inefficient in finding the path to approach the transmitter.
    \item \textit{Dither-Free Gradient Ascent (DGA)}: Also uses a single photodiode, but estimates the gradient from past movements. This is a novel method we propose to investigate  single-source navigation with light.\\
    \textit{Pros:} Lightweight; minimal sensor hardware.\\
    \textit{Cons:} Requires prior position estimates, sensitive to IMU drift.
\end{enumerate}

By developing and comparing these different approaches, we aim to determine which strategy offers the best balance of sensor requirements, computational complexity, and guidance efficiency for our LTA platform.

\subsubsection{Bearing Angle Guidance (BAG)}
\label{BearingANGLE}

Our first approach is a direct measurement strategy. The high-level idea is to build a ``compass-like'' sensor that instantaneously computes the light's Angle-of-Arrival (AoA), or bearing angle ($\theta_{bearing}$). This angle is then fed directly to the flight controller as the desired heading command ($\psi_{d}$). This approach simplifies the control logic, as the drone only needs to align its heading angle $\psi_{d}$ with the estimated bearing $\theta_{bearing}$ while maintaining a constant forward velocity.

This method is implemented using a circular array of $N$ photodiodes.
Each photodiode is oriented at a known direction, e.g., $0^\circ, 45^\circ, 90^\circ, \ldots 315^\circ$. The direction vector $\boldsymbol{n}_{i}$ of each diode $i$ is represented as: 

\begin{equation}
\boldsymbol{n}_{i} =
\begin{bmatrix}
\cos(\theta_i) \\
\sin(\theta_i)
\end{bmatrix}
\quad \text{where} \quad
\theta_i = i \cdot \frac{2\pi}{N} \quad \text{for } i = 0, 1, ..., N-1
\end{equation}

To find the direction of the transmitting light, a weighted average of all sensor vectors is calculated. Each vector $\boldsymbol{n}_{i}$ is weighted by its corresponding signal strength $M'_{peak,i}$ (obtained via FFT as described later). The weighted average produces a single, consolidated bearing vector $\boldsymbol{v}_{bearing}$:

\begin{equation}
\boldsymbol{v}_{bearing} = \frac{\sum_{i=0}^{N-1} M'_{peak,i} \cdot \boldsymbol{n}_{i}}{\sum_{i=0}^{N-1} M'_{peak,i}}
\label{eq:bearing_vector}
\end{equation}

Then, the drone's bearing $\theta_{bearing}$ is updated with the new vector information $\boldsymbol{v}_{bearing}$ using the $atan2$ function:

\begin{equation}
\theta_{bearing} = atan2(\boldsymbol{v}_{bearing, y}, \boldsymbol{v}_{bearing, x})
\end{equation}

%Aligning the drone with the light source while maintaining a constant forward velocity.

% \begin{figure}[tb!]
%   \centering
%   \includegraphics[width=0.5\linewidth]{./img/esalgorithm.drawio.png}
%   \caption{Dither Extremum Seeking Algorithm.}
%   \label{fig:ditherES}
% \end{figure}

\begin{figure}[tb!]
    \centering
    \begin{minipage}[t]{0.5\linewidth}
        \centering
        \includegraphics[width=\linewidth]{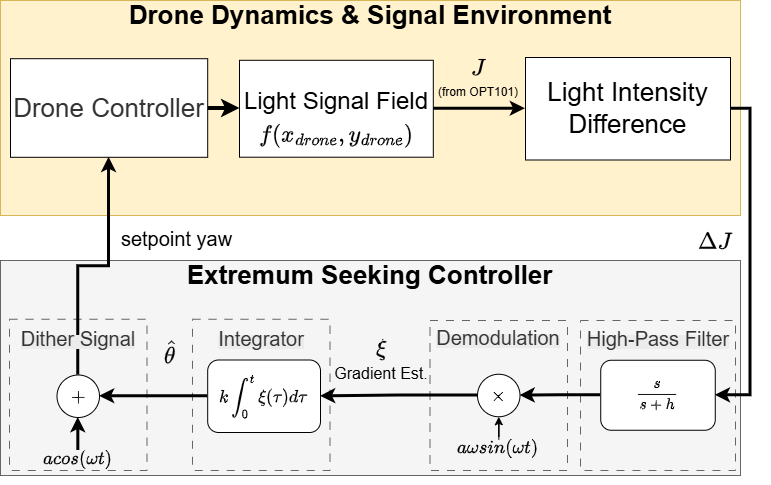}
        \captionof{figure}{Dither Extremum Seeking Algorithm.}
        \label{fig:ditherES}
    \end{minipage}
    \hspace{5mm} 
    \begin{minipage}[t]{0.4\linewidth}
        \centering
        \includegraphics[width=\linewidth]{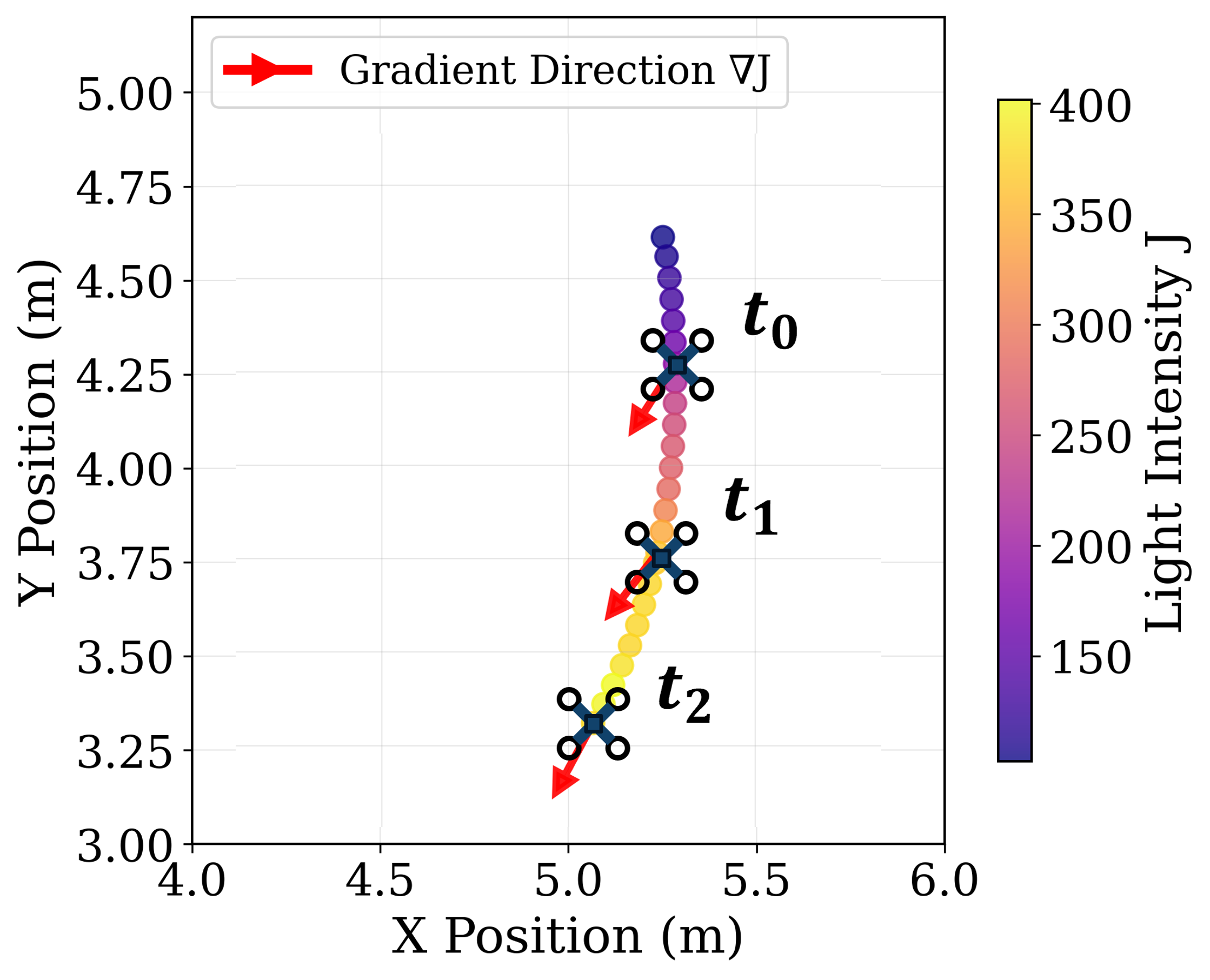}
        \caption{Dither-Free Gradient Ascent.}
        \label{fig:gradientFree}
    \end{minipage}
\end{figure}

% \begin{figure}[tb!]
%   \centering
%   \includegraphics[width=0.35\linewidth]{./img/drone_gradient_numerical.png}
%   \caption{Dither-Free Gradient Ascent.}
%   \label{fig:gradientFree}
% \end{figure}

This method presents a clear trade-off. The primary advantage is the simplicity and responsiveness of the algorithm. It provides an instantaneous heading measurement without requiring complex estimations or exploratory motions. The main disadvantage is the hardware complexity, which requires multiple sensors that increase the weight and cost compared to methods that can operate with a single sensor.

\subsubsection{Dither Extremum Seeking (DES)}

To reduce hardware requirements, DES employs a single photodiode. The underlying principle is gradient ascent: the drone attempts to continuously move in the direction of maximum light intensity ($J$). The main challenge is that with a single sensor, the gradient ($\nabla \boldsymbol{J}$) cannot be measured directly, as the drone lacks the multi-point data required for its calculation. Therefore, DES uses a model-free, real-time optimization technique that actively ``probes'' the environment to find the maximum gradient \cite{azzollini2021uavbasedsearchrescueavalanches, ZHANG2007245}.

The DES algorithm is shown in \autoref{fig:ditherES}. In a nutshell, the algorithm adds a small sinusoidal path ($a\cos(\omega t)$) into the drone's heading command. The sinusoid forces a continuous, small exploratory oscillation, or ``wiggle.'' As the drone oscillates, its single photodiode measures the fluctuations in the light intensity ($J$) and processes them in real-time to estimate the gradient's direction, directing the drone along the path of steepest ascent.

This method shifts the complexity from hardware to software. The primary advantage is that a single photodiode is needed. The main disadvantage is the inherent inefficiency of the flight path. The constant exploratory oscillation means the drone does not fly in a direct line to the target. Furthermore, this approach requires a more sophisticated onboard control algorithm to perform the real-time signal processing necessary to estimate the gradient. These shortcomings motivate us to develop a third strategy that attempts to remove this inefficient exploratory motion by using the drone's own state estimates to compute the gradient.

\subsubsection{Dither-Free Gradient Ascent (DGA)}

This strategy eliminates the inefficient probing motion by estimating the gradient from the drone's recent flight history. The objective is to retain the hardware simplicity of a single sensor while achieving a more direct and efficient flight path.
This method requires position information, but we do not rely on any external positioning system. Instead, the algorithm estimates the drone's location using its internal IMU data and the commanded speed ($v_{k}$). At each time step $k$, the estimated 2D position is updated as:

\begin{equation}
    \begin{bmatrix}
        x_{k} \\
        y_{k}
    \end{bmatrix} = \begin{bmatrix}
        x_{k-1} \\
        y_{k-1}
    \end{bmatrix} + v_{k} \begin{bmatrix}
        cos(\theta_{k-1}) \\
        sin(\theta_{k-1}
    \end{bmatrix} \Delta t
\end{equation}

While this estimation method is known to accumulate drift over long periods, it could be sufficiently accurate over short intervals to estimate the local gradient. As the drone moves, the controller stores a history of recent measurements as position-intensity pairs ($\boldsymbol{p}_{i} = (\boldsymbol{r}_{i}, J_{i})$). It then performs a least-square regression on this set of points to fit a first-order model (a plane), $\hat{J}(\boldsymbol{r}) = ax + by$, to the local light field, as visualized in \autoref{fig:gradientFree}. The resulting coefficients $(a, b)$ represent the estimated local gradient $\hat{\nabla}J = [a, b]^T$, which is used to set the yaw command $\hat\theta = atan2(b, a)$ and steer the drone along this numerically determined path of steepest ascent.

This method eliminates the overhead of the dither signal, providing a more direct path with a single sensor.
This efficiency, however, is not without cost. Unlike the model-free DES algorithm, this method's performance is directly coupled to the quality of its position estimates. If the odometry from the IMU is noisy or drifts significantly, the gradient calculation will be corrupted, potentially leading to navigation failure. This strategy, therefore, trades the control complexity of the DES method for a critical dependency on accurate state estimation.

\subsubsection{Comparing Navigation Strategies in Simulation}

\begin{table}[tb!]
\centering
\caption{Comparative Characterization of Navigation Algorithms}
\label{tab:algo_comparison}
\resizebox{\textwidth}{!}{%
\begin{tabular}{@{}lccc@{}}
\toprule
\textbf{Characteristic} & \textbf{Bearing Angle} & \textbf{Dither Extremum Seeking} & \textbf{Dither-Free Gradient} \\ \midrule
\textbf{Sensor Number} & Photodiode Array ($N>1$) & Single Photodiode & Single Photodiode \\
% & & & \\
\textbf{Position Data} & Not Required & Not Required & Required \\
%& & & \\
\textbf{Computational Load} & Low & Medium & High \\
\bottomrule
\end{tabular}
}
\end{table}

\begin{figure}[tb!]
    \centering
    \begin{minipage}[t]{0.5\linewidth}
        \centering
        \includegraphics[width=\linewidth]{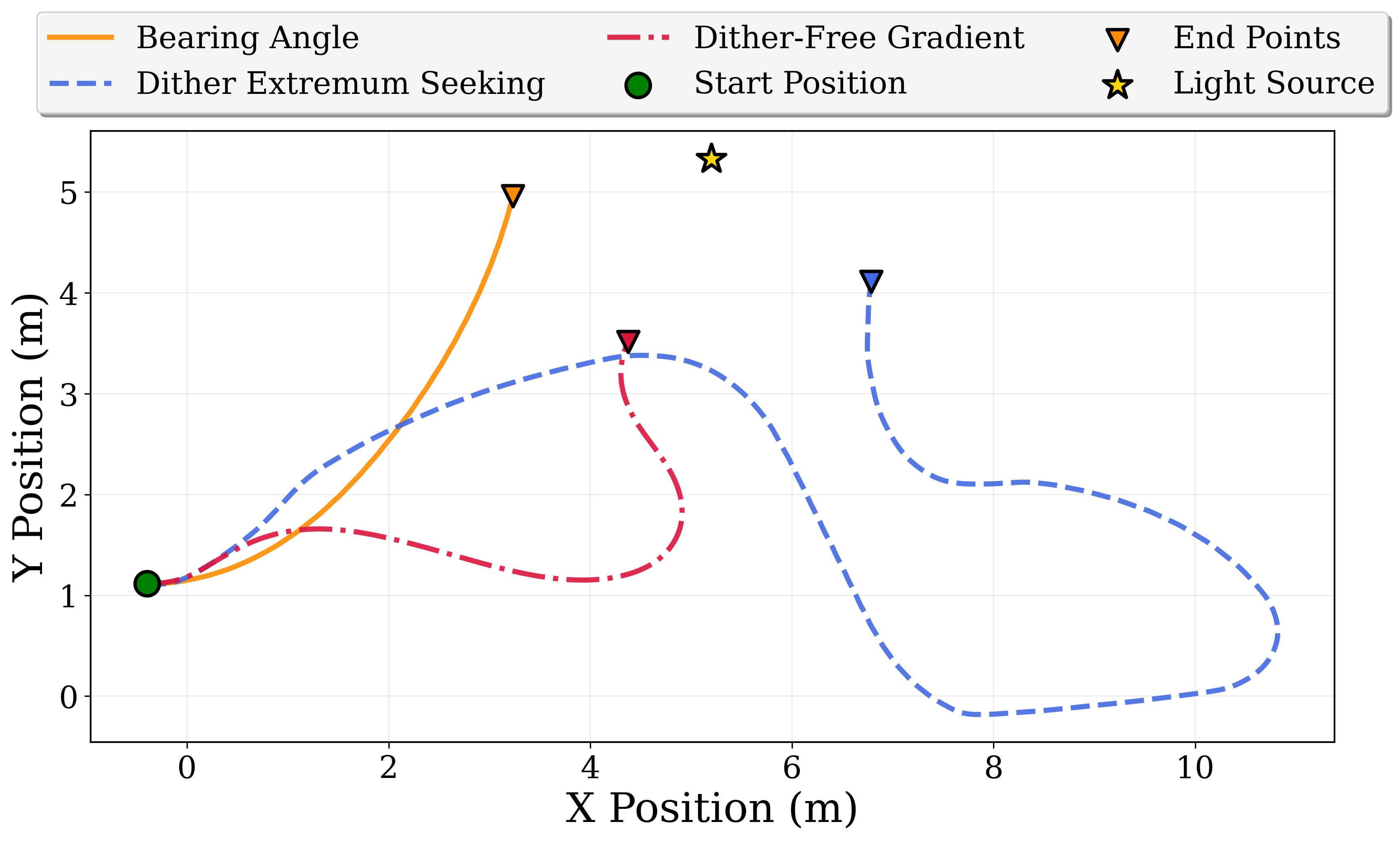}
        \captionof{figure}{Comparison of the three navigation strategies in a simulation environment.}
        \label{fig:simCompareNavigation}
    \end{minipage}
    \hspace{5mm} 
    \begin{minipage}[t]{0.4\linewidth}
        \centering
        \includegraphics[width=\linewidth]{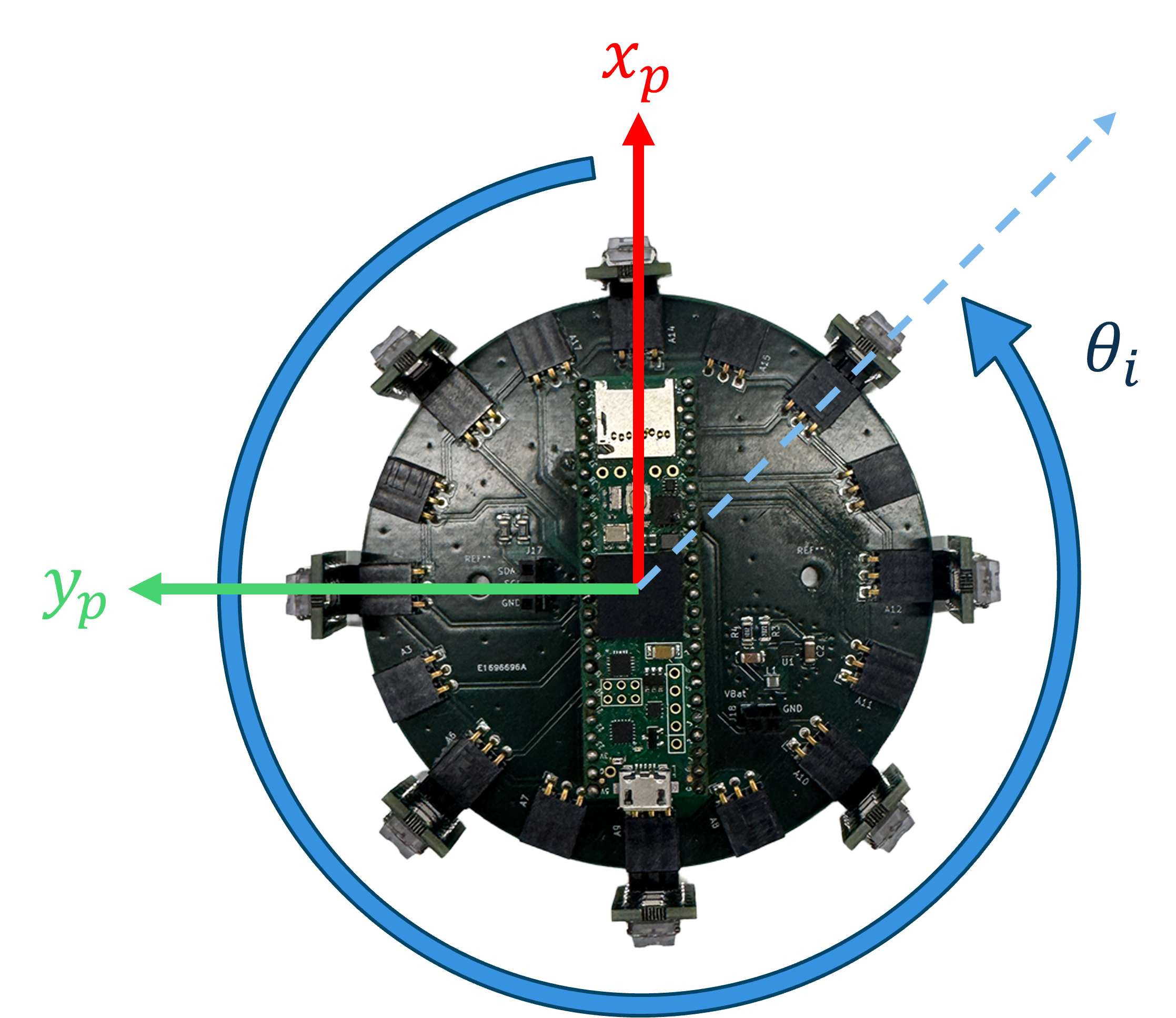}
        \caption{Navigation board: sensors \& MCU.}
        \label{fig:sensor_board}
    \end{minipage}
\end{figure}

The three navigation strategies present design trade-offs between hardware complexity, computational load, and reliance on state estimation, as summarized in \autoref{tab:algo_comparison}.
%
%BAG is computationally simple but requires multiple photodiodes. On the other hand, both the DES and DGA operate with a single photodiode, which is advantageous for payload-constrained LTA platforms. The primary difference between these two single-sensor methods lies in the need of prior position estimates. DES is ``model-free'' (requiring no position data) but must use an inefficient ``wiggling'' motion to search the gradient. DGA avoids this inefficient motion by using prior position estimates, but is vulnerable to erroneous estimations.
%
To evaluate more thoroughly the design trade-offs, we implement the three algorithms in our simulator, described in \autoref{sec:platform_comparison}. In that setup, a light source is placed seven meters away from the drone, and the drone must get at least two meters from the target (this safe close-distance is required due do the diameter of the envelope).
The resulting trajectories, plotted in \autoref{fig:simCompareNavigation}, show the significant performance differences between the algorithms. The Bearing Angle method (orange) provides the most direct and smooth trajectory, confirming the advantage of having a sensor array. On the other hand, DES (blue), while using minimal resources, requires continuous random probing. This exploratory motion results in a significantly longer and convoluted path. DGA provides a shorter travel path, which is a compromise between the two extreme methods. DAG has a brief overshoot at the beginning, but quickly updates its gradient to converge efficiently on the target.

The simulations provide valuable insight, but still leave important questions open: What is the minimum number of photodiodes that the Bearing method needs in practice to perform efficiently? And, how would the algorithms perform in real environments where wind can be present? In the next subsection, we identify an optimal number of sensors for BAG and in our evaluation section we compare all three algorithms under various challenging conditions.

\subsection{Analysis Of Optimal Photodiode Count for the BAG Algorithm}
\label{subsec:pd_analysis}

To determine the optimal number of photodiodes for the Bearing method, first, we describe the hardware we develop and how we process the light signals, and then, we perform the photodiode analysis.

\subsubsection{Transmitter and receivers}
\label{sensorsVLC}

Our light-based navigation system must operate with both natural and artificial light, requiring it to handle a dynamic range of intensities from complete darkness to direct sunlight. To ensure the drone can reliably distinguish the target emitter from ambient light, we employ a simple modulation strategy. The emitter oscillates at a specific frequency, and the receiver uses a Fast Fourier Transform (FFT) to isolate that frequency from background noise. Next, we discuss our hardware setup for this approach.

\textbf{Emitter.}
The emitter is shown in \autoref{fig:exploded_view}. It is the same LED used for energy harvesting in \autoref{sec:energy-harvesting}, but modulated by an Arduino DUE and a DFR0457 MOSFET switcher. The LED emits a specific frequency to differentiate it from ambient light: $f_{mod}$, square wave with 50\% duty cycle. Since warehouses and greenhouses use high power LEDs, the MOSFET switcher is used to cleanly modulate the 100W LED. The LED has a 120-degree field of view with a 60-degree half-power angle. No external optics are used.

% \begin{figure}[tb!]
%     % \centering
    
%     % \begin{minipage}[b]{0.35\linewidth}
%         \centering
%         \includegraphics[width=0.4\linewidth]{img/allphoto_coordinate.png}
%         \captionof{figure}{Navigation board: sensors \& MCU.}
%         \label{fig:sensor_board}
    % \end{minipage}
%     \hfill % Adds horizontal space
%     \begin{minipage}[b]{0.6\linewidth}
%         \centering
%         \includegraphics[width=\linewidth]{img/paperadc.png}
%         \captionof{figure}{Sensor raw response.}
%         \label{fig:sensor_value}
%     \end{minipage}
%     \begin{figure}[tb!]
%   \centering
%   \includegraphics[width=0.6\linewidth]{./img/simulation_trajectory_comparison_20251028_185945.png}
%   \caption{Comparison of the three navigation strategies in a simulation environment}
%   \label{fig:simCompareNavigation}
% \end{figure}

% \end{figure}

\textbf{Receiver.}
We build a custom sensor board for the photodiodes and the MCU processing their signals, shown in \autoref{fig:sensor_board}. The PCB follows a modular design. Instead of having a fixed number of sensors, the board has female sockets in a circle with a $22.5^\circ$ offset. Small, pluggable modules holding a single OPT101 light sensor can be inserted into these sockets. This modularity allows us to easily test algorithms with different number of sensors (up to 16 for a full 360-degree view) and adjust the sensitivity of each module via different resistors.
A Teensy 4.1 microcontroller (MCU) processes the light intensity signals from each sensor.

\textbf{Signal Processing.} 
The MCU continuously samples each sensor at a set frequency ($f_{sample}$). Once the buffer is full, the signal follows a Fast Fourier Transfrom (FFT).
To identify the peak signal corresponding to our emitter ($f_{mod}$, e.g., 150 Hz), we first calculate its corresponding target bin, $k_{target}$, in the FFT array:

\begin{equation}
k_{target} = \text{round}\left(\frac{f_{mod} \cdot N_{FFT}}{f_{sample}}\right)
\end{equation}

To account for small frequency drifts, we search a small window around this target bin to find the exact peak magnitude ($M_{peak}$). %in the bin $k_{peak}$.
Finally, we calculate the Signal-to-Noise Ratio (SNR) following the standard process of first filtering the bins of the known harmonics, and then, calculating the SNR as:

\begin{equation}
\text{SNR} = \frac{M_{peak}}{M_{noise}} = \frac{M(k_{peak})}{\frac{1}{N_{noise}} \sum_{k \in \mathcal{K}{noise}} M(k)}
\end{equation}

where $\mathcal{K}{noise}$ is the set of all frequency bins excluding the windows around its harmonics. This SNR value is vital, as it tells the navigation algorithms whether a sensor's reading is reliable enough to use.

\subsubsection{Optimal Number of Photodiodes for the Bearing Angle Method}
\label{OptimalPhoto}

The Bearing Angle Guidance (BAG) algorithm estimates the light’s direction by computing a weighted average of signals from multiple photodiodes. This approach introduces a critical design trade-off between sensing accuracy and payload efficiency. Increasing the number of sensors improves angular precision but also adds weight and power consumption. To identify the most effective configuration for our LTA platform, we experimentally evaluate how BAG's performance scales as the number of photodiodes decreases.

We test three configurations using our modular sensor board: a high-density array with 16 photodiodes ($22.5^\circ$ spacing), a medium-density array with 8 photodiodes ($45^\circ$ spacing), and a sparse array with 4 photodiodes ($90^\circ$ spacing). Each configuration is tested in two environments: \textit{indoor}, with moderate ambient light and reflective surfaces; and \textit{outdoor}, under strong sunlight and minimal reflections.

\begin{figure}[tb!]
    \centering % Center the entire figure

    % First subfigure
    \begin{subfigure}[b]{0.3\linewidth}
        \centering
        \includegraphics[width=\linewidth]{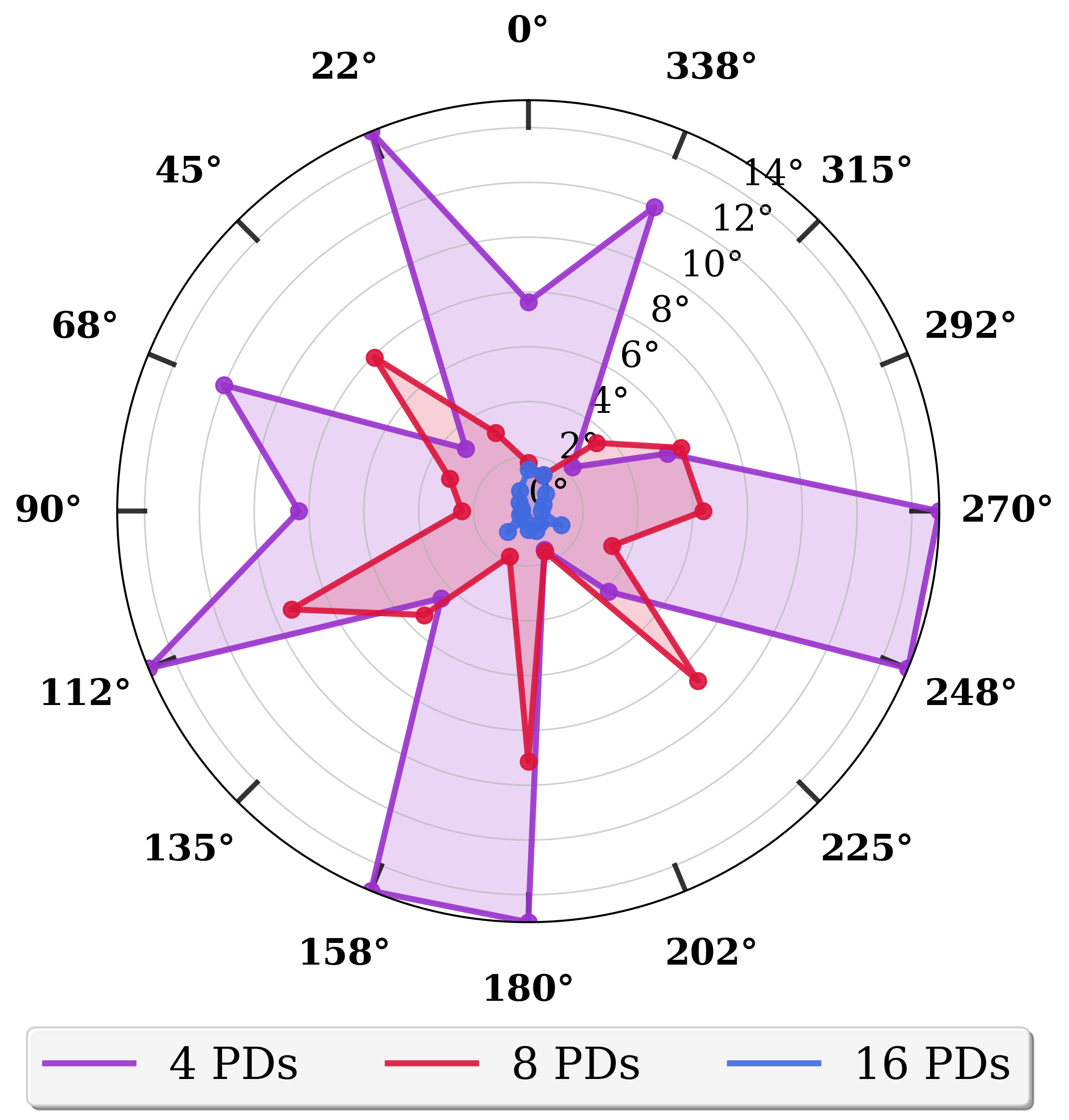}
        \caption{Indoors}
        \label{fig:indoorBearing}
    \end{subfigure}
    \hspace{5mm}  % This command creates the space between the two images
    % IMPORTANT: Do not leave a blank line here
    % Second subfigure
    \begin{subfigure}[b]{0.3\linewidth}
        \centering
        \includegraphics[width=\linewidth]{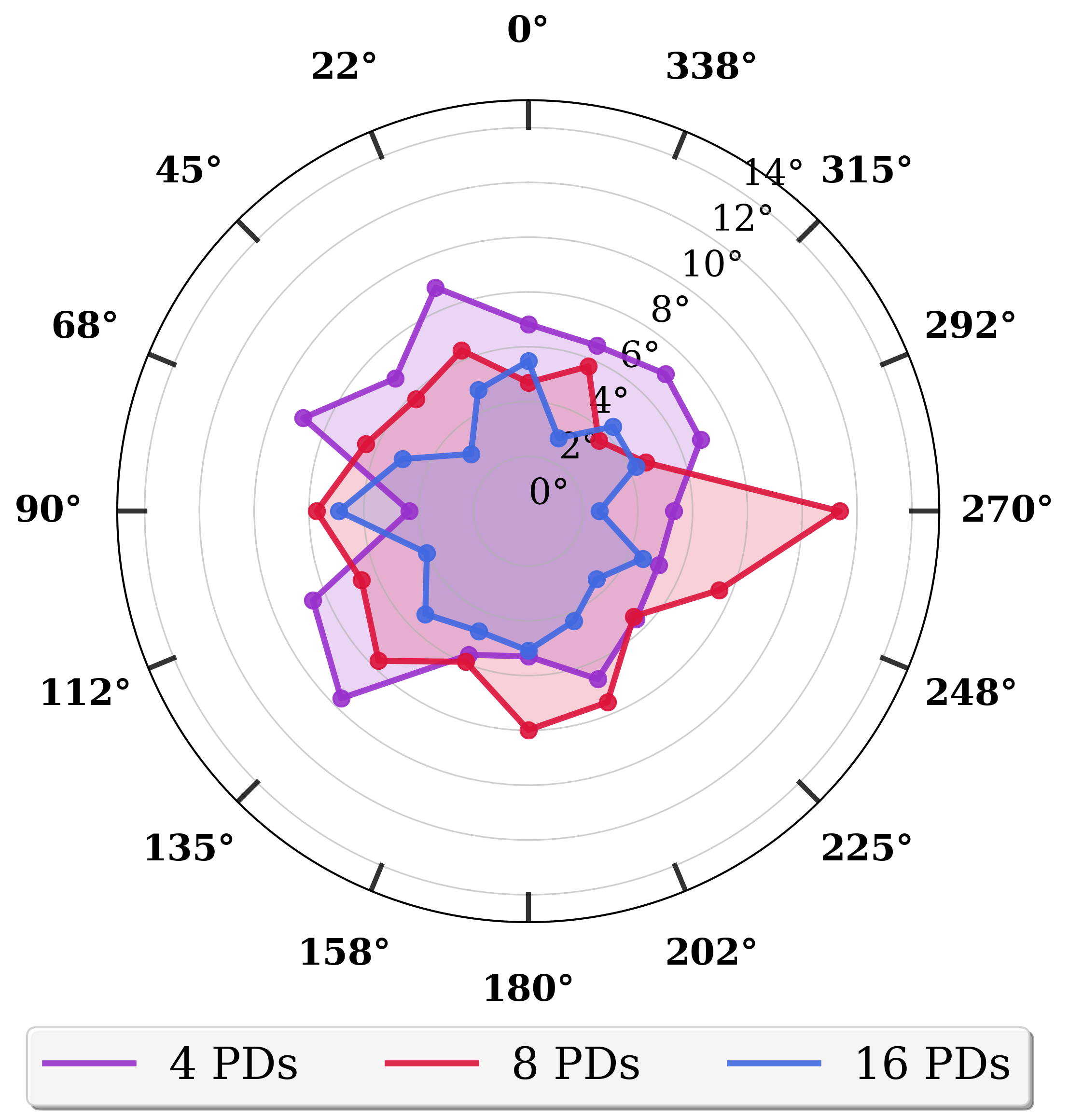}
        \caption{Outdoors}
        \label{fig:outdoorbearing}
    \end{subfigure}

    \caption{Bearing angle estimation for indoor and outdoor scnearios.}
    \label{fig:bearingCombine}
\end{figure}
To measure bearing accuracy, the emitter is positioned 3\,m in front of the sensor array. The array is rotated in $22.5^\circ$ increments over a full $360^\circ$ rotation, yielding 16 ground-truth angles. At each orientation, we record 50 measurements to compute the median angular error. The results are shown in \autoref{fig:bearingCombine}, where the angle represents the ground truth direction, and the radius of each point indicates the median error for that direction.

\begin{enumerate}
    \item \textbf{Indoor results} (\autoref{fig:indoorBearing}):
A clear relationship emerges between sensor density and accuracy. The 16-photodiode array achieves excellent precision, maintaining errors below 2$^\circ$ at all angles. The 8-photodiode array remains reliable, showing slightly higher but consistent errors. Twelve directions have errors below 4$^\circ$, and the other four are below 8$^\circ$. In contrast, the 4-photodiode array proved highly unreliable, with several directions exceeding 10$^\circ$ error. Low-error readings occur only when the light source is positioned symmetrically between two adjacent sensors, allowing their signals to balance. At other orientations, a single diode dominates the estimate, making the result highly sensitive to reflected light from nearby surfaces. These reflections act as secondary emitters, producing significant noise and estimation errors.
    \item \textbf{Outdoor results} (\autoref{fig:outdoorbearing}): Overall, the strong presence of sunlight degrades the signal-to-noise ratio for all configurations, leading to a general increase in the median error. However, the performance gap between arrays narrows: the 16- and 8-photodiode arrays maintain similar accuracy, while the 4-photodiode array becomes more stable than indoors. This improvement is likely due to the dominance of a single, bright light source (the Sun) and the lack of (reflecting) surfaces in outdoor scenarios, which minimize the impact of reflections and multipath effects.
\end{enumerate}

Based on these results, we select the 8-photodiode configuration as the optimal choice for the BAG algorithm. It strikes the best balance among accuracy, weight, and system complexity. The 8-sensor setup avoids the extreme unreliability of the 4-photodiode array in reflective indoor environments while offering substantial weight savings compared to the 16-sensor configuration. For the DES and DAG algorithms, we use a single photodiode.

\section{Evaluation}

In this section, we empirically validate the performance of our integrated LTA platform. First, we begin with a comprehensive energy analysis. Second, we establish the effective operational range of our ``point-and-go'' navigation system. Third, we compare our three navigation algorithms (BAG, DES, and DGA), first in a controlled indoor environment to evaluate their path efficiency and speed, and then in challenging outdoor tests to determine their real-world robustness against wind.

\subsection{Platform Energy Analysis}

To validate our LTA drone's energy efficiency, we analyze its power consumption compared to conventional drones. Our analysis demonstrates that our approach has two key advantages for endurance: first, leveraging buoyancy significantly reduces power requirements for hovering and flying, and second, onboard solar charging provides a crucial mechanism to enable sustained autonomous operation.

\begin{figure}[tb!]
    \centering
    \begin{minipage}[t]{0.4\linewidth}
        \centering
        \includegraphics[width=\linewidth]{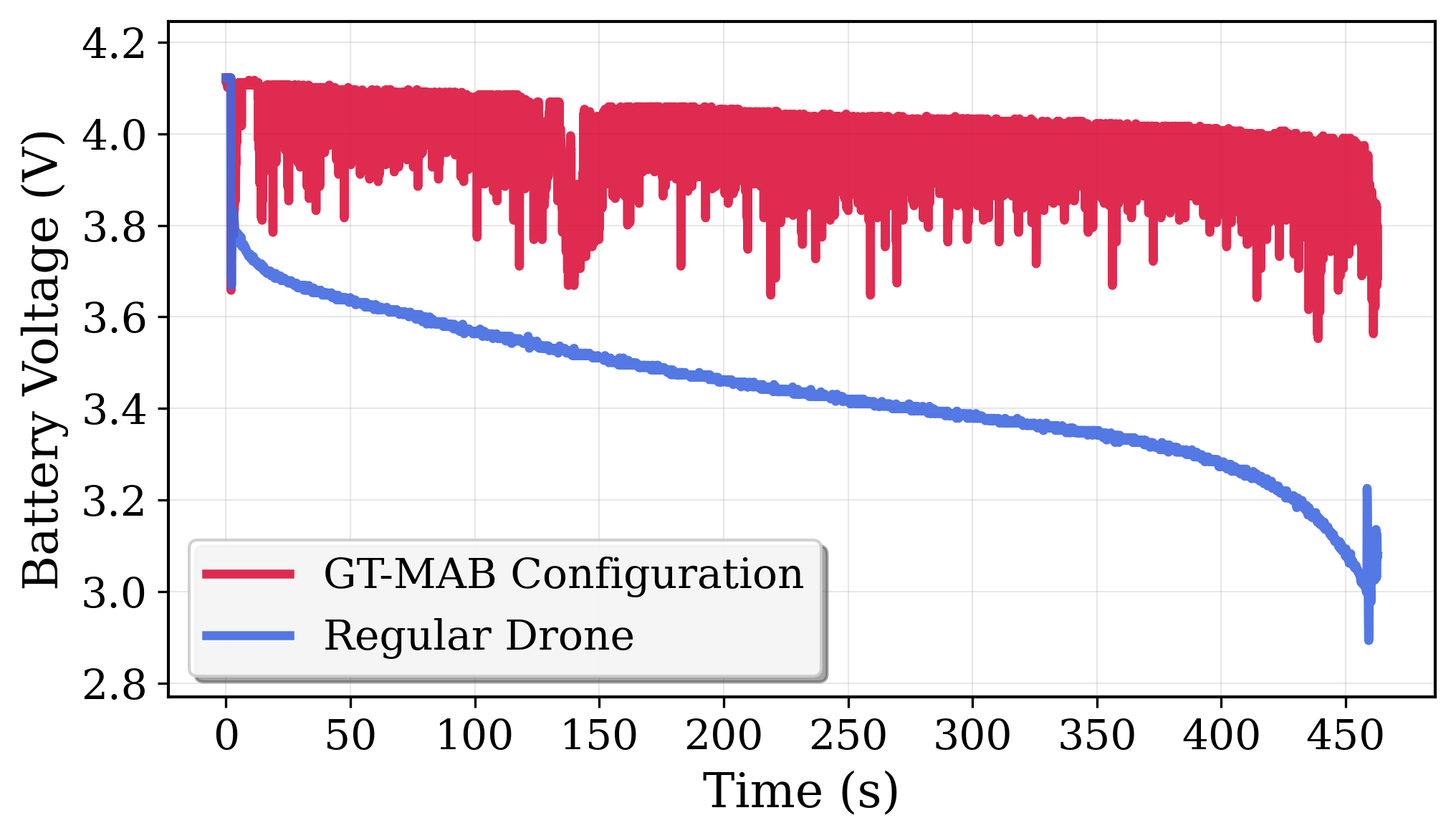}
        \captionof{figure}{Battery voltage over time while hovering}
        \label{fig:BatteryComp}
    \end{minipage}
    \hspace{5mm} 
    \begin{minipage}[t]{0.55\linewidth}
        \centering
        \includegraphics[width=\linewidth]{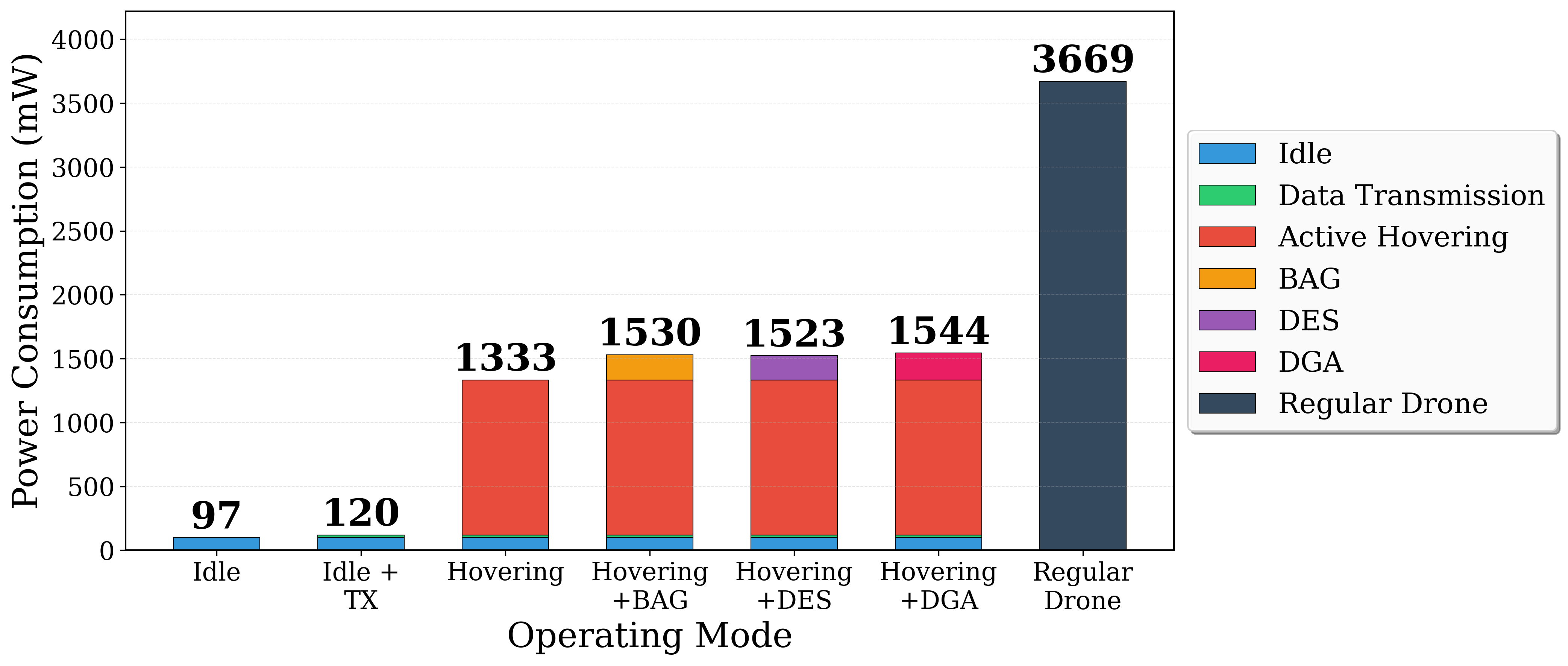}
        \caption{Power budget of different operations}
        \label{fig:powerconsumption}
    \end{minipage}
\end{figure}

\begin{figure}[tb!]
    \centering 
    \begin{minipage}[t]{0.42\linewidth}
        \centering
        \includegraphics[width=\linewidth]{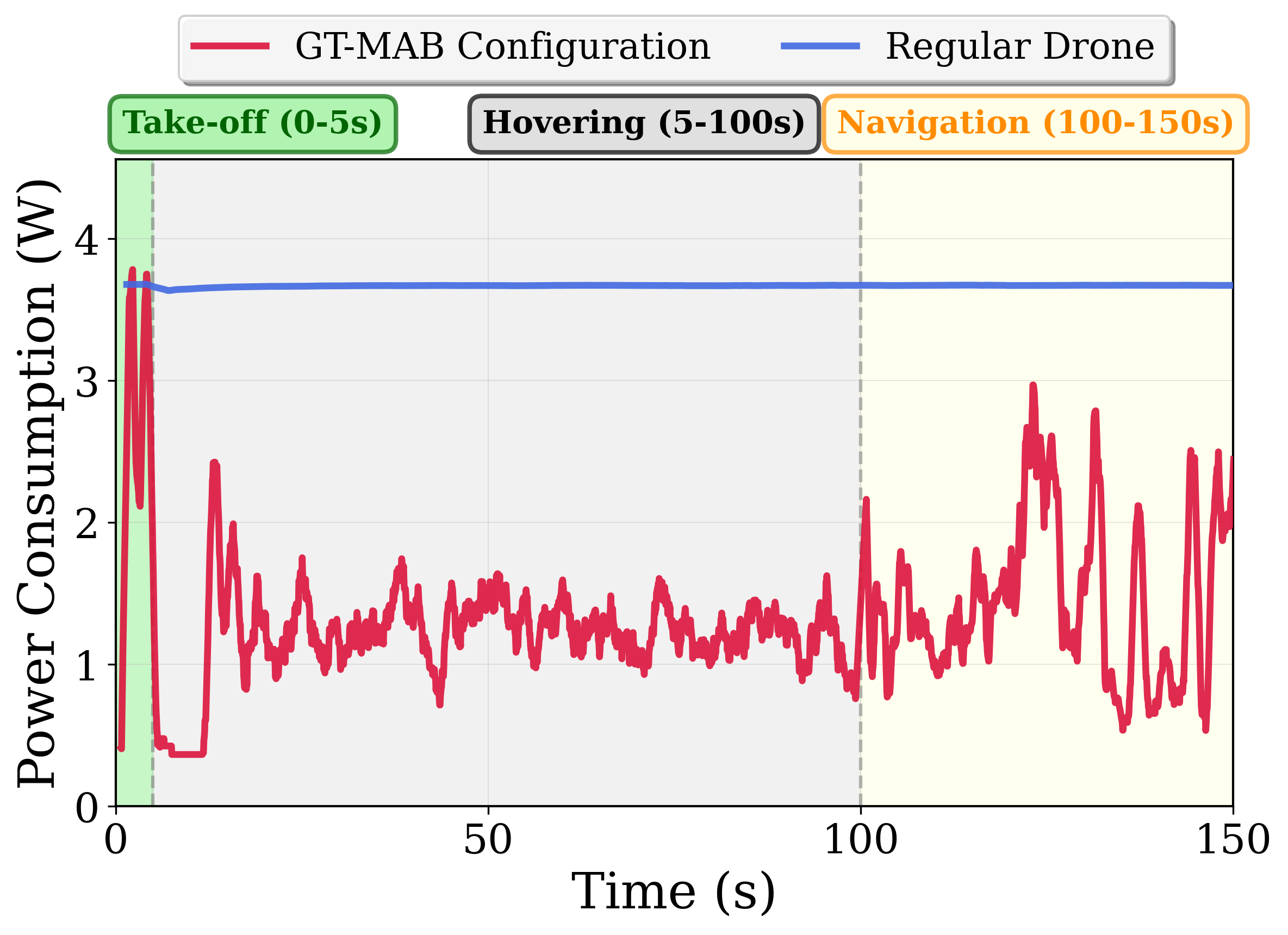}
        \captionof{figure}{Power consumption: LTA vs. standard drone}
        \label{fig:powerCompare}
    \end{minipage}
\hspace{5mm} 
    \begin{minipage}[t]{0.42\linewidth}
        \centering
        \includegraphics[width=\linewidth]{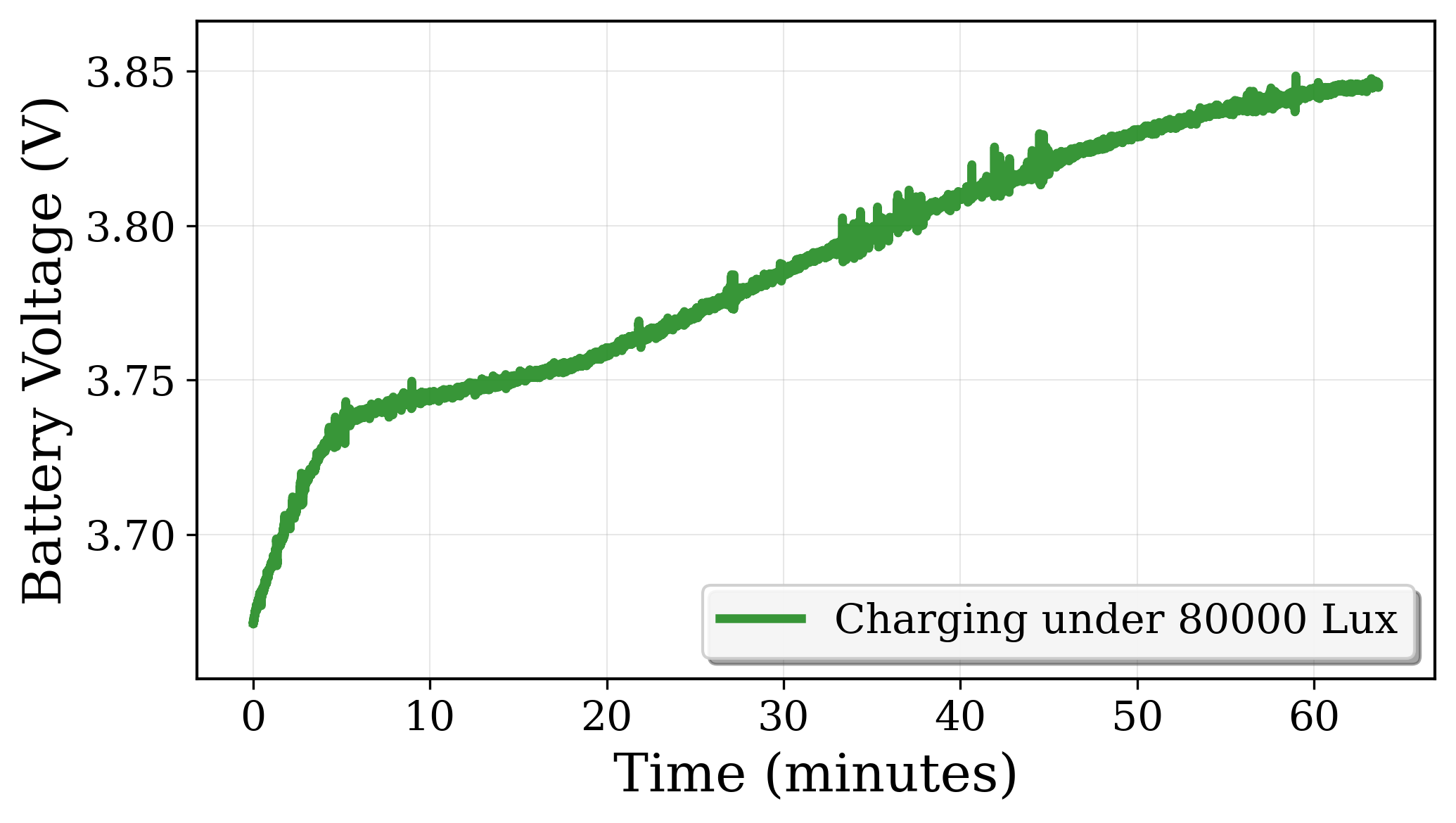}
        \captionof{figure}{Power production: Solar energy harvesting}
        \label{fig:chargingTime}
    \end{minipage}
\end{figure}

\begin{itemize}
    \item \textbf{Hovering Efficiency.} To quantify the fundamental energy savings from buoyancy, we compare the hovering efficiency of our LTA drone against a standard microdone. Both platforms use identical batteries and hover at the same altitude indoors for 450 seconds. 
    \autoref{fig:BatteryComp} shows the battery voltage over time for this test. The standard quadrotor's voltage (blue line) drops rapidly as it constantly consumes power to fight gravity, whereas our LTA platform's voltage (red line) drops only a small fraction. This significantly smaller power consumption for lift confirms the primary advantage of the LTA design. The small fluctuations seen in the LTA drone's voltage trace simply reflect intermittent motor use for minor stabilization adjustments to maintain the instructed position, not continuous lifting effort.
    
    \item \textbf{Operational Power Budget.} Beyond the energy needed for hovering, other functions also consume power. We characterize the operational power budget of various required activities, shown in ~\autoref{fig:powerconsumption}. The platform consumes the least power when the motors are off: idle (97\,mW) and when wireless data transmission is added (120\,mW). Actively stabilizing while hovering requires more power (1333 mW), showing that while buoyancy eliminates the cost of lift, precise position holding still requires non-trivial energy. However, the stabilization energy is almost one third of the energy consumed by a standard drone (black bar). Running the different light-based navigation algorithms (BAG, DES, DGA) adds a modest overhead, bringing the total power consumption during navigation to the 1523-1544\,mW range. 

    \item \textbf{Energy-Harvesting Potential.} To assess the potential for autonomous operation, we need to quantify the ratio between energy consumption and production. Regarding energy consumption, the most important component are the rotors. \autoref{fig:powerCompare} shows the power consumption of our LTA platform during three tasks: (i) take-off, (ii) stabilization while hovering, and (iii) flying towards a given point. Note that the LTA platform not only saves energy while hovering but also during flight time. The figure also shows the hovering power of a standard drone. Regarding energy production, \autoref{fig:chargingTime} shows the steady increase in battery recharging over one hour under an illumination of 80000 lux\footnote{To put this light intensity in context, one needs to consider that standard office lights radiate several thousand lux. In warehouses, where ceilings are much higher, the light intensity on-flight (i.e., closer to the LEDs) is tens of thousands of lux. In greenhouses, sunlight can easily provide levels above 100\,k lux during daylight, and during darker periods, the light fixtures provide strong illumination for plants, including the IR bands, which provide the strongest energy for harvesting, but we did not consider greenhouse LEDs.}. 
    With these values for energy production and consumption, our platform obtain a charging ratio of 4:1 for flying, that is, to fly for a time period $t$, the platform needs to charge for $4t$. The charging ratio for stabilization (active hovering) is 3:1. Thus, depending on the ratio between flying and hovering, our platform could provide periodic flights of approximately 18 minutes for every hour of recharging with daylight.  
    %This capability, combined with the inherently low power required for buoyant hovering, creates a viable path toward sustained, long-duration autonomous operation, particularly for monitoring tasks involving extended stationary periods. 
\end{itemize}

\subsection{Platform Range Analysis}

Our light-based navigation requires low infrastructure: a single emitter within the photodiodes' field-of-view, rather than a dense, precisely mapped anchor grid. In the visible light positioning (VLP) literature, higher anchor density correlates strongly with a lower positioning error. For example, a previous study reports an average error of $\approx\!4$\,cm using 4 LEDs over a 1m x 1m area (4 LED/m$^2$) \cite{Niu2021ICC}. While a 2m x 2m testbed with four LEDs ($1$\,LED/m$^2$) yields a mean error of $\approx\!20.6$\,cm under mobile (6-DoF) flight, illustrating the accuracy–density trade-off \cite{10257235}. In contrast to those multi-anchor settings, our navigation objective is not to provide centimeter-accurate localization. Instead, we propose a reliable directional cue and sufficient signal quality to drive goal-seeking navigation. 

\begin{figure}[tb!]
\centering 
\begin{subfigure}[b]{0.45\linewidth}
\centering
    \includegraphics[width=0.8\linewidth]{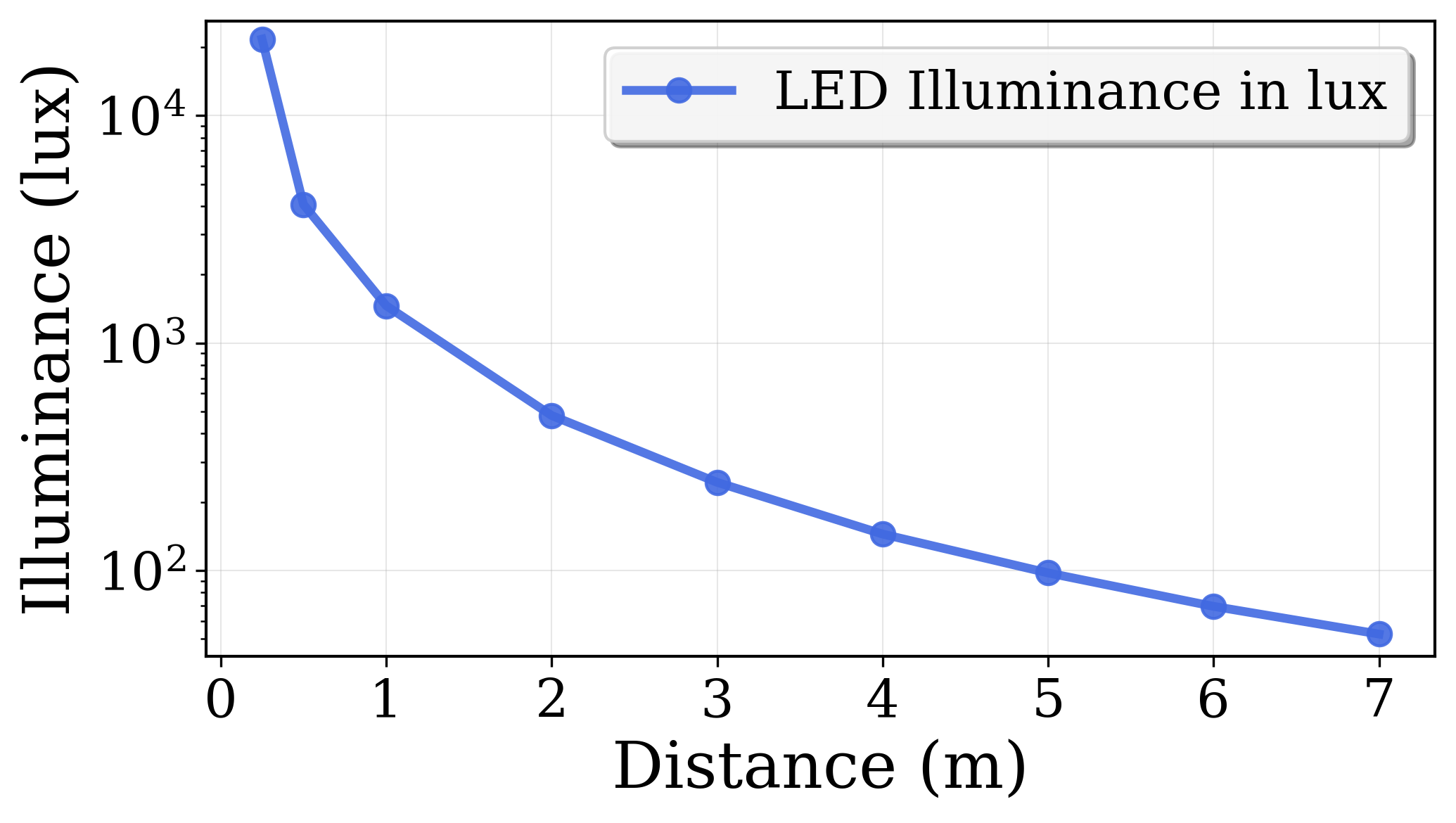} % <-- !! REPLACE THIS !!
    \caption{LED Lux measurement compare to the distance}
    \label{fig:LuxMeasure}
\end{subfigure}
\hspace{4mm} 
\begin{subfigure}[b]{0.45\linewidth}
    \centering
    \includegraphics[width=0.8\linewidth]{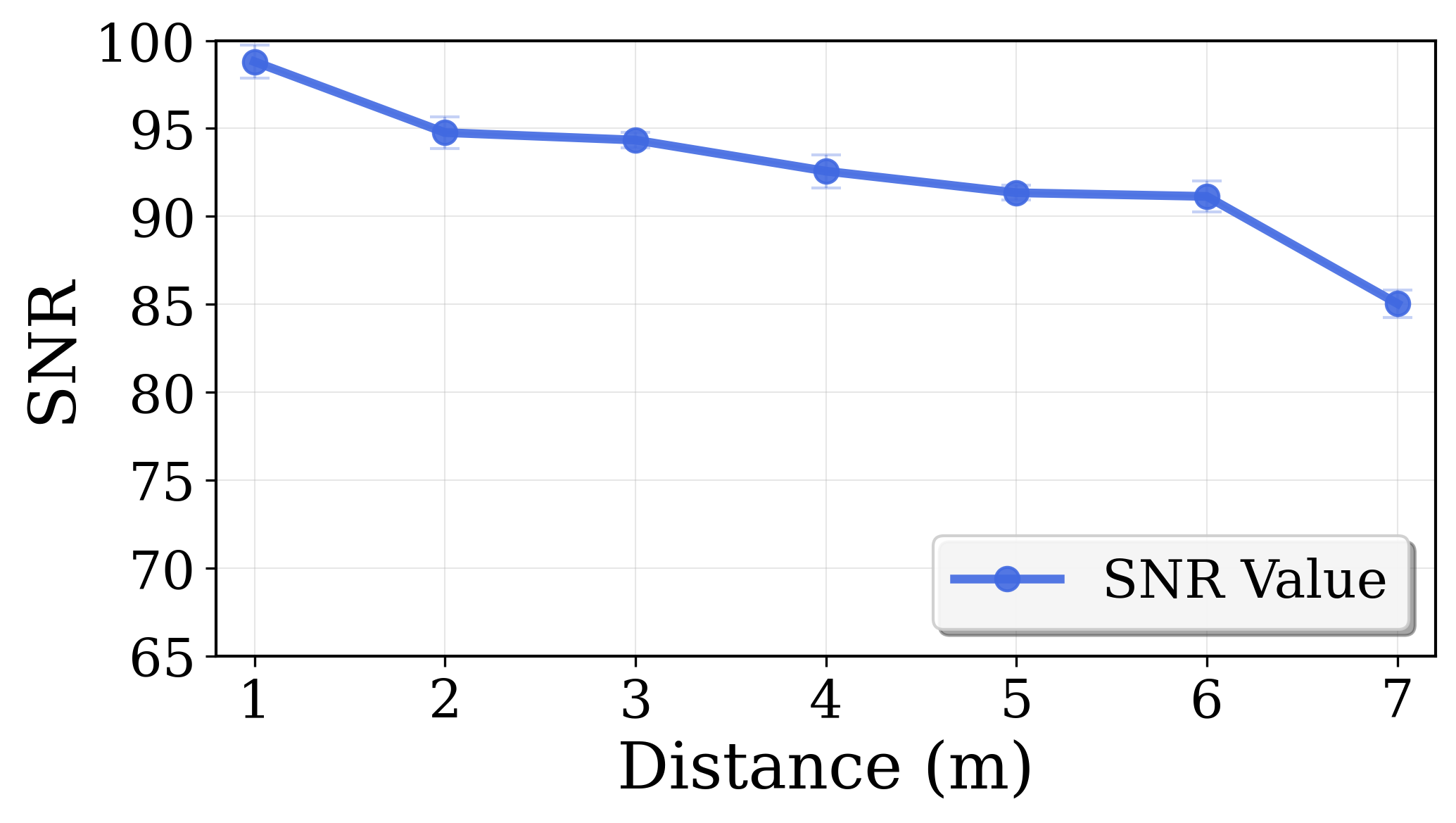}
    \caption{System's SNR measurement compare to the distance}
    \label{fig:PhotoSNR}
\end{subfigure}
% --- Main Caption and Label ---
\caption{Lux and SNR values over distance}
\label{fig:lightMeasurement}
\end{figure}

To measure our operating range, we empirically characterize the received signal and noise as a function of distance. First, to establish baseline signal strength, we measure illuminance (lux) versus distance for our emitter in a dark room  (\autoref{fig:LuxMeasure}). At 7 m, the emitter still produces several tens of lux, and our demodulation pipeline consistently recovers the beacon’s ID at these levels. In fact, prior studies have shown that, in dark scenarios, illumination of a few lux can still be used to decode FSK signals~\cite{tapia2024sol}.
Second, to consider interference, we evaluate the SNR over distance under typical ambient lighting (\autoref{fig:PhotoSNR}), computing SNR from the FFT magnitude as described in the prior section. The system maintains high SNR across the tested range, remaining robust at 7 m, indicating that the narrowband demodulation sufficiently suppresses unmodulated ambient light.

Together, these measurements demonstrate a reliable operational range of at least 7 m for single-beacon guidance, even with interference from ambient light. This is a significant finding: a 7 m radius from a single emitter covers an area of over 150 m$^2$. This supports an extremely sparse deployment, which is orders of magnitude lower density than multi-anchor VLP baselines requiring 1 LED/m$^2$ or more for grid-based localization. Our results, therefore, show that effective navigation can be achieved with minimal infrastructure.

\subsection{Navigation Evaluation in \textbf{Indoor} environment}

Having defined three distinct navigation algorithms in \autoref{sec:light-navigation}, we now evaluate their practical performances in an autonomous tracking task. This experiment aims to quantify the real-world trade-offs between the hardware simplicity of single-photodiode methods (DES and DGA) and the potential efficiency of the multi-sensor Bearing Angle (BAG) algorithm. We compare them on two primary metrics: navigation efficiency (total path length) and speed (total travel time).

\begin{figure}[tb!]
    \centering 
    \begin{subfigure}[b]{0.4\linewidth}
        \centering
        \includegraphics[width=0.9\linewidth]{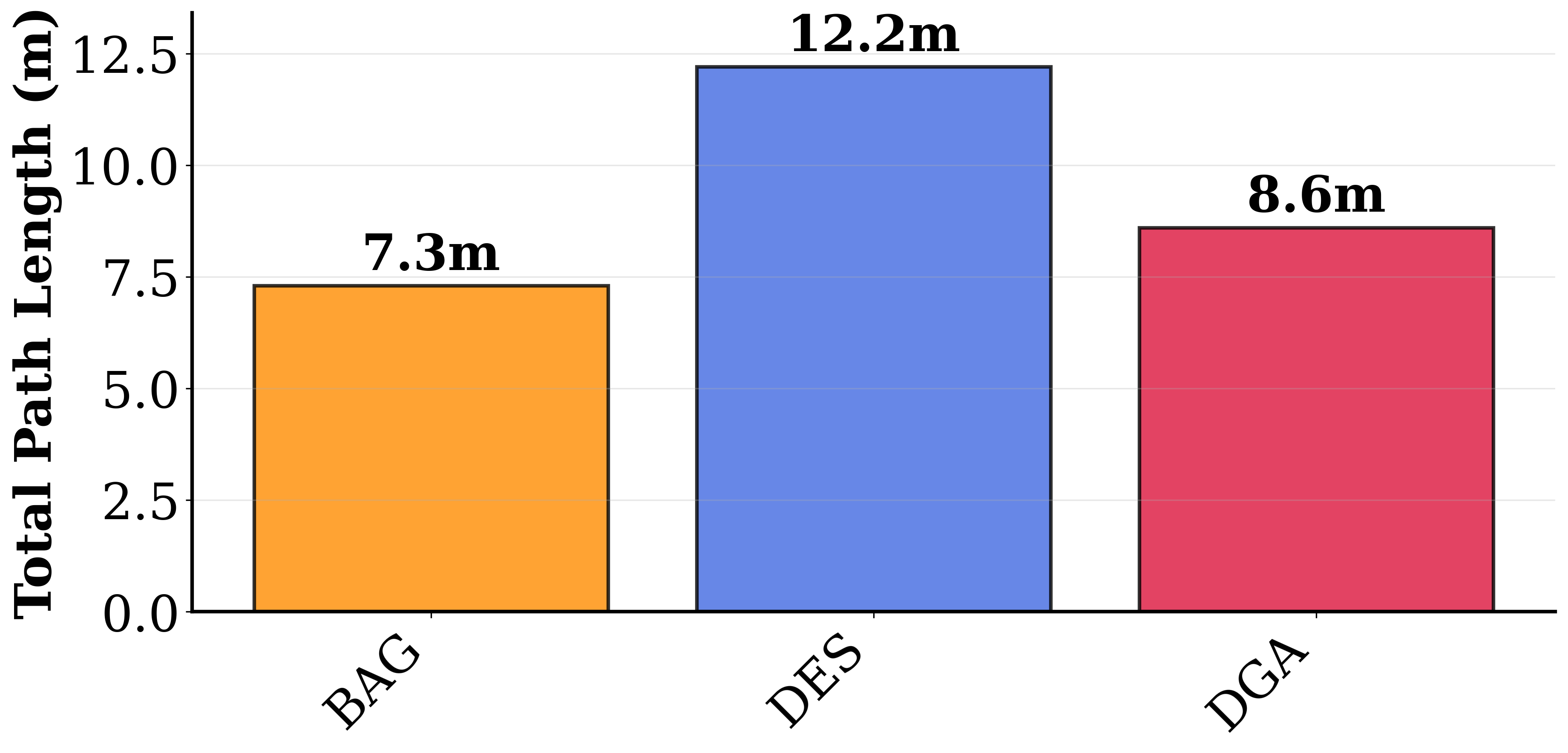}
        \caption{Indoor travel path length }
        \label{fig:indooPathQuant}
    \end{subfigure}
    \hspace{5mm} 
    \begin{subfigure}[b]{0.4\linewidth}
        \centering
        \includegraphics[width=\linewidth]{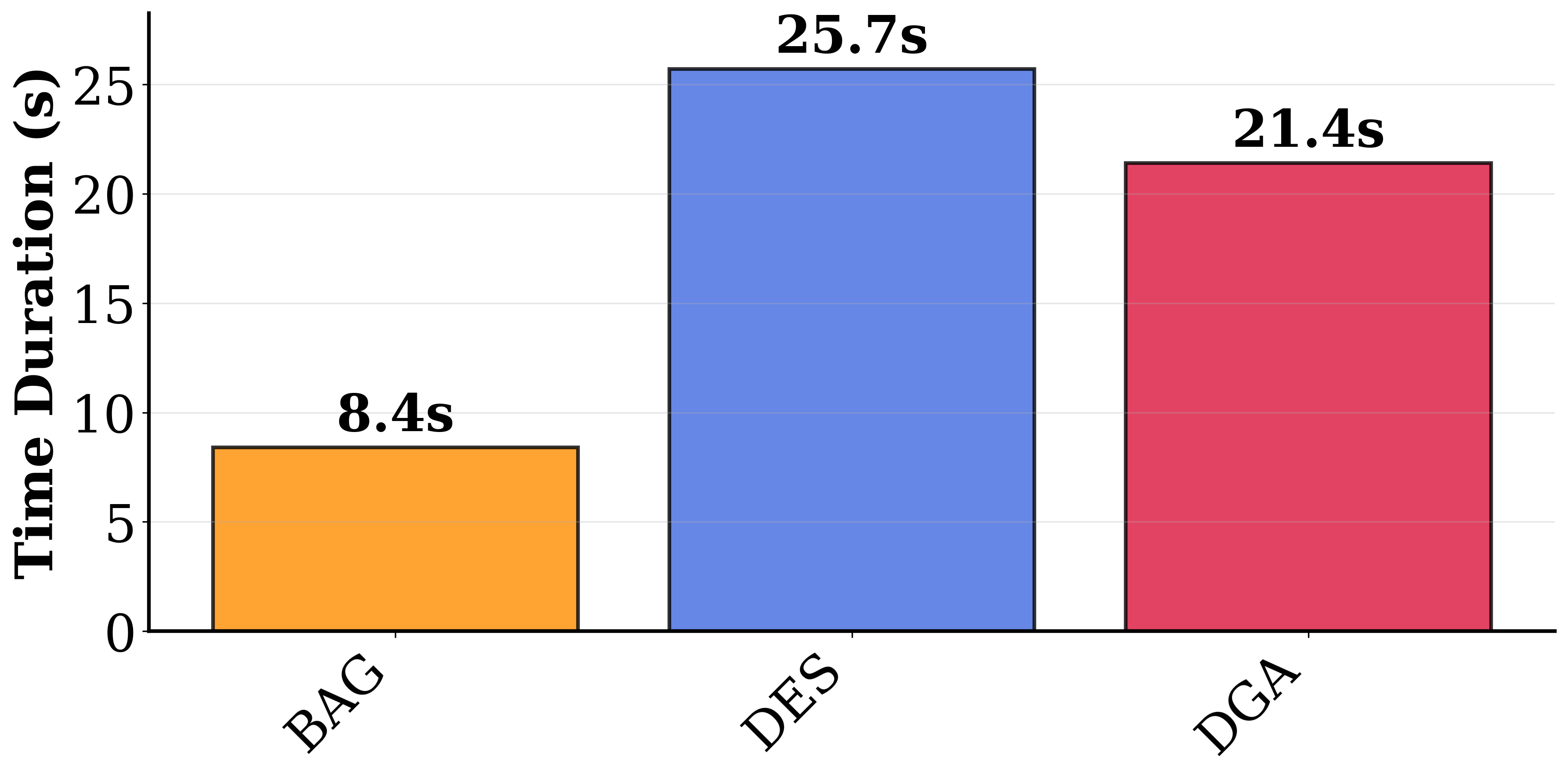}
        \caption{Indoor travel time}
        \label{fig:outdooPathQuant}
    \end{subfigure}
    \caption{Comparison of travel path length and time for three navigation algorithms}
    \label{fig:lightSourceIndoor}
\end{figure}

\textbf{Experimental setup.} We conduct the test in an indoor space, which represents our target scenarios (e.g., warehouses) and minimizes external disturbances like wind. In each trial, the drone started at a fixed distance of 7 meters from the light emitter and autonomously navigate toward it. A test is considered successful when the drone gets within 1-meter distance of the emitter. The ground truth location of the drone, for all indoor and outdoor scenarios, is obtained via visual odometry. To isolate the performance of the navigation algorithm, the drone maintains a constant forward speed of 0.5 m/s and a hovering altitude of 0.5 m throughout each experiment. 
%To ensure a fair comparison, we configured the key parameters for each navigation algorithm to reach the target in our experiment setup. 
The Bearing Angle algorithm uses an 8-photodiode array, the configuration we previously identified as optimal in \cref{OptimalPhoto}. The Dither Extremum Seeker algorithm and the Dither-Free Gradient Ascent use a single photodiode placed at $0°$ on the board. %Due to our confined experiment space, the dither angle should be wide to prevent collapse with the wall. It also needs a moderate dither frequency for the drone to respond sufficiently. Thus, its parameters were set to $\omega = 0.3$ and $a = 90°$. The Dither-Free Numerical Gradient Ascent algorithm used the same single photodiode. It stores the magnitude response of the filtered frequency measurements in a spatial map with a 2cm x 2cm resolution to compute the local light gradient.

\begin{figure}[tb!]
    \centering % Center the entire figure
    \begin{subfigure}[b]{0.45\linewidth}
        \centering
        \includegraphics[width=0.9\linewidth]{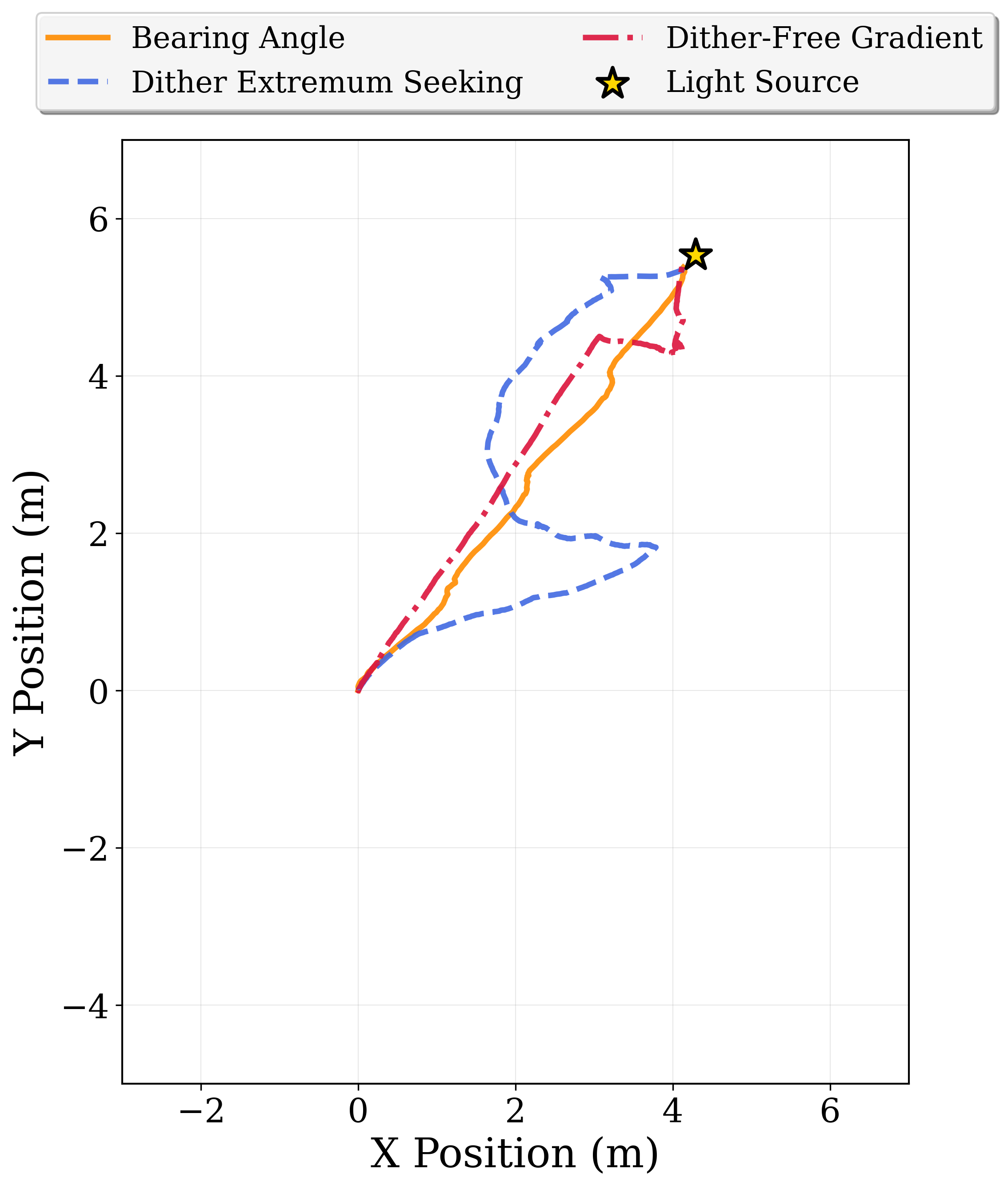}
        \caption{Indoor travel trajectory }
        \label{fig:indoorTraj}
    \end{subfigure}
    \hspace{5mm}
    \begin{subfigure}[b]{0.45\linewidth}
        \centering
        \includegraphics[width=0.9\linewidth]{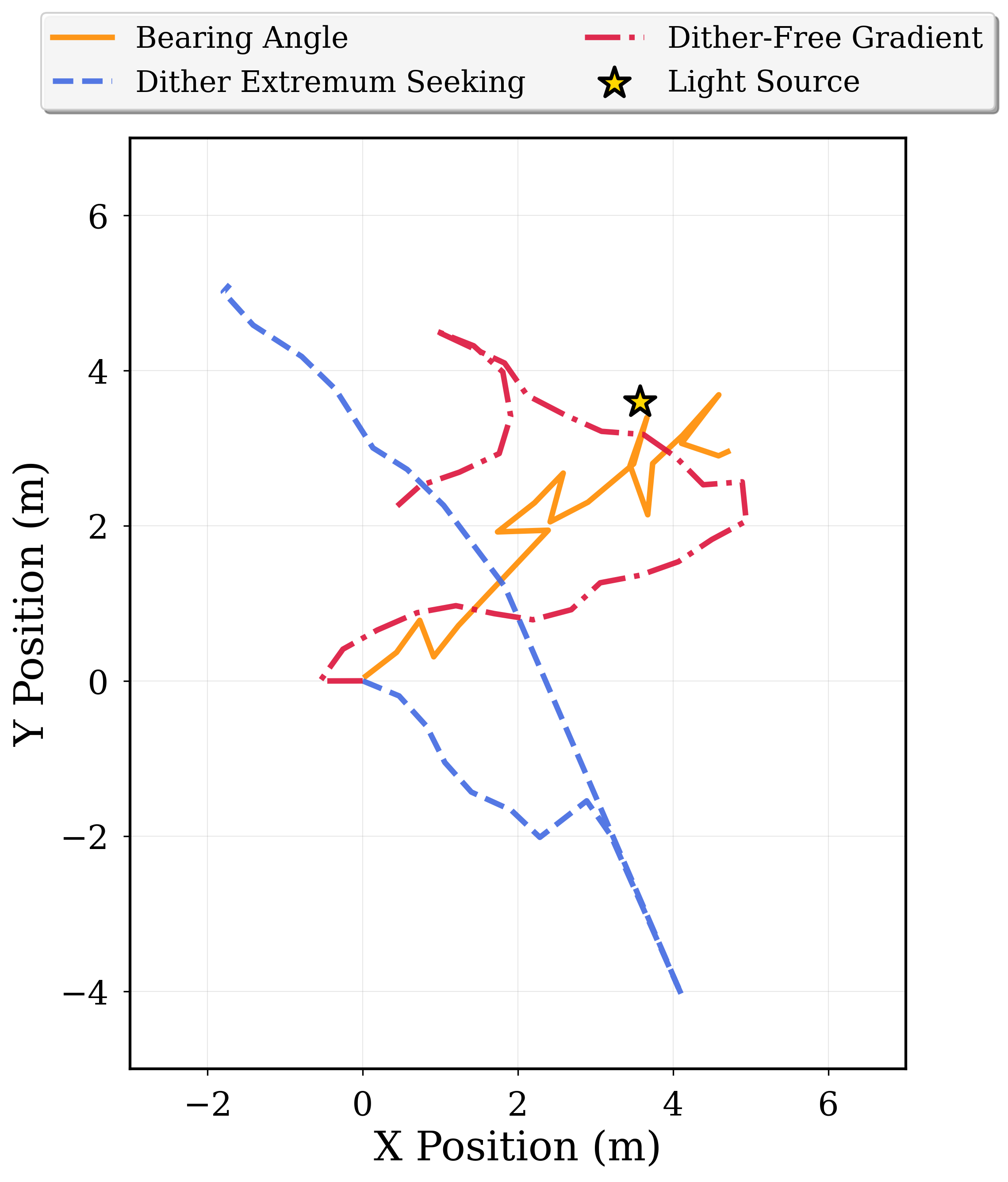}
        \caption{Outdoor travel trajectory (Wind: 8\,m/s)}
        \label{fig:outdoorTraj}
    \end{subfigure}
    \caption{Comparison of travel path length and time for three navigation algorithms}
    \label{fig:flightTrajectory}
\end{figure}

\textbf{Results and Analysis}. The resulting flight paths, in ~\autoref{fig:indoorTraj}, show clear performance differences. BAG (orange) produces a smooth and direct trajectory. On the other hand, DES (blue) shows constant wavering, an inefficient motion inherent to its probing-based logic. DGA (red) produces a more direct path than DES, but its trajectory shows moments of straight flight followed by sharp corrective turns as its gradient estimate is updated.

The quantitative results in \autoref{fig:indooPathQuant} confirm these qualitative observations. BAG is the most efficient, completing the task with the shortest path (7.3\,m). DES is the least efficient, as expected, requiring the longest path (12.2\,m). DGA achieves a reasonably efficient path length of 8.6\,m, much closer to BAG than to DES.
However, the travel time in \autoref{fig:outdooPathQuant} captures a different trend. BAG is the fastest (8.4\,s), consistent with its direct path. However, DGA, despite its efficient path, is significantly slower (21.4\,s). This delay is a key finding: it is not caused by inefficient flight (like DES) but by the computational overhead of its gradient estimation and the time required to stop and make sharp corrective turns.

Based on these findings, the BAG algorithm is the superior solution for this source-seeking problem, demonstrating the highest efficiency in both path length and travel time. Although DGA's path is efficient, its significant computational and maneuvering delays make it less effective.

\subsection{Navigation Evaluation in \textbf{Outdoor} environment}

Although our primary use case is indoors, we conduct an outdoor evaluation to stress-test the robustness and practical limitations of our three navigation algorithms. This outdoor setting introduces the primary challenge of wind, a significant complicating factor for any LTA platform given its large surface area and susceptibility to aerodynamic forces. %This test is particularly important as it validates the real-world viability of our novel ``point-and-go'' strategies. %It is important to recall that while related concepts like DES have been tested with light on standard drones \cite{wu2020inflightrangeoptimizationmulticopters} and BAG with RF signals \cite{6580223}, our implementation of BAG for aerial light guidance and our DGA algorithm are novel. 
Therefore, this experiment moves beyond a sterile lab environment to determine if these novel algorithms are resilient enough to handle dynamic, unpredictable disturbances.

\begin{figure}[tb!]
    \centering
    % \begin{minipage}[b]{0.46\linewidth}
    %     \centering
    %     \includegraphics[width=\linewidth]{img/distance_to_light_20251014_161920.png}
    %     \captionof{figure}{Distance Versus Time For Outdoor Experiment.}
    %     \label{fig:outdoorDrone}
    % \end{minipage}
    % \hspace{5mm} % Adds horizontal space
    \begin{minipage}[b]{0.46\linewidth}
        \centering
        \includegraphics[width=\linewidth]{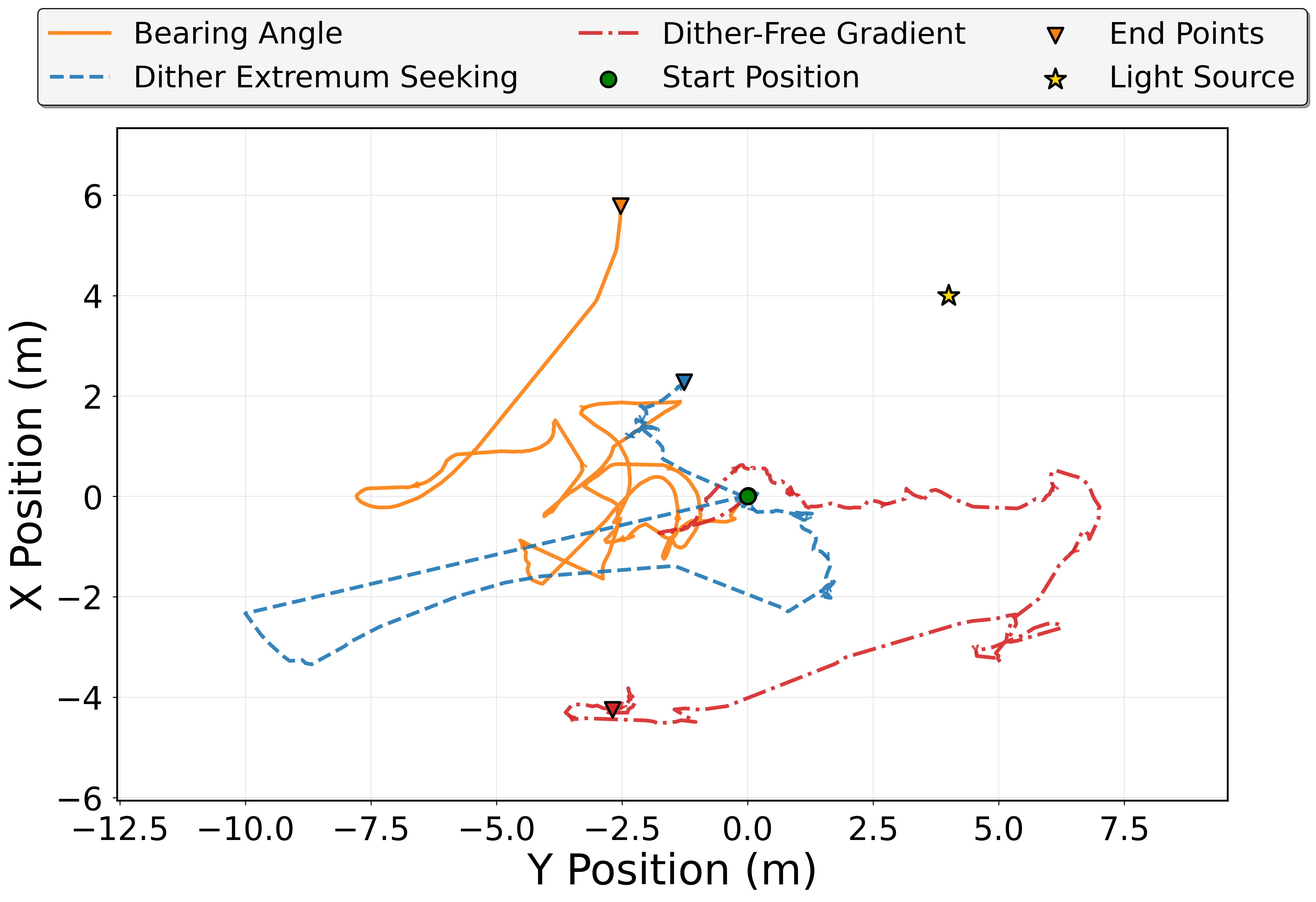}
        \captionof{figure}{Strong Wind (14\,m/s):  outdoor fail case.}
        \label{fig:failCase}
    \end{minipage}
\end{figure}

\subsubsection{Case 1: Performance in Moderate Wind}
We first evaluate the algorithms under moderate wind outdoors (around 8\,m/s, 30\,km/hr). The results show a clear difference in robustness. The flight paths in \autoref{fig:outdoorTraj} %and the distance-to-target plot in \autoref{fig:outdoorDrone} 
show that BAG is the only algorithm to successfully converge.  Although its trajectory is jittery, BAG's reliance on an instantaneous, direct measurement of the light's bearing allows it to continuously correct for wind gusts, reaching the 1-meter success threshold in under 20\,s.
In contrast, both single-sensor methods fail in this condition, each for distinct reasons. The DES method fails catastrophically as its logic depends on correlating small, controlled dither inputs with light changes. Unpredictable wind gusts overpower these controlled movements, completely decoupling the algorithm's measurements from its commands and causing it to fly away from the target. The DGA method also fails to converge. Its reliance on IMU-based dead reckoning to build a gradient map is severely corrupted by wind-induced drift. As the wind pushes the drone, its internal position estimate becomes inaccurate, leading to the severe overshooting seen in its trajectory.

\subsubsection{Case 2: Performance in High Wind}
We then test the platform in a high wind environment (14\,m/s, 50\,km/hr), with results shown in \autoref{fig:failCase}. In this scenario, no algorithm could successfully navigate to the target. Even the robust BAG algorithm, while actively attempting to correct its heading, is ultimately overwhelmed. The wind forces exceed the stabilization capacity of the motors, making guided navigation impossible regardless of the algorithm's strategy.

\textbf{Analysis.}
These outdoor experiments confirm that the navigation strategy is critical for robustness. The single-sensor approaches (DES and DGA) are unsuited for dynamic environments because their guidance logic relies on wiggling motions or historical data, which wind disturbances immediately corrupt. The BAG algorithm is far more resilient due to its reliance on instantaneous, direct measurements. Our tests show that our novel implementation of BAG with visible light provides reliable guidance in moderate wind, but its operation is ultimately bounded by the LTA platform's physical thrust limitations in high-wind conditions. The evaluation under different wind conditions is a valuable contribution of our work. Prior studies state that their LTA platforms can handle wind conditions, but they do not perform empirical studies, they only provide estimates for milder wind conditions than our platform: BEAVIS < 8\,m/s; Skye < 3.5\,m/s \cite{sharma2023beavis, burri2013design}.

\section{Related Work}

Our work draws upon insights from three domains to attain a persistent, autonomous LTA drone. This requires addressing challenges in platform design, autonomous navigation, and energy endurance. This section reviews the state-of-the-art in these three areas. 

\subsection{LTA Simulators}

Simulators are a critical tool for designing and testing drones, and while many simulators exist for standard multi-rotor drones \cite{shah2017airsimhighfidelityvisualphysical, kong2022marsimlightweightpointrealisticsimulator, yang2021digital, panerati2021learningflygym}, the nascent field of hybrid LTA drones lacks such robust simulation support. The unique aerodynamics of LTA platforms, particularly their large surface area, complex drag forces, and specific propeller-envelope interactions, are not captured by conventional drone simulators. The literature has few dedicated LTA simulators, each with significant limitations.

A holistic work in this area is done by Zufferey et al. which comprehensively models their blimp platform and verifies its accuracy using data from real-world experiments \cite{zufferey2006flying}. However, this framework has two major limitations: it still relies on physical experiments to inform the model and it does not integrate a flight controller into the simulation. Another recent simulator addresses the second limitation by integrating a controller \cite{Price_2022}. This work, however, does not simulate the key aerodynamic forces from first principles. It only provides a simple function to approximate the aerodynamic effect.

Our work addresses these gaps by introducing two key contributions. First, we integrate a CFD component to model complex aerodynamics directly from a 3D design, replacing the need for physical experiments and imprecise approximations. Second, we ``close the loop'' by integrating a complete flight controller system, which was missing from the original framework. This transforms the model into a true flight dynamics simulator, capable of evaluating a platform's design, stability and autonomous performance.

\subsection{Localization Systems for UAVs}

% Other localization methods 
Autonomous drone navigation in GPS-denied environments has been addressed with various technologies, and there is no one-size-fits-all solution. Every technology poses distinct trade-offs, including those based on light. Radio-frequency (RF) systems like Wi-Fi, Bluetooth, and Ultra-Wideband (UWB) have different trade-offs. Wi-Fi and Bluetooth can leverage existing infrastructure, making them low-cost, but they suffer from low accuracy of several meters and high latency \cite{7039067, 7120085, 6823667, Ariante2022}. UWB offers robust, centimeter-level accuracy, but requires deploying a costly and complex network of base stations and has a limited range \cite{10.1109/msp.2005.1458289}. 
Another line of research uses sound-based systems~\cite{chevtchenko2025drone}. These methods are resilient to low-light conditions and provide good accuracy, but require a dense infrastructure, are susceptible to noise, and can raise privacy concerns. 

% VLP
Vision-based methods, relying on cameras, are also widely used \cite{5646820}, but they can be computationally intensive and unreliable in feature-poor or poorly-lit environments \cite{electronics12071533}, and similar to sound-based systems, cameras can raise privacy concerns. As an alternative, Visible Light Positioning (VLP) leverages existing LED lighting infrastructure, offering a low-cost, high-accuracy solution using simple photodiodes  \cite{s20051382, 9780735, zhu2024survey, bastiaens2024visible, yang2015wearables, yasir2014indoor}. 
Many of these traditional VLP methods are, however, unsuitable for a lightweight, low-cost drone, as drones have 6-DoF movement and have constantly changing tilting motions.
The majority of existing VLP research is validated only for static receivers that remain parallel to the light source, an assumption that is immediately invalidated by a flying drone \cite{li2014epsilon, zhuang2019-vlp-noise-mitigation, plets2019efficient}.
Methods that do attempt a full 3D localization are often infeasible for a lightweight, low-cost drone due to high complexity. For example, angle-based approaches can achieve a high accuracy but require bulky, complex multi-sensor receivers or custom-built transmitters that are costly and difficult to scale \cite{6823667, csahin2015hybrid}. Signal-strength-based methods, while simpler in hardware, often rely on computationally intensive algorithms (like particle swarm optimization) that are ill-suited for a drone's limited onboard processor \cite{cai2017-vlp-pso}.
Only a few studies have implemented VLP methods for \textbf{mobile} drones. For example, one study reports an average error of $\approx\!4$\,cm using 4 LEDs~\cite{Niu2021ICC}, while another single-photodiode approach, using also four LEDs, reports an accuracy of $\approx\!20.6$\,cm under mobile (6-DoF) flight~\cite{10257235}. These studies, however, require a high LED density: 4 LED/m$^2$ for~\cite{Niu2021ICC} and $1$\,LED/m$^2$ for~\cite{10257235}

Our work takes a different approach by focusing on source-seeking rather than high-precision localization. The objective is not to estimate precise Cartesian coordinates, but to enable a simple and robust ``point-and-go'' navigation system. This method avoids the need for a dense, complex infrastructure by operating with a single light beacon that can be deployed every 150\,m$^2$.

% Our contribution
% ⁠Then we highlight our contribution, stating clearly that we propose a novel method but we do not estimate cartesian coordinates (to be fari and avoid overselling ourwork)

\subsection{Extending the Flight Time of LTAs}

Overcoming the limited flight time of battery-powered drones is a critical research challenge. Several strategies exist, but each comes with significant drawbacks. Tethered systems, for instance, provide continuous power from a ground station but inherently restrict the drone's operational range to the cable's length and introduce complex drag forces \cite{walendziuk2020power, dynamics5020017, al2021self}. Wireless power transfer (WPT) using lasers or inductive fields is also being explored, but these methods remain highly experimental and are severely limited in range, often requiring the drone to hover over a specific pad or stay within a few meters of a source \cite{anand2015wireless, en18020351, junaid2016design}.

Onboard solar harvesting is the most promising approach for untethered, autonomous endurance, but it presents fundamental trade-offs for different types of UAVs. Fixed-wing drones offer large wing surfaces ideal for mounting panels but are inefficient or incapable of hovering \cite{chu2021development, articleLiller, oettershagen2016perpetual}. Conversely, multi-rotor drones, which excel at hovering, suffer from a critically limited surface area. This forces researchers to add external structures to mount panels, which introduces parasitic weight and complexity, often negating the energy gains \cite{abidali2024development, goh2019100}.

LTA platforms offer a promising solution to this trade-off, uniquely combining a large envelope surface for energy harvesting with an intrinsic, near-zero-power ability to hover. Despite this clear advantage, research in this specific area is sparse. Existing work has provided only simulations for the placement of panels on an envelope (without empirical validation)~\cite{manikandan2024parametric}, or presented preliminary measurements of the power harvested from a single LTA design \cite{articlelifeNano}. However, no work has provided a systematic framework for comparing the efficiency of different solar cell types for this application or, more critically, analyzed how the placement of rigid panels impacts the LTA's aerodynamic stability and flight performance. Our work directly addresses this gap by providing this comparative analysis, balancing energy harvesting with flight stability.

\section{Conclusion}

Our work establishes a practical pathway toward energy-autonomous, light-guided LTA drones. Based on our high-fidelity simulator, we identified a stable LTA configuration that we enhanced with lightweight solar harvesting and a novel, ``point-and-go'' light-based navigation algorithm. Our evaluations in controlled and real-world environments show a net-positive energy harvesting and demonstrate reliable 7-meter navigation to a single light beacon. These results confirm the platform's potential for persistent, self-sustaining operation using a simple infrastructure. With this foundation, the remaining challenges to bring persistent LTA drones closer to practical deployment are enhancing flight stability in higher-wind conditions, integrating energy-aware mission planning, and expanding the single-beacon guidance logic to more complex, multi-target navigation. Overall, our findings mark a first step and widen the design space for compact, energy-autonomous aerial systems, laying the groundwork for translating the promise of LTA drones into a practical reality for long-duration applications.

%%
%% The acknowledgments section 
\begin{acks}
This project has received funding from the European Union’s EU Framework Programme for Research and Innovation under the HORIZON-MSCA-DN-2022 Grant Agreement N°101119652.
\end{acks}

% \bibliographystyle{ACM-Reference-Format}

\end{document}